\documentclass[runningheads]{llncs}

\usepackage[year=2026,ID=13621]{eccv}
\usepackage{eccvabbrv}

\usepackage{graphicx}
\usepackage{booktabs}

\usepackage[accsupp]{axessibility}  % Improves PDF readability for those with disabilities.

\usepackage{hyperref}
\hypersetup{
    colorlinks=true,
    urlcolor=magenta,
}
\usepackage{orcidlink}
\usepackage{marvosym}
\usepackage{algorithm}
\usepackage{algpseudocode}
\usepackage{array}
\newcommand{\corrauth}{\textsuperscript{\textrm{\Letter}}}
\begin{document}

% ---------------------------------------------------------------
% TODO REVIEW: Replace with your title
\title{DICT: Data Injection and Contrastive Trajectory Refinement for Conditional Image Generation with Diffusion Models} 

% TODO REVIEW: If the paper title is too long for the running head, you can set
% an abbreviated paper title here. If not, comment out.
\titlerunning{DICT}

% TODO FINAL: Replace with your author list. 
% Include the authors' OCRID for the camera-ready version, if at all possible.

\author{Chunnan Shang \and
Xin Zhang\and 
Zhizhong Wang \corrauth \and
Hongwei Wang \corrauth }

% TODO FINAL: Replace with an abbreviated list of authors.
\authorrunning{C.~Shang et al.}
% First names are abbreviated in the running head.
% If there are more than two authors, 'et al.' is used.

% TODO FINAL: Replace with your institution list.
\institute{Zhejiang University, Hangzhou, China\\
\email{ \{chunnan.22, hongweiwang\}@intl.zju.edu.cn \\
\{22421264, endywon\}@zju.edu.cn} \\
\url{https://github.com/scn-00/DICT}
}

\maketitle
\let\thefootnote\relax\footnotetext{\corrauth Corresponding authors.}

\begin{abstract}
Diffusion models have become a dominant paradigm for conditional image generation, yet existing approaches generally follow two directions: task-specific designs that can improve performance but limit generalization, and training-free loss guidance that compresses rich conditions into scalar objectives and applies stepwise guidance, leading to information bottlenecks and error accumulation along the sampling trajectory. Given the urgent need for an effective unified framework across diverse conditional image generation tasks, we propose {\bf D}ata {\bf I}njection and {\bf C}ontrastive {\bf T}rajectory Refinement ({\bf DICT}), a training-free inference method that enhances conditional image generation without introducing task-dependent architectures. DICT introduces Data Injection, where noise-perturbed conditional signals are integrated into early denoising stages; by performing guided denoising on these injected signals, DICT adaptively selects and distills task-salient information from the raw condition, effectively preserving spatial richness and ensuring precise condition-to-generation alignment. Furthermore, DICT applies Contrastive Trajectory Refinement across adjacent denoising states, enabling pairwise comparisons that progressively improve sample quality. These designs keep inference simple while improving cross-task transfer under a unified diffusion formulation. Extensive experiments on conditional image generation tasks (e.g., style transfer, image super-resolution, and image deblurring) show consistent gains in fidelity and perceptual quality over representative task-specific and loss-guided baselines.
\end{abstract}

\section{Introduction}
\label{sec:intro}

Diffusion models (DMs)~\cite{ho2020denoising,rombach2022high} have become a dominant paradigm for conditional image generation, largely due to their strong priors and flexible sampling dynamics. In many practical conditional image generation settings (e.g., style transfer, image super-resolution, and image deblurring), the informative value of the given conditions is often restricted. For instance, a condition might capture merely partial target features~\cite{lei2025stylestudio,liu2024stylecrafter} or represent a corrupted observation~\cite{kim2025flowdps,wu2024seesr}. Such suboptimal conditions inherently hinder faithful generation, requiring the model to extract useful signals while suppressing misleading cues.

Existing approaches can be categorized into two major classes of methods: (1) {\it condition-specific designs}, which tailor architectures or constraints to particular condition types, such as feature injection and fusion for style transfer~\cite{liu2024stylecrafter,deng2024z,wang2024instantstyle} or data-consistency and trajectory relaxation for image super-resolution and image deblurring tasks~\cite{kim2025flowdps,rout2023solving,xu2025rethinking}; and (2) {\it generic inference-time loss guidance}, which formulates condition satisfaction as a scalar loss and utilizes its gradients to guide the reverse diffusion trajectory without additional training~\cite{ye2024tfg,yu2023freedom,castillo2025adaptiveag}.

While broadly applicable, loss-guided inference suffers from two key limitations. First, it compresses the condition reference into a single scalar loss, so guidance is driven primarily by the gradient of this objective. This scalarization creates an information bottleneck: the structured, high-dimensional content in the condition reference is not fully retained in the resulting gradient signal, leading to updates that are often too coarse to provide fine-grained control in complex tasks. For instance, in style transfer, a loss function might indicate how far a stylized image is from the target style overall, but it lacks the granular information needed to specify specific elements like texture patterns and brush strokes, which exist only within the high-dimensional data itself and are not reducible to a single scalar. Furthermore, loss guidance using scalar objectives leads to compounding error accumulation. In the iterative denoising process, inaccuracies at any step persist and propagate along the sampling trajectory, increasingly biasing the generation and degrading final sample quality. Alternatively, while latent initialization methods like SDEdit~\cite{meng2021sdedit} avoid scalarization by adding noise to a reference image and denoising it, they typically employ a one-shot integration strategy. This lacking of continuous condition anchoring often leads to trajectory drift, and their inability to adaptively filter out irrelevant condition information restricts their general efficacy.

To address these issues, we propose {\bf D}ata {\bf I}njection and {\bf C}ontrastive {\bf T}rajectory Refinement ({\bf DICT}), a general inference framework for conditional image generation that goes beyond scalar loss guidance and avoids task-specific redesign. Our motivation is twofold. First, although task-specific architectures or constraints can better exploit a given condition, they often encode condition-dependent assumptions and thus transfer poorly across tasks and condition types. To improve generality, DICT avoids explicit constraints and instead leverages the {\it condition as data}: we diffuse the condition signal and inject its noise-perturbed counterpart into the early denoising steps. This design allows the model to access richer conditional information than a scalar loss, selectively amplifying informative cues while suppressing redundant or degraded components. Second, even with a better condition representation, stepwise loss optimization remains prone to error accumulation along the reverse process. To stabilize sampling in a model-agnostic manner, compatible with U-Net~\cite{rombach2022high} and Transformer-based backbones~\cite{peebles2023scalable}, we exploit the temporal structure of the reverse diffusion process and introduce a {\it Contrastive Trajectory Refinement} mechanism. Concretely, we optimize a contrastive trajectory objective that enforces step-to-step improvement, ensuring that the prediction at each timestep is more aligned with the condition signal than the prediction at the previous timestep, thereby reducing error propagation and promoting consistent refinement along the sampling path.

Overall, our main contributions are as follows:
\begin{itemize}
  \item We analyze limitations of existing conditional image generation methods: condition-specific designs can improve task performance but generalize poorly across tasks, while generic inference-time loss guidance compresses rich conditions into scalar objectives, leading to information bottlenecks and error accumulation along the sampling process.
  \item We propose {\bf D}ata {\bf I}njection and {\bf C}ontrastive {\bf T}rajectory Refinement ({\bf DICT}), a training-free inference method that enhances conditional image generation without introducing task-dependent architectures. DICT injects noise-perturbed conditional signals in early denoising steps to strengthen condition alignment, and further refines the reverse process via a contrastive trajectory objective across adjacent denoising states to progressively improve quality.
  \item Extensive experiments on conditional image generation tasks (e.g., style transfer, image super-resolution, and image deblurring) demonstrate consistent improvements in fidelity and perceptual quality over representative task-specific and loss-guided baselines.
\end{itemize}

\section{Related Work}
\label{sec:relatedwork}

{\bf Style Transfer.} We focus on text-driven style transfer, which aims to generate images that follow the text content while adopting the style of a reference image. Existing methods can be divided into two categories: those that learn feature mappings and those that exploit prior features. Feature-mapping approaches learn to project style/content attributes into dedicated spaces and model their relations, e.g., mapping style features into the diffusion model’s CLIP space (StyleShot~\cite{gao2025styleshot}), learning a style feature space from CLIP image features (StyleCrafter~\cite{liu2024stylecrafter}), or explicitly decomposing CLIP image features into content and style representations (CSGO~\cite{xing2024csgo}, DEADiff~\cite{qi2024deadiff}); related work also maps style attributes via lightweight tuning with human feedback (StyleDrop~\cite{sohn2023styledrop}). Prior-feature exploitation approaches instead leverage pretrained representations to extract or inject style cues, such as subtracting text content embeddings from image embeddings (InstanStyle~\cite{wang2024instantstyle}), decoupling content via ControlNet~\cite{zhang2023adding} to obtain cleaner style features (StyleStudio~\cite{lei2025stylestudio}), or injecting and aligning attention features (queries/keys/values) from the style image (StyleAlign~\cite{hertz2024style}).

{\bf Super-resolution and Deblurring.} We consider super-resolution and deblurring, recovering a clean image from a low-resolution or blurred observation. Existing approaches mainly follow two directions: methods based on strict constraints and methods based on flexible sampling. Constraint-based methods enforce explicit data-consistency to match the degraded observation, e.g., multi-stage consistency (SITCOM~\cite{alkhouri2025sitcom}), correction/projection constraints (PSLD~\cite{rout2023solving}, ~\cite{xu2025rethinking}), or lightweight gradient-based consistency enforcement (DCDP~\cite{li2024decoupled}). Flexible sampling methods relax or reinterpret the reverse diffusion process, such as viewing posterior sampling as Bayesian filtering (FPS-SMC~\cite{dou2024diffusion}), formulating sampling as discretized optimal control (DOC~\cite{li2024solving}), or steering via vector-field objectives and scheduled constraint relaxation (FlowChef~\cite{patel2024steering}, FlowDPS~\cite{kim2025flowdps}). Beyond these general strategies, many task-specific designs have also been explored for super-resolution, e.g., SeeSR~\cite{wu2024seesr} and InvSR~\cite{yue2025arbitrary}.

Compared to task-specific methods, DICT directly conditions on the given image and adaptively extracts useful cues by injecting noise-perturbed conditional signals into early denoising steps and denoising them along the reverse process, avoiding biases from learned condition-dependent feature mappings and the partial information captured by fixed pretrained features. Moreover, DICT does not rely on strict task-specific consistency constraints or modified sampling procedures; instead, it leverages the conditional input to apply a Contrastive Trajectory Refinement objective across adjacent denoising states, enabling pairwise step-to-step comparisons that progressively improve condition alignment and sample quality in a unified and more general manner.

{\bf Loss-gradient Guidance.} Training-free loss-guided methods have emerged as a versatile paradigm to incorporate complex constraints into pretrained diffusion models without the need for task-specific fine-tuning. These methods typically leverage the gradients of external loss functions to shift the sampling trajectory towards a target region defined by the condition. For instance, FreeDoM~\cite{yu2023freedom} achieves conditional generation by utilizing off-the-shelf networks to construct energy functions, employing an iterative time-travel strategy to enforce semantic consistency. TFG~\cite{ye2024tfg} unifies them into a comprehensive design space, proposing an efficient search strategy to dynamically optimize guidance parameters for various tasks. However, a major bottleneck of these methods is the increased computational overhead caused by repeated gradient computations. To address this, Adaptive Guidance~\cite{castillo2025adaptiveag} accelerates inference by adaptively omitting the unconditional score evaluation once it converges with the conditional estimate.

Unlike previous methods, DICT goes beyond scalar loss gradients by injecting a noise-perturbed condition in early denoising steps to exploit rich conditional signals. It further introduces a Contrastive Trajectory Refinement that enforces step-to-step improvement, mitigating error accumulation, and improving quality.

\section{Method}

\subsection{Revisit Latent Diffusion Models}

The latent diffusion models~\cite{rombach2022high} operate in latent space, using a forward process that gradually adds noise to reach a standard Gaussian and a reverse process that steadily denoises to generate samples.

Thus, the clean latent vector $z_0$ can be first obtained:
\begin{equation}
\begin{aligned}
z_0  = E(x),
\label{eq:1}
\end{aligned}
\end{equation}
where $E$ is an encoder of~\cite{rombach2022high}. $x$ denotes the clean data.

The forward diffusion is a Markov chain, where the latent vector $z_t$ at each state $t \in (1,T)$ is obtained by adding Gaussian noise to the clean latent vector $z_{0}$:

\begin{equation}
\begin{aligned}
z_t = \sqrt{\bar{\alpha}_t} z_{0} + \sqrt{1 - \bar{\alpha}_t} \epsilon,  
\label{eq:2}
\end{aligned}
\end{equation}
where $\bar{\alpha}_t = \prod_{s=1}^{t} (1-\beta_t) $, $\beta_t \in (0,1)$ adopts a fixed variance schedule. $\epsilon \sim \mathcal{N}(0, \mathbf{I})$ is the noise. 

The reverse diffusion starts from a Gaussian sample $z_T \sim \mathcal{N}(0, \mathbf{I})$ and progressively denoises. Based on PLMS~\cite{liu2022pseudo}, each preceding state $z_{t-1}$ is obtained: 
\begin{equation}
\begin{aligned}
z_{0|t} =& \frac{z_t - \sqrt{1 - \bar{\alpha}_t} \, \epsilon_\theta(t)}{\sqrt{\bar{\alpha}_t}},\\
z_{t-1} = \sqrt{\alpha_{t-1}} \, z_{0|t} + &\sqrt{1 - \alpha_{t-1} - \sigma_t^2} \epsilon_\theta(t) + \sigma_t z, 
\label{eq:3}
\end{aligned}
\end{equation}
where $z \sim \mathcal{N}(0, \mathbf{I})$ and $\epsilon_\theta(t)=\sum_{i=0}^{k-1}w_i\epsilon_\theta(z_{t-i},t-i)$. $k=4$ is for the historical steps. $w_i$ is the weighting coefficient. $\epsilon_\theta$ is predicted noise, implemented by a backbone such as U-Net~\cite{rombach2022high} or DiT~\cite{peebles2023scalable}. For this paper, we use the U-Net as our foundational model. $\sigma_t^2=0$ for deterministic sampling. 

Finally, the target image $x$ can be obtained via decoding:
\begin{equation}
\begin{aligned}
x = D(z_{0}),
\label{eq:4}
\end{aligned}
\end{equation}
where $D$ represents the decoder of ~\cite{rombach2022high}.

\subsection{Task Setup for Conditional Image Generation}

Conventionally, super-resolution and deblurring are treated as fundamentally distinct from text-driven style transfer due to their differing natures—explicit physical degradations versus abstract semantic conditions. To bridge this conceptual gap, we mathematically frame them as generalized posterior sampling, providing a unified formulation under conditional image generation.

Formally, let $y$ denote the unknown clean image (or latent) to be generated, and let $x$ denote the available condition, which can be a semantic/style specification (e.g., text and a style reference image) or a degraded observation (e.g., low-resolution/blurred image). Both tasks can be cast as sampling from the conditional distribution:
\begin{equation}
\begin{aligned}
y \sim p(y|x), \quad p(y|x) \propto p(x|y)p(y),
\label{eq:5}
\end{aligned}
\end{equation}
where $p(y)$ is the diffusion prior learned from data, and $p(x|y)$ measures the compatibility between the generated image $y$ and the condition $x$. Equivalently, we can target the Maximum a Posteriori (MAP) estimation:
\begin{equation}
\begin{aligned}
y^* = arg \min\limits_{y} - logp(x|y) - logp(y).
\label{eq:6}
\end{aligned}
\end{equation}

{\bf Style Transfer.} For text-driven style transfer, the condition $x=(t,s)$ combines a text prompt $t$ (content) and a style image $s$ (style). Under our unified posterior view:
\begin{equation}
p(y|t,s) \propto p(t,s|y)p(y).
\label{eq:7}
\end{equation}

Assuming $t$ and $s$ are conditionally independent given $y$, we can factorize the conditional term as $p(t,s|y) = p(t|y)p(s|y)$. This leads to:
\begin{equation}
p(y|t,s) \propto p(s|y) p(y|t),
\label{eq:8}
\end{equation}
where $p(y|t) \propto p(t|y)p(y)$ represents the text-conditioned prior. In practice, $p(y|t)$ is implicitly learned by a pretrained diffusion model (e.g., via classifier-free guidance), while $p(s|y)$ is explicitly defined using an energy-based surrogate:
\begin{equation}
-\log p(s|y) \approx \mathcal{L}_{style}(y,s).
\label{eq:9}
\end{equation}
Therefore, MAP inference under these conditions can be formulated as:
\begin{equation}
y^* = \arg \min\limits_{y} \mathcal{L}_{style}(y,s) - \log p(y|t),
\label{eq:10}
\end{equation}
where the gradient of $-\log p(y|t)$ is provided by the conditioned score function of the diffusion model.

{\bf Super-resolution and Deblurring.} For super-resolution or deblurring, the condition $x$ results from a known degradation operator $\mathcal{A}(\cdot)$:
\begin{equation}
\begin{aligned}
x = \mathcal{A}(y)+\epsilon, \quad \epsilon \sim \mathcal{N}(0, \sigma^{2}\mathbf{I}),
\label{eq:11}
\end{aligned}
\end{equation}
which yields a standard data-fidelity term $- logp(x|y) \propto ||x-\mathcal{A}(y)||_{2}^2$. Therefore, MAP inference under degraded conditioning can be written as:
\begin{equation}
\begin{aligned}
y^* = arg \min\limits_{y}  \frac{1}{2\sigma^2}||x-\mathcal{A}(y)||_{2}^2 - logp(y).
\label{eq:12}
\end{aligned}
\end{equation}

This unified view motivates inference procedures that exploit conditional signals while remaining robust to condition information.

\subsection{Overall Framework of DICT}

\begin{figure*}[t]
  \centering
   \includegraphics[width=1\linewidth]{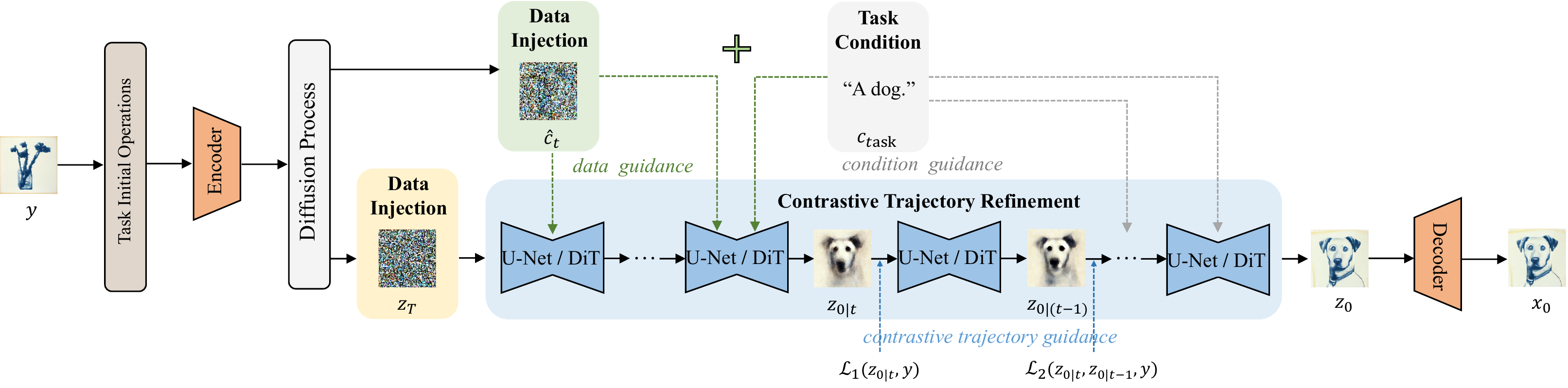}
   \caption{Overview of DICT. Data Injection is integrated into the early reverse diffusion stages to retain conditional richness, leveraging guided denoising to adaptively extract task-salient information from the perturbed condition. Simultaneously, Contrastive Trajectory Refinement spans the entire trajectory to ensure coherence and refine the generative path for superior quality. Task-specific conditions, where applicable, remain persistent throughout the process to ensure consistent guidance.}
   \label{fig:1}
\end{figure*}

We introduce {\bf D}ata {\bf I}njection and {\bf C}ontrastive {\bf T}rajectory Refinement, hereafter {\bf DICT}, for conditional image generation (Fig.~\ref{fig:1}), a general inference method beyond loss guidance. It injects noise-perturbed conditions in early denoising and enforces step-to-step improvement with a contrastive trajectory objective.

{\bf Data Injection.} In conditional image generation, the fidelity of the output is dictated by how effectively the condition $x$ is integrated into the sampling trajectory. Existing training-free methods predominantly rely on a scalar-guidance paradigm, where rich, high-dimensional conditions $x$ are compressed into a single scalar loss objective. This drastic reduction in dimensionality creates a severe information bottleneck: a scalar value is fundamentally insufficient to encapsulate the spatial complexity and fine-grained textures of the condition. Consequently, the resulting gradients often fail to recover subtle cues, leading to a significant loss of fidelity as the model hallucinates details to fill the information void. 

\begin{figure}[t]
  \centering
   \includegraphics[width=1\linewidth]{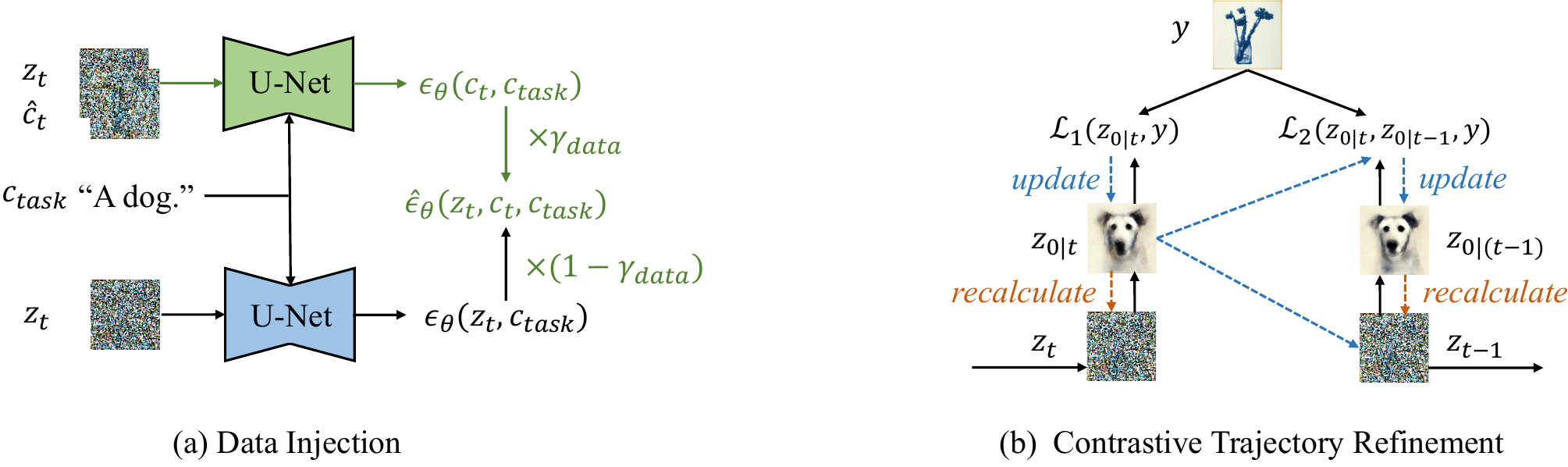}
   \caption{The detailed schematics illustrate the Data Injection and Contrastive Trajectory Refinement within the DICT framework. }
   \label{fig:2}
\end{figure}

To circumvent this information collapse, we propose a direct Data Injection mechanism. Instead of relying solely on a simplified scalar proxy, we incorporate noise-perturbed versions of the condition $x$ directly into the diffusion inference loop during the early denoising stages (Fig.~\ref{fig:1}, Fig.~\ref{fig:2}(a)). By denoising these injected signals alongside the latent variables throughout the reverse process, the model can adaptively distill task-relevant information from the raw condition. This enables the generative trajectory to be grounded in the target data manifold, ensuring that the final output preserves high-frequency details and structural nuances that are typically lost in traditional loss-guided frameworks.

Specifically, given a condition $x$, we choose an operation $\mathcal{M}(\cdot)$ to transform $x$ into a higher-quality initial condition that amplifies useful signals for guidance:
\begin{equation}
 \hat{x} = \mathcal{M}(x), 
\label{eq:13}
\end{equation}
where $\mathcal{M}$ is a task-dependent but lightweight preprocessing step that better aligns the condition with the guidance objective. It is worth noting that although $\mathcal{M}$ is inherently task-specific (analogous to task-dependent loss functions), it introduces no trainable parameters and does not modify the internal structure of the base model. Consequently, our ``architecture-independent'' claim remains valid. Details of $\mathcal{M}$ are provided in Sec.~B of the Supplementary Materials.

We then use the processed condition $\hat{x}$ to initialize the reverse diffusion with $z_T$ at time step $T$, and to compute the condition latent $\hat{c}_t$  used at each denoising step $t$:
\begin{equation}
\begin{aligned}
\hat{z} =\; &E(\hat{x}), \\
z_{T} = \sqrt{\bar{\alpha}_T}\hat{z} &+ \sqrt{1-\bar{\alpha}_T}\epsilon,\\
\hat{c}_{t} = \sqrt{\bar{\alpha}_t}\hat{z} + \sqrt{1-\bar{\alpha}_t}&\epsilon_{ti}, i=1,\dots, N_{iter1},
\label{eq:14}
\end{aligned}
\end{equation}
where $\epsilon \sim \mathcal{N}(0, \mathbf{I})$. For $i=1$, we set $\epsilon_{ti} = \epsilon$; otherwise, $\epsilon_{ti} = \epsilon_\theta(t)$ (Eq.~\ref{eq:15}). $N_{iter1}$ denotes the number of iterations. The condition latent $\hat{c}_{t}$ is a noise-perturbed version of the condition $x$ that preserves its high-dimensional spatial richness. Rather than injecting the condition data directly, we add noise to it to facilitate selective information exploitation. This strategy is grounded in the observation that task-relevant features—such as stylistic textures in style transfer or semantic structures in image restoration—typically manifest as the dominant signals within the image space. Stochastic noise effectively disrupts redundant and irrelevant information, whereas the dominant conditional structures remain resilient and perceptible to the diffusion prior, thanks to their high-dimensional robustness. Consequently, compared to conventional guidance methods that compress rich conditions into a single scalar loss—thereby creating an irreversible information bottleneck—our noise injection strategy effectively maintains the integrity of the conditional cues. By grounding the sampling process in the noise-perturbed image manifold rather than a simplified scalar proxy, DICT enables the model to adaptively distill task-salient signals while suppressing harmful interference, leading to superior generation quality and fidelity.

Subsequently, we inject lightweight variants of the condition data into the early reverse-diffusion steps to improve output fidelity and sampling efficiency:
\begin{equation}
\begin{aligned}
c_t  = \alpha&_{data} \times z_t + (1-\alpha_{data}) \times \hat{c}_{t},\\
\quad \quad \hat{\epsilon}_\theta(z_t, c_t, c_{task}) & = \gamma_{data}   \times \epsilon_\theta(c_{t}, c_{task}) \\
  &  \qquad  + (1 - \gamma_{data} ) \times \epsilon_\theta(z_t, c_{task}), \\
\epsilon_\theta(t) & = \hat{\epsilon}_\theta(z_t, \hat{c}_t, c_{task}),
\label{eq:15}
\end{aligned}
\end{equation}
where $z_t$ and $\hat{c}_{t}$ are integrated via weighted summation, enabling their information to interact and be adaptively refined during denoising. $\alpha_{data}$ and $\gamma_{data}$ are weighting factors, with values reported in Sec.~B of the Supplementary Materials. $c_{task}$ denotes the task-level condition: the text prompt for style transfer, and empty for super-resolution and deblurring.

{\bf Discussion.} SDEdit~\cite{meng2021sdedit} synthesizes images by noising a user-provided image to a fixed timestep $t_0$ and then denoising it to reach the data manifold. Our DICT differs in three key aspects: (i) {\em Role of the Condition.} SDEdit treats the condition primarily as a latent initialization (a starting point). It relies on the model’s unconditional prior to ``hallucinate'' missing details based on the noisy layout of the input. In contrast, DICT treats the condition as an active reference source. Instead of just starting from latent initialization, DICT explicitly distills specific high-dimensional cues (e.g., stylistic textures for style transfer or semantic skeletons for restoration) from the condition throughout the denoising guidance process. This ensures the output is grounded in the condition’s actual features rather than just its coarse layout. (ii) {\em Integration Frequency.} SDEdit is a one-shot strategy that only conditions the generative process at the initial timestep $t_0$. This often leads to a ``drift'' where the model loses track of the condition in later stages. In contrast, DICT performs multi-step injection over a subset of the denoising trajectory. This continuous anchoring provides a more direct and stable path toward the target, effectively reducing suboptimal solutions common in fixed-initialization methods. (iii) {\em Adaptive Selectivity.} SDEdit lacks a mechanism to distinguish between relevant and irrelevant information—it attempts to reconstruct the entire noisy input. DICT, however, leverages the noise prediction, i.e., $\epsilon_{ti}=\epsilon_{\theta}(t)$ in Eq.~\ref{eq:15} as a natural filter. By injecting noise and denoising under guidance, DICT adaptively ignores harmful or irrelevant content (e.g., the objects in a style image) and only attends to the task-salient signals.

We then compute the predicted clean latent $z_{0|t}$:
\begin{equation}
\begin{aligned}
z_{0|t} = \frac{z_t - \sqrt{1 - \bar{\alpha}_t} \, \epsilon_\theta(t)}{\sqrt{\bar{\alpha}_t}},
\label{eq:16}
\end{aligned}
\end{equation}
where $z_{0|t}$ is expected to be consistent with the condition $x$. We therefore optimize $z_{0|t}$ to better capture condition-relevant target information:
\begin{equation}
\begin{aligned}
\mathcal{L}_{1}(z_{0|t},x)  = & f_{loss}(D(z_{0|t}), x), \\
z_{0|t} = z_{0|t}  - \eta_1 &\times \nabla_{z_{0|t}}\mathcal{L}_{1}(z_{0|t},x),
\label{eq:17}
\end{aligned}
\end{equation}
where $f_{loss}$ is the task loss function. $\eta_1$ controls the step size of gradient updates to $z_{0|t}$. More details are in Sec.~B and Sec.~C of the Supplementary Materials.

After that, we obtain the $z_{t-1}$ with rich information: 
\begin{equation}
\begin{aligned}
z_{t-1} = \sqrt{\alpha_{t-1}} \, z_{0|t} + \sqrt{1 - \alpha_{t-1} - \sigma_t^2} \, \epsilon_\theta(t) + \sigma_t z,
\label{eq:18}
\end{aligned}
\end{equation}
where $\sigma_t=0$. While Eq.~\ref{eq:17} encourages $z_{0|t}$ to match $x$, it does not guarantee that predictions at later steps are more accurate than those at earlier steps, since errors can accumulate monotonically along the trajectory without an explicit correction mechanism.

{\bf Data Injection as a Non-gradient Score Proxy. } To bypass scalar gradient bottlenecks, our Data Injection (Eq.~\ref{eq:15}) uses latent spatial displacement as a score surrogate. DICT models $\hat{c}_t$ as an empirical Dirac distribution to implicitly maximize high-dimensional likelihood and preserve complex information. Details are in Sec.~A of the Supplementary Materials.

\begin{table}[t]
\centering
\caption{Hyperparameter settings for different tasks in DICT.}
\label{tab:hy}
\setlength{\tabcolsep}{3pt} % 在这里设置列间距，默认通常是 6pt
\scalebox{0.78}{
\begin{tabular}{lccc}
\toprule
Hyperparameters & Style Transfer & Image Super-resolution & Image Deblurring \\ 
\midrule
Base Model   & SD v1.4   & LDM  & LDM  \\
Sampling Steps   & 50 & 200 & 200   \\
Latent Weight ($\alpha_{data}$) & 0.88 & 0.88  & 0.88   \\
Noise Weight ($\gamma_{data}$)  & 0.2  & 0.1   & 0.1 \\
Iterative Updates ($N_{iter1}$, $N_{iter2}$) & 3, 3 & 3, 3  & 3, 4  \\
Initial Update Rate ($\eta_{1}$, $\eta_{2}$)    & 2$^\diamond$, 2$^\diamond$ & [0.08, 0.015, 0.006]$^{\dagger}$, 0.02   & [0.13, 0.015, 0.006]$^*$, 0.02  \\
Data Injection Steps ($T_1$)   & 1-18  & 1-8  & 1-50   \\
Contrastive Trajectory Margin ($\alpha_{margin}$)& 1.0  & 1.0  & 1.0 \\ 
\bottomrule
\multicolumn{4}{l}{\footnotesize $^{\diamond}$ This task employs a varying parameter schedule~\cite{ye2024tfg} for $\eta_1$ and $\eta_2$.}\\
\multicolumn{4}{l}{\footnotesize $^{\dagger}$ This task employs a three-stage schedule for $\eta_1$: steps 1--130, 131--160, and 161--200.}\\
\multicolumn{4}{l}{\footnotesize $^{*}$ This task employs a three-stage schedule for $\eta_1$: steps 1--130, 131--150, and 151--200.}\\

\end{tabular}
}
\end{table}

{\bf Contrastive Trajectory Refinement. } In conditional image generation tasks, relying on a scalar-level loss $\mathcal{L}_1$ has fundamental limitations. Such a loss provides only a static target at a single timestep and does not reflect the temporal structure of the reverse-diffusion trajectory. In practice, this leads to locally optimal updates at each discrete timestep, rather than a coherent refinement trajectory, i.e., $z_{0|t-1}$ is not guaranteed to be a principled improvement over $z_{0|t}$. Without an explicit mechanism to correct earlier deviations, small inaccuracies introduced at early steps can propagate and compound, biasing the sampling trajectory and degrading the final result. To address these issues, we leverage the reverse-diffusion trajectory structure (Fig.~\ref{fig:1} and Fig.~\ref{fig:2}(b)) and impose a Contrastive Trajectory Refinement principle: each intermediate prediction should be closer to the condition target than the previous one. This trajectory-aware design supports incremental correction and favors more reliable denoising paths, thereby mitigating error accumulation via a contrastive trajectory objective: 
\begin{equation}
\begin{aligned}
\mathcal{L}_{2} = \max \big( \mathcal{L}_{1}(z_{0|t-1},y) &- \mathcal{L}_{1}(z_{0|t},y) + \alpha_{margin}, 0 \big),\\
z_{0|t-1} = z_{0|t-1} &- \eta_2 \times \nabla_{z_{0|t-1}}\mathcal{L}_{2},
\label{eq:19}
\end{aligned}
\end{equation}
where $\alpha_{margin}$ enforces a margin, requiring the negative sample $z_{0|t}$ to remain at least a fixed distance away from the positive sample $z_{0|t-1}$. $\eta_2$ controls the step size of the gradient-based update applied to $z_{0|t-1}$. All parameter settings are provided in Sec.~C of the Supplementary Materials. With this strengthened trajectory prior, $z_{0|t-1}$ is encouraged to be closer to the target than $z_{0|t}$.

Ultimately, we obtain the noisy latent $z_{t-1}$ which integrates more accurate and higher-quality condition information:
\begin{equation}
\begin{aligned}
z_{t-1} = \sqrt{\bar{\alpha}_{t-1}} z_{0|t-1} + \sqrt{1 - \bar{\alpha}_{t-1}} \epsilon.
\label{eq:20}
\end{aligned}
\end{equation}

Iterating this reverse process yields the final output: 
\begin{equation}
\begin{aligned}
x_0 = D(z_{0}),
\label{eq:21}
\end{aligned}
\end{equation}
where $x_0=y$ denotes the final output, i.e., the target sample generated under the provided condition.

{\bf Contrastive Trajectory Refinement as a Discrete Lyapunov Stabilizer.} Solver truncation introduces Local Truncation Errors (LTE, $\mathcal{O}(\Delta t^{k+1})$) causing tangential drift. Our $\mathcal{L}_2$ acts as a discrete Lyapunov constraint enforcing monotonic energy decay ($\Delta \mathcal{L}_1 \le -\alpha_{margin}$). Functioning as a soft Manifold Projection Operator, it generates a normal force to mitigate orthogonal LTE, promoting coherence. Details are in Sec. A of the Supplementary Materials.

\begin{figure*}[t!]
\centering
\subfloat[\label{fig:4a} Qualitative comparisons of text-to-image style transfer task. ]{
\resizebox{1\textwidth}{!}{
\setlength{\tabcolsep}{0.05cm} % 调整列间距
\renewcommand{\arraystretch}{0.5}  % 调整行距
\begin{tabular}{ccccccccccccc}
 Text & Style & {\bf DICT} & StyleShot & StyleStudio & StyleCrafter & DEADiff & InstantStyle & StyleAlign & CSGO & StyleDrop & TFG & FreeDom \\
\includegraphics[width=0.14\linewidth]{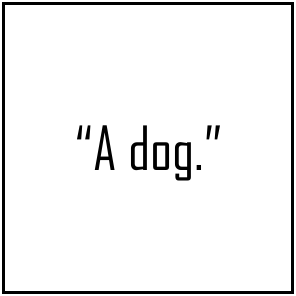} & \includegraphics[width=0.14\linewidth]{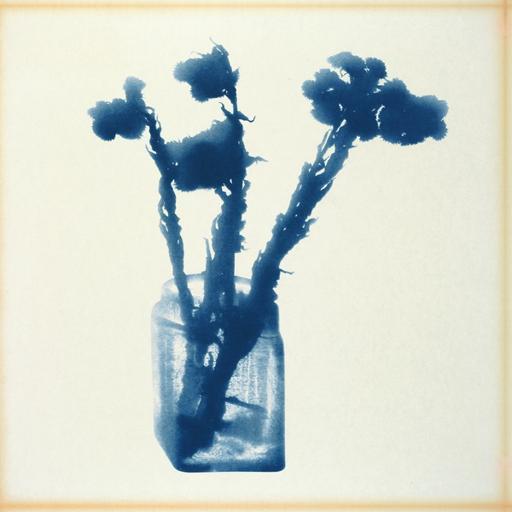}  & \includegraphics[width=0.14\linewidth]{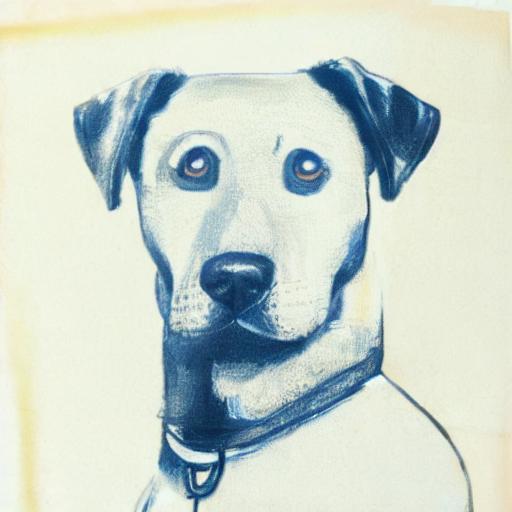} &
\includegraphics[width=0.14\linewidth]{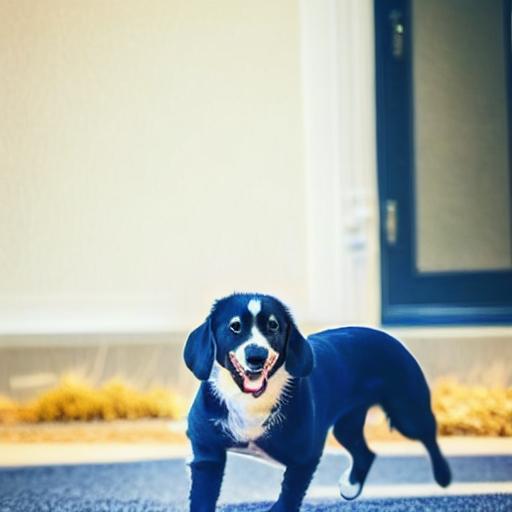} & \includegraphics[width=0.14\linewidth]{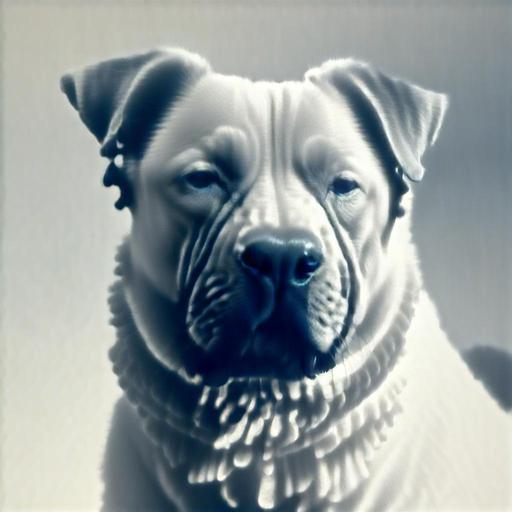} &
\includegraphics[width=0.14\linewidth]{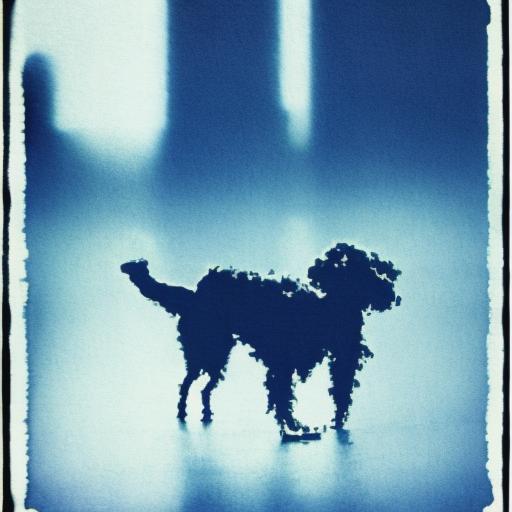} &  \includegraphics[width=0.14\linewidth]{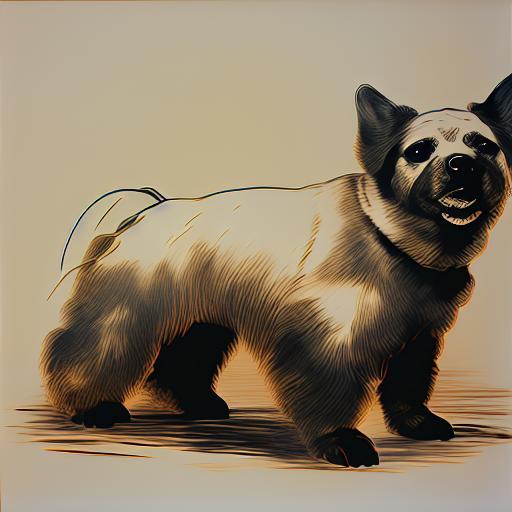} &  \includegraphics[width=0.14\linewidth]{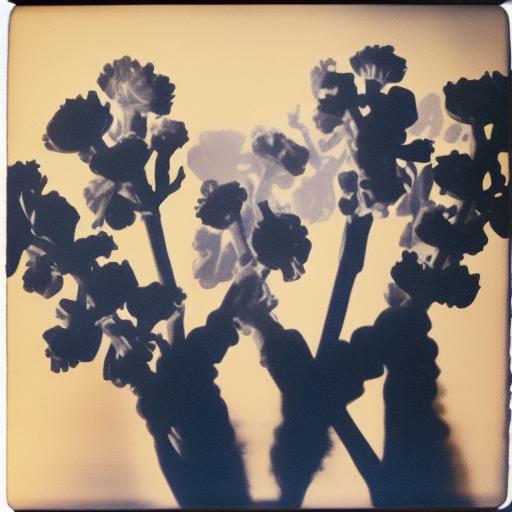} &  \includegraphics[width=0.14\linewidth]{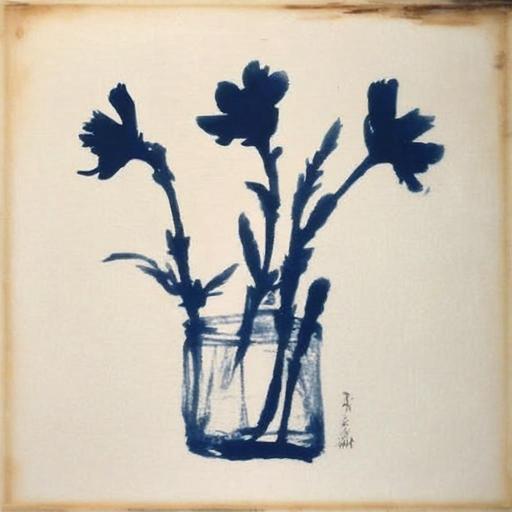}& \includegraphics[width=0.14\linewidth]{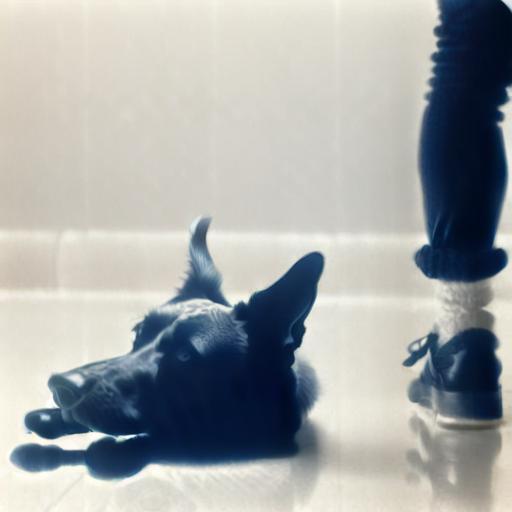} &  
\includegraphics[width=0.14\linewidth]{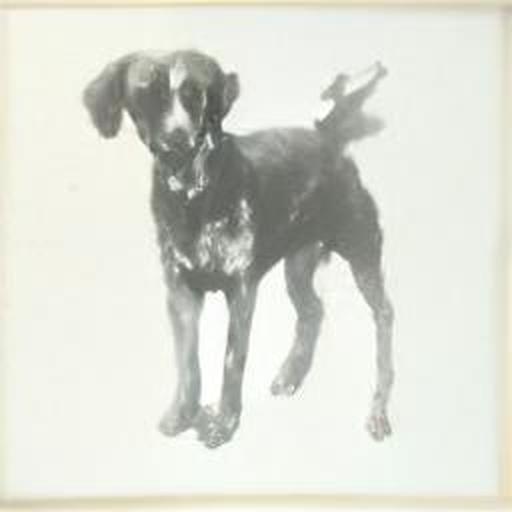} & \includegraphics[width=0.14\linewidth]{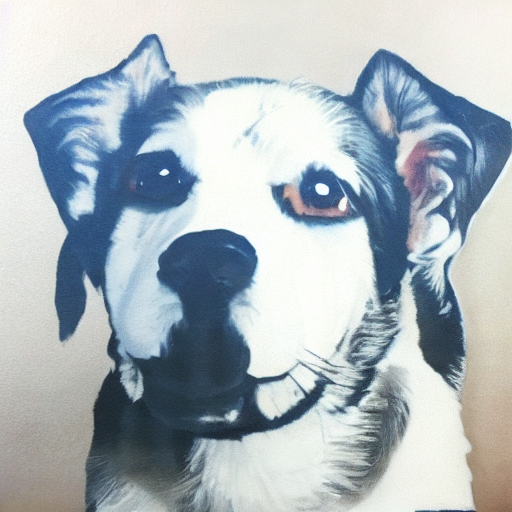} &  
\includegraphics[width=0.14\linewidth]{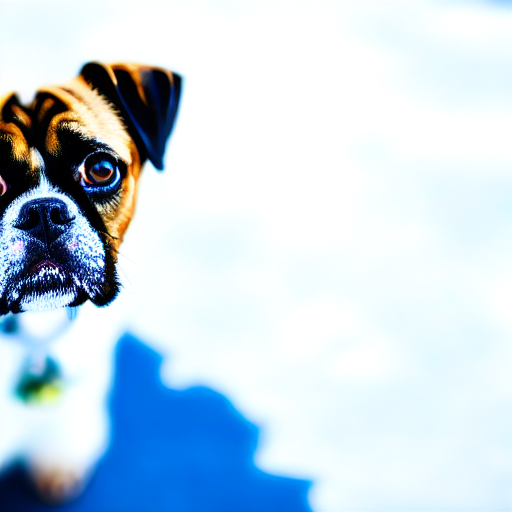} \\
\includegraphics[width=0.14\linewidth]{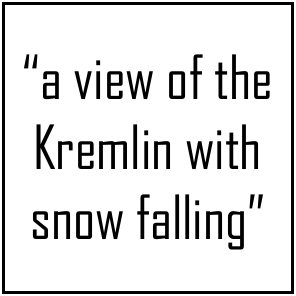} & \includegraphics[width=0.14\linewidth]{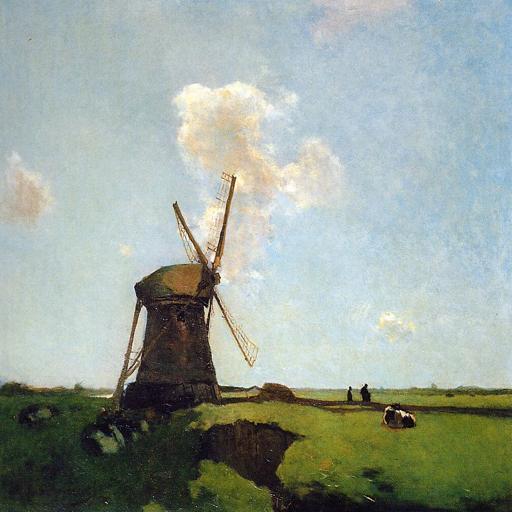}  & \includegraphics[width=0.14\linewidth]{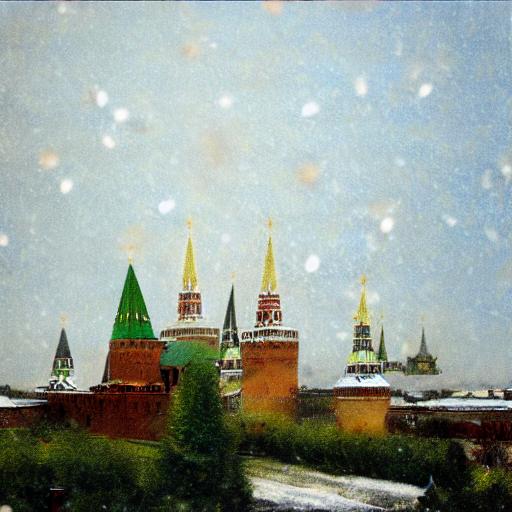} &
\includegraphics[width=0.14\linewidth]{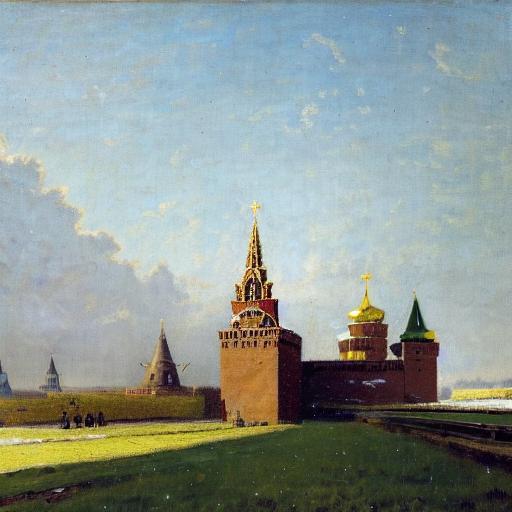} & \includegraphics[width=0.14\linewidth]{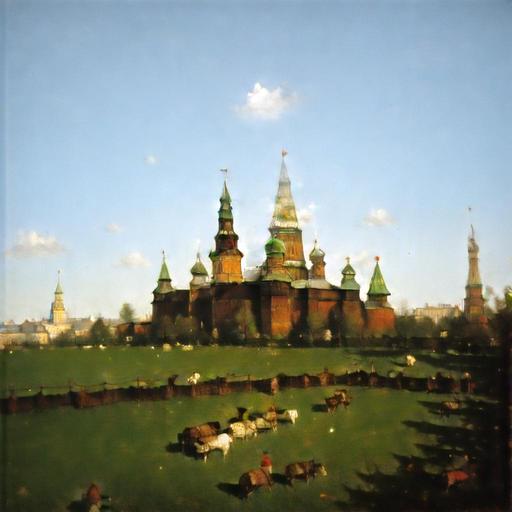} &
\includegraphics[width=0.14\linewidth]{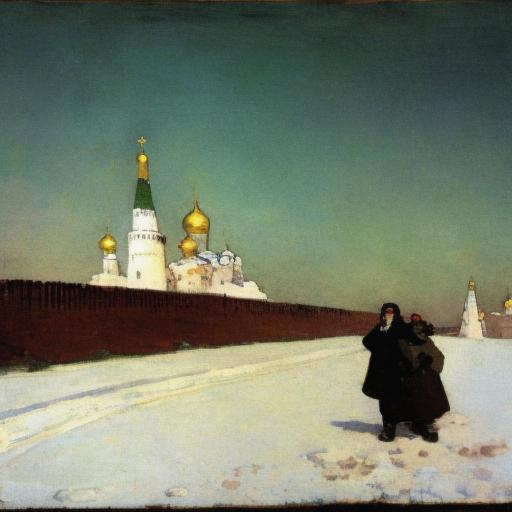} &  \includegraphics[width=0.14\linewidth]{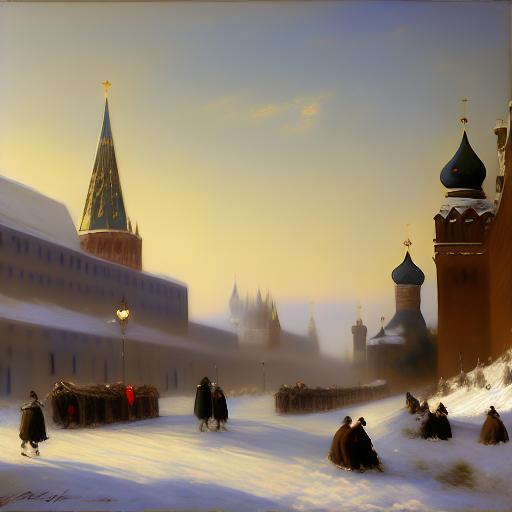} &  \includegraphics[width=0.14\linewidth]{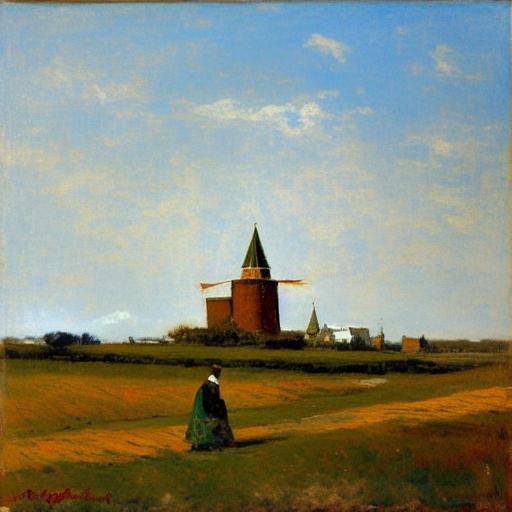} &  \includegraphics[width=0.14\linewidth]{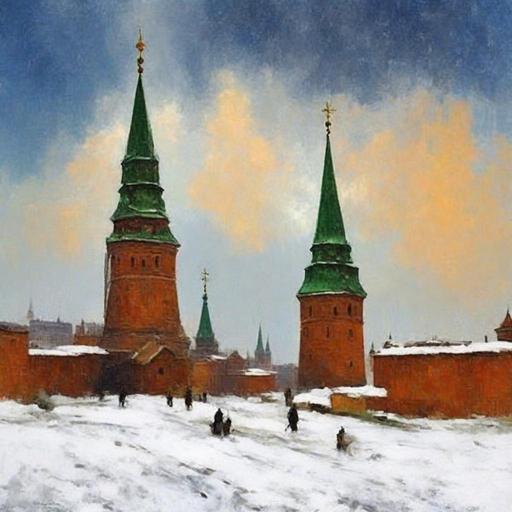}& \includegraphics[width=0.14\linewidth]{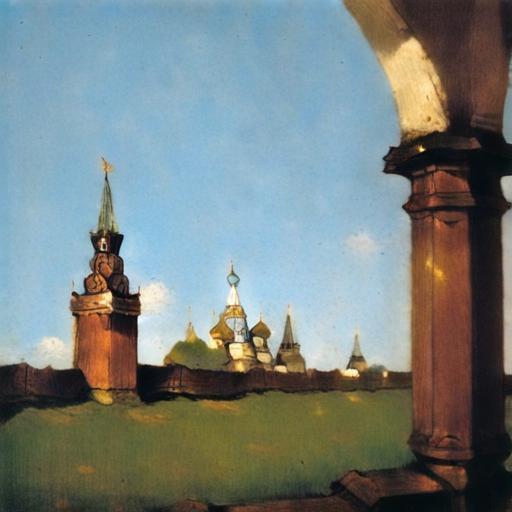} &  
\includegraphics[width=0.14\linewidth]{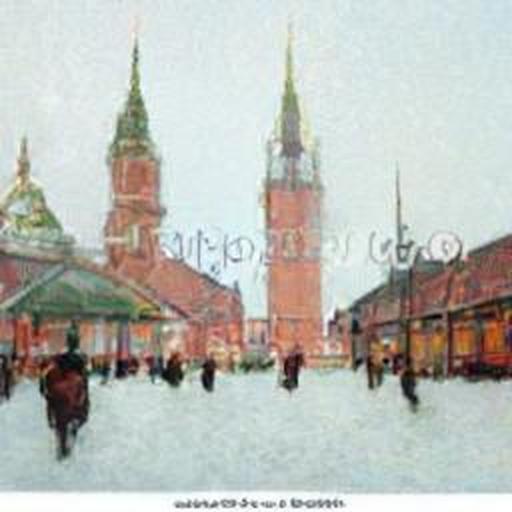}& \includegraphics[width=0.14\linewidth]{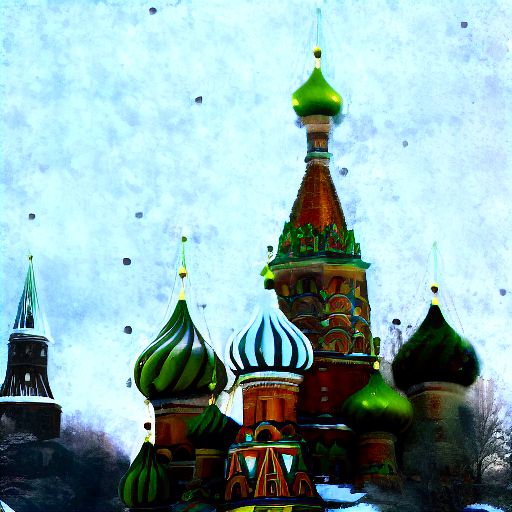} &  
\includegraphics[width=0.14\linewidth]{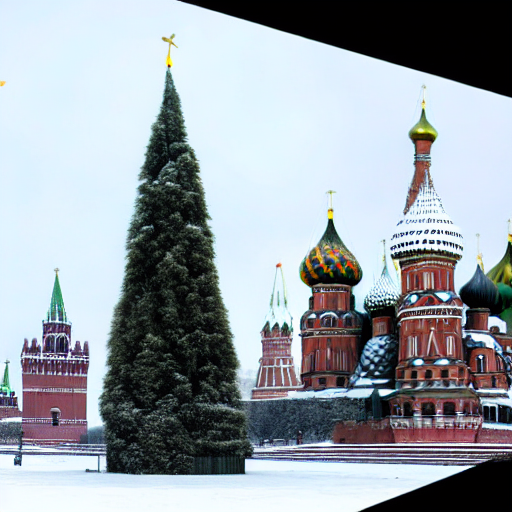}\\
\includegraphics[width=0.14\linewidth]{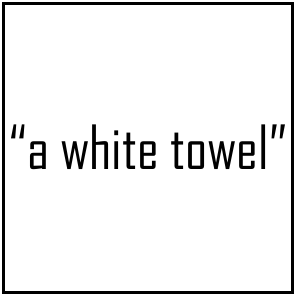} & \includegraphics[width=0.14\linewidth]{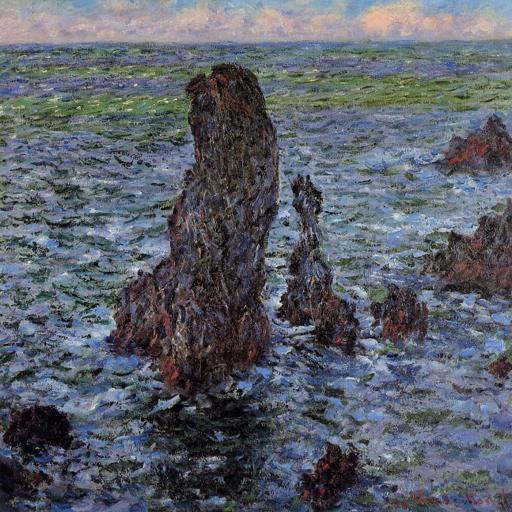}  & \includegraphics[width=0.14\linewidth]{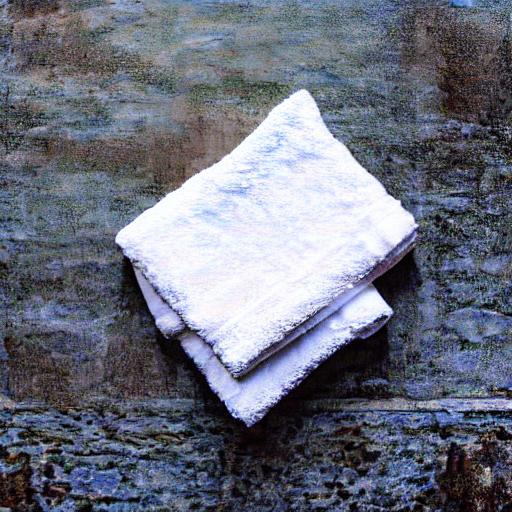} &
\includegraphics[width=0.14\linewidth]{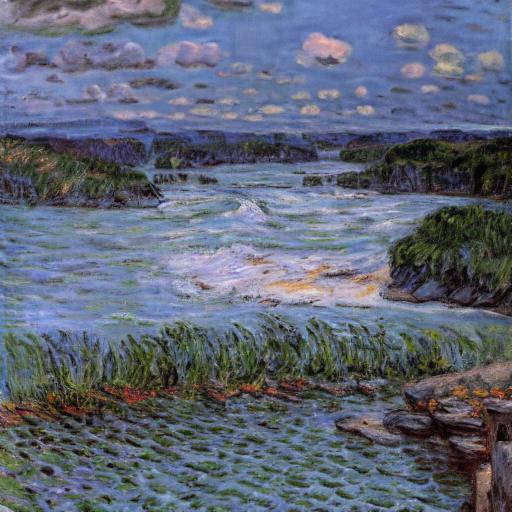} & \includegraphics[width=0.14\linewidth]{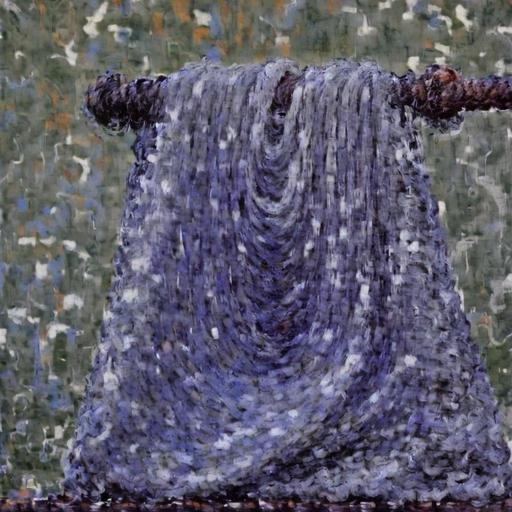} &
\includegraphics[width=0.14\linewidth]{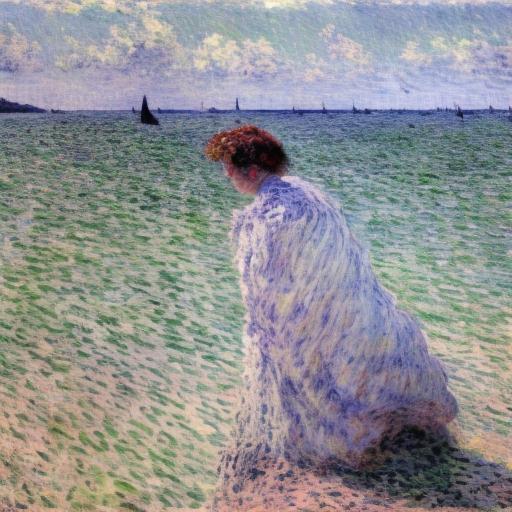} &  \includegraphics[width=0.14\linewidth]{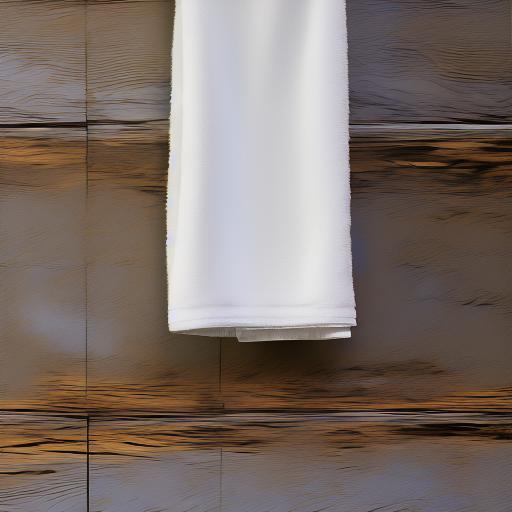} &  \includegraphics[width=0.14\linewidth]{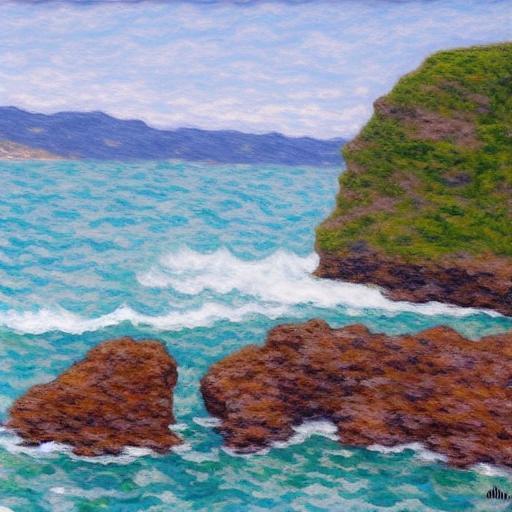} &  \includegraphics[width=0.14\linewidth]{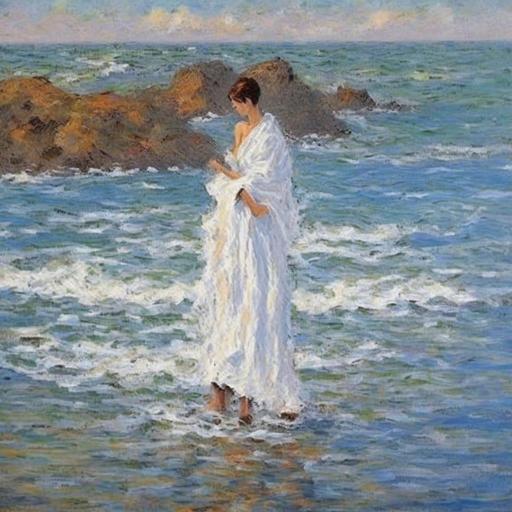}& \includegraphics[width=0.14\linewidth]{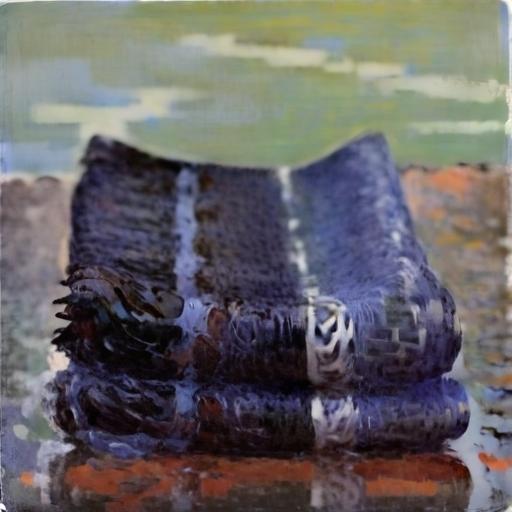} &  
\includegraphics[width=0.14\linewidth]{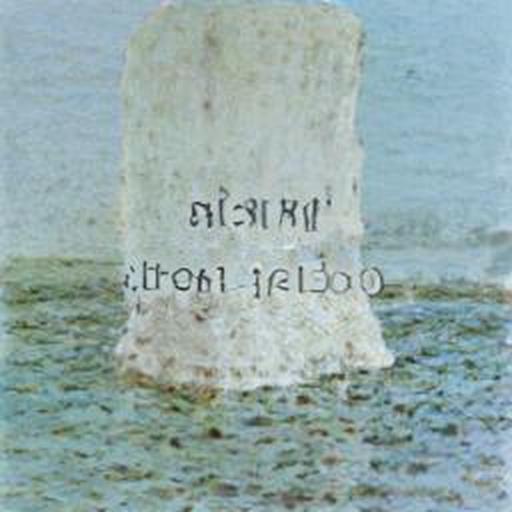}& \includegraphics[width=0.14\linewidth]{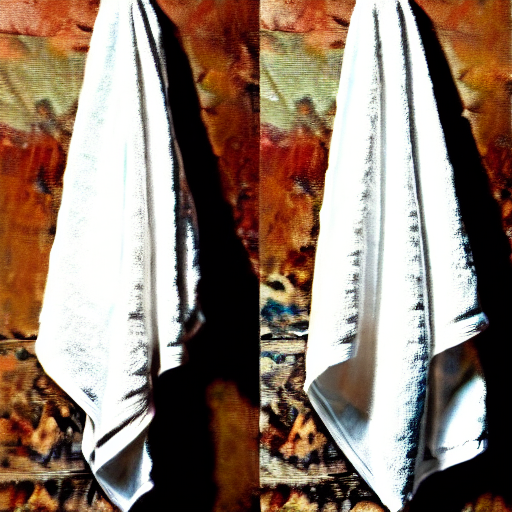} &  
\includegraphics[width=0.14\linewidth]{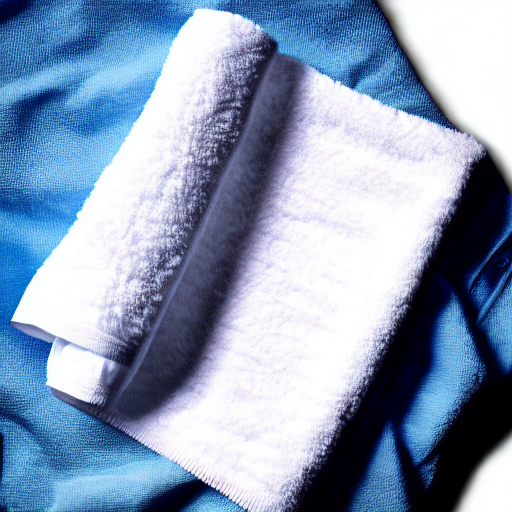}\\
\includegraphics[width=0.14\linewidth]{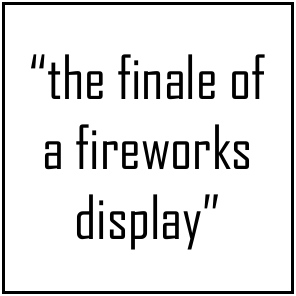} & \includegraphics[width=0.14\linewidth]{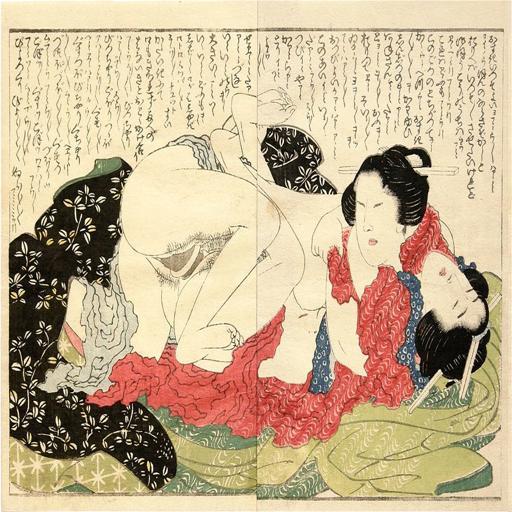}  & \includegraphics[width=0.14\linewidth]{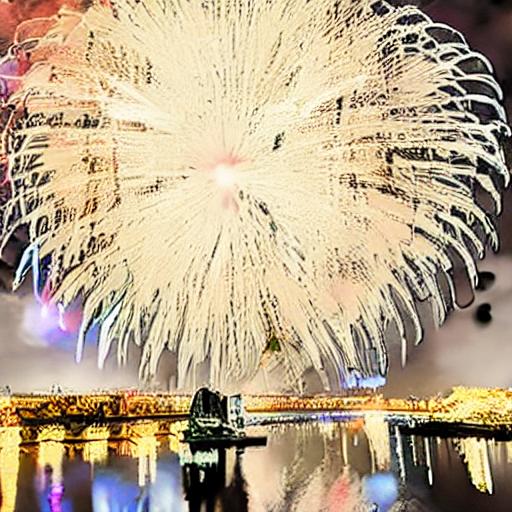} &
\includegraphics[width=0.14\linewidth]{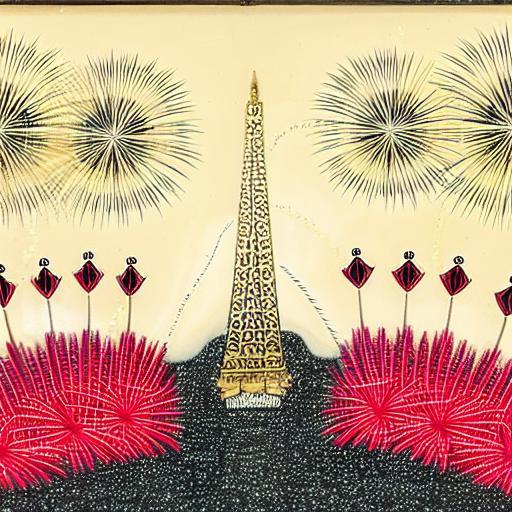} & \includegraphics[width=0.14\linewidth]{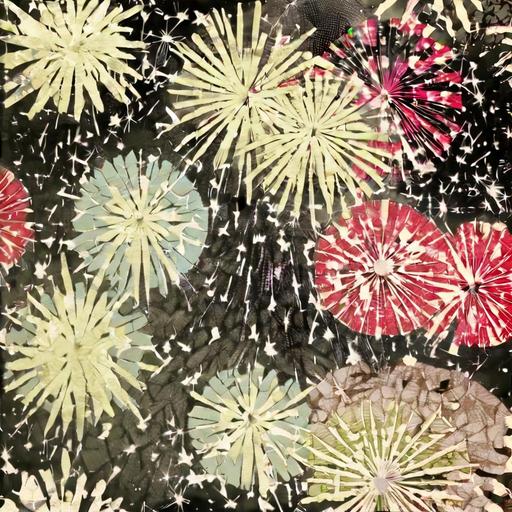} &
\includegraphics[width=0.14\linewidth]{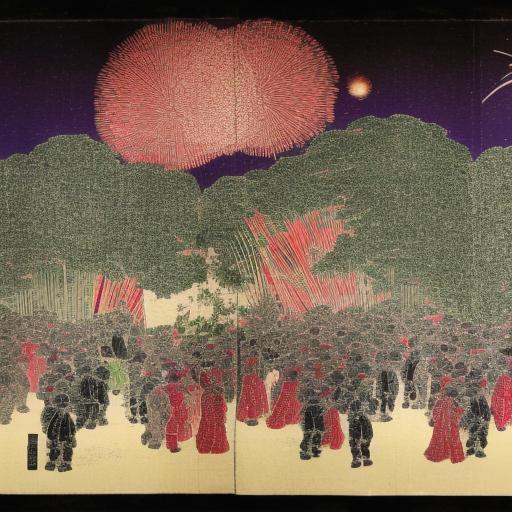} &  \includegraphics[width=0.14\linewidth]{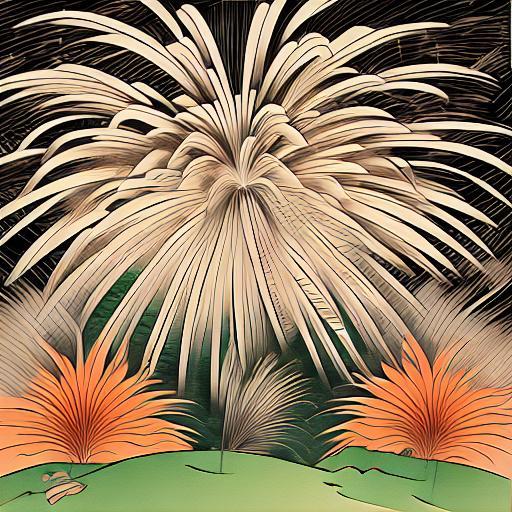} &  \includegraphics[width=0.14\linewidth]{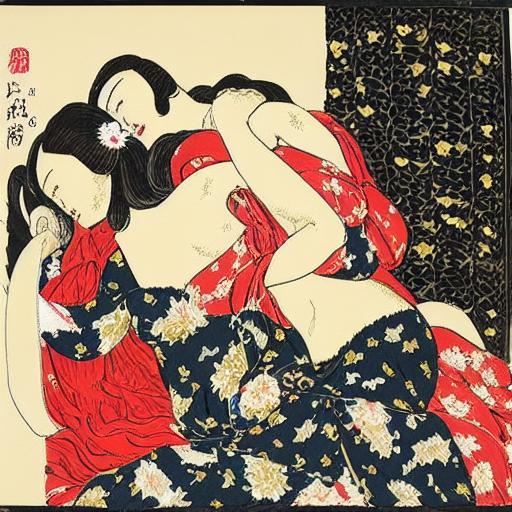} &  \includegraphics[width=0.14\linewidth]{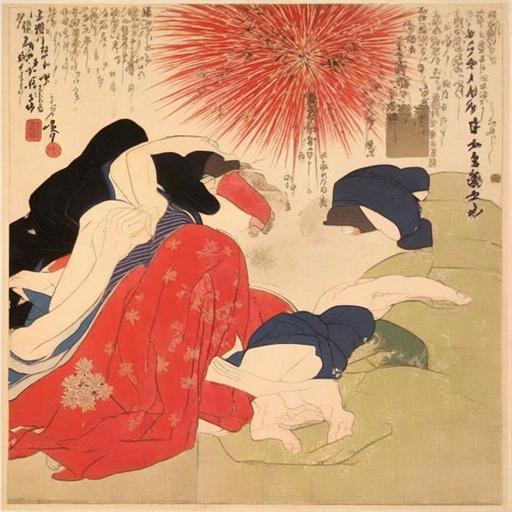}& \includegraphics[width=0.14\linewidth]{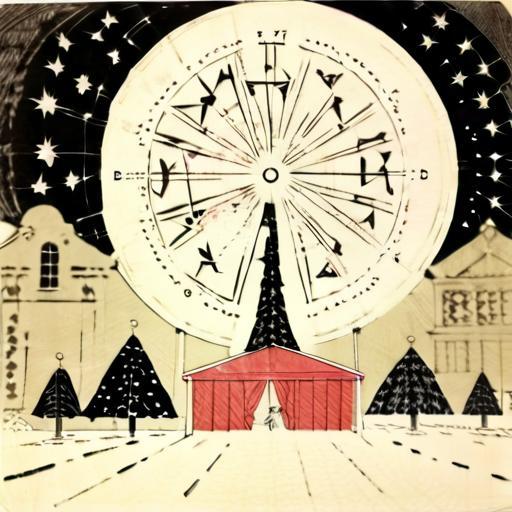} &  
\includegraphics[width=0.14\linewidth]{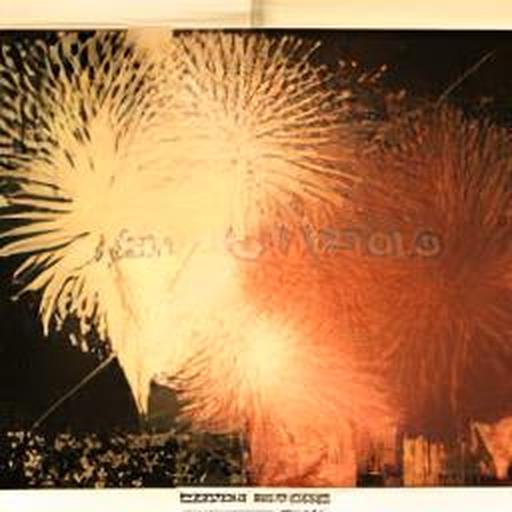} & \includegraphics[width=0.14\linewidth]{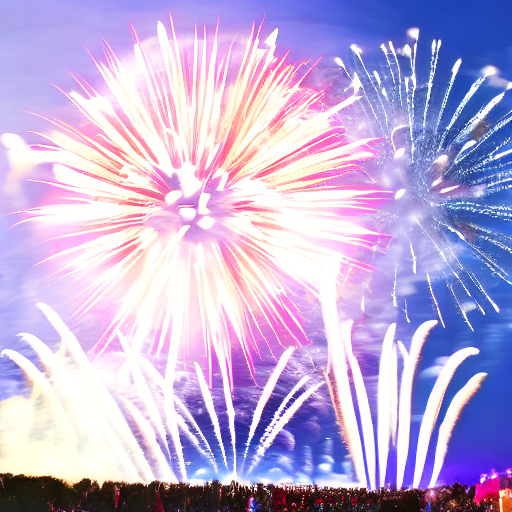} &  
\includegraphics[width=0.14\linewidth]{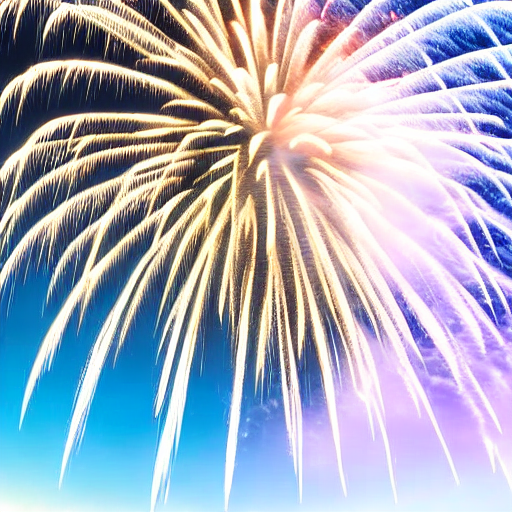}
\end{tabular}
}}
\vspace{5pt}
\subfloat[\label{fig:4b} Qualitative comparisons of image super-resolution task.  ]{
\resizebox{1\textwidth}{!}{
% \scriptsize
\setlength{\tabcolsep}{0.05cm} % 调整列间距
\renewcommand{\arraystretch}{0.5}  % 调整行距
\begin{tabular}{ccccccccccccc}
 Condition & Target & {\bf DICT} & InvSR & PSLD  & FPS-SMC$^*$ & SITCOM$^*$ &  DMAP & FlowDPS & FlowChef & DOC$^*$ & TFG$^*$ & FreeDom$^*$\\
\includegraphics[width=0.14\linewidth]{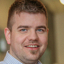} & \includegraphics[width=0.14\linewidth]{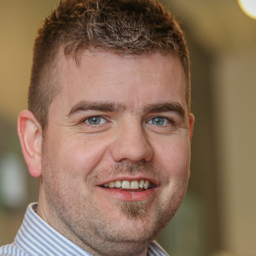}  & \includegraphics[width=0.14\linewidth]{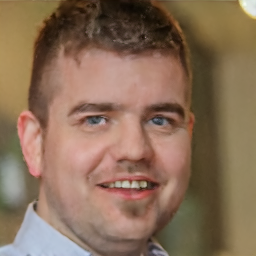} &
\includegraphics[width=0.14\linewidth]{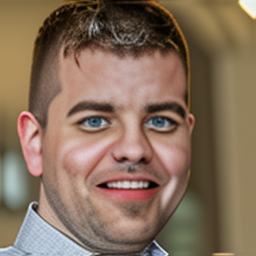} & \includegraphics[width=0.14\linewidth]{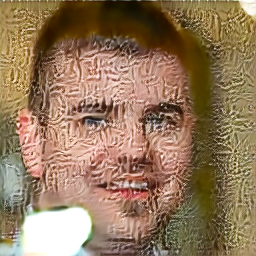} &
\includegraphics[width=0.14\linewidth]{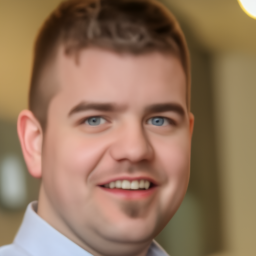} &  \includegraphics[width=0.14\linewidth]{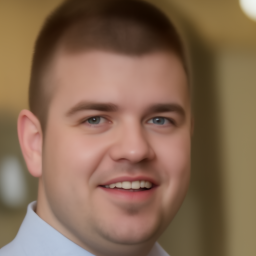} &  \includegraphics[width=0.14\linewidth]{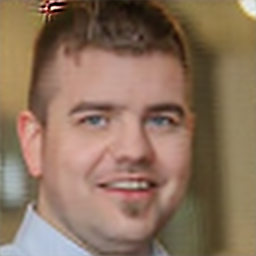} &  \includegraphics[width=0.14\linewidth]{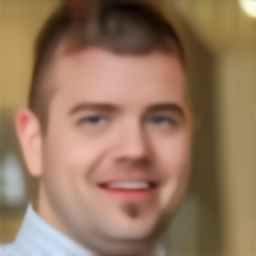}& \includegraphics[width=0.14\linewidth]{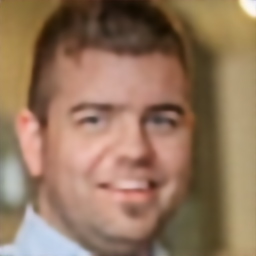} &  
\includegraphics[width=0.14\linewidth]{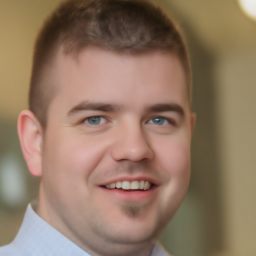} & \includegraphics[width=0.14\linewidth]{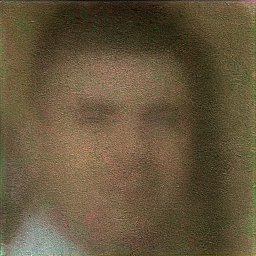} &  
\includegraphics[width=0.14\linewidth]{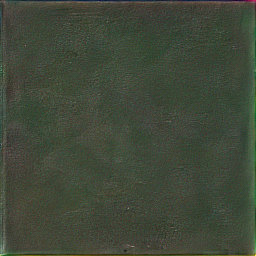}\\
\includegraphics[width=0.14\linewidth]{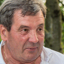} & \includegraphics[width=0.14\linewidth]{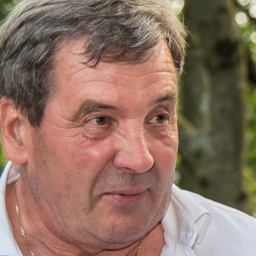}  & \includegraphics[width=0.14\linewidth]{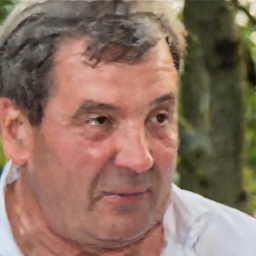} &
\includegraphics[width=0.14\linewidth]{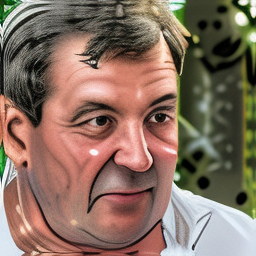} & \includegraphics[width=0.14\linewidth]{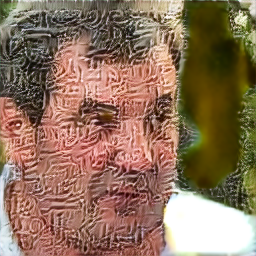} &
\includegraphics[width=0.14\linewidth]{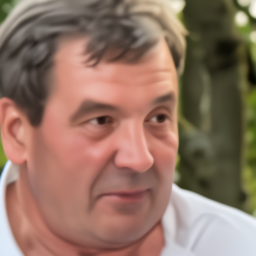} &  \includegraphics[width=0.14\linewidth]{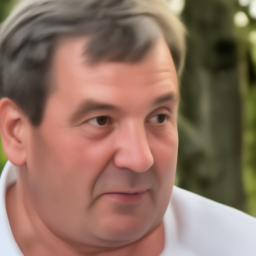} &  \includegraphics[width=0.14\linewidth]{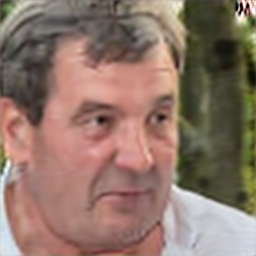} &  \includegraphics[width=0.14\linewidth]{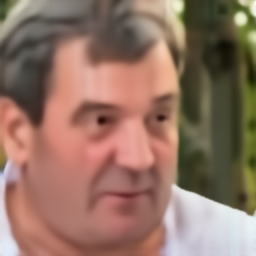}& \includegraphics[width=0.14\linewidth]{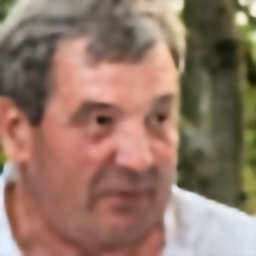} &  
\includegraphics[width=0.14\linewidth]{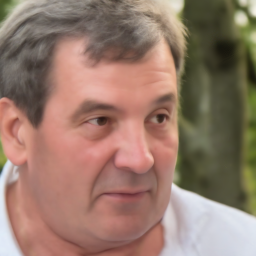} & \includegraphics[width=0.14\linewidth]{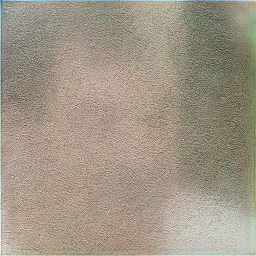} &  
\includegraphics[width=0.14\linewidth]{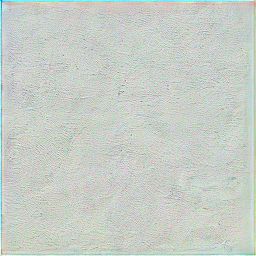}\\
\includegraphics[width=0.14\linewidth]{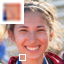} & \includegraphics[width=0.14\linewidth]{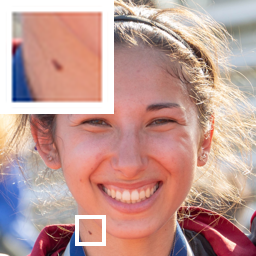}  & \includegraphics[width=0.14\linewidth]{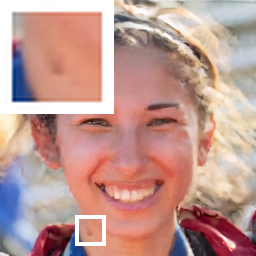} &
\includegraphics[width=0.14\linewidth]{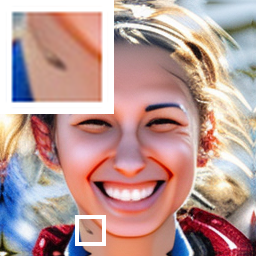} & \includegraphics[width=0.14\linewidth]{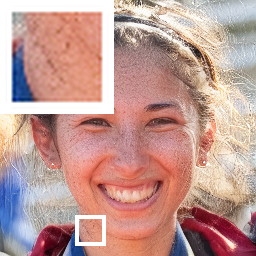} &
\includegraphics[width=0.14\linewidth]{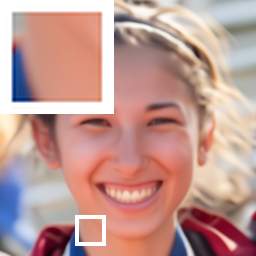} &  \includegraphics[width=0.14\linewidth]{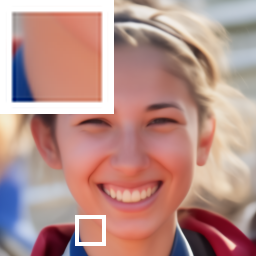} &  \includegraphics[width=0.14\linewidth]{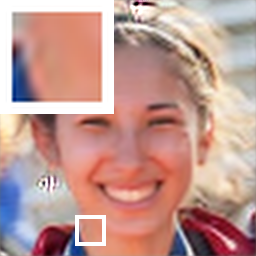} &  \includegraphics[width=0.14\linewidth]{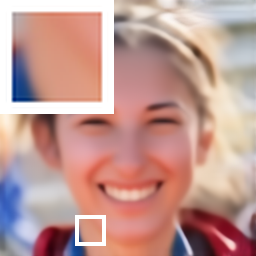}& \includegraphics[width=0.14\linewidth]{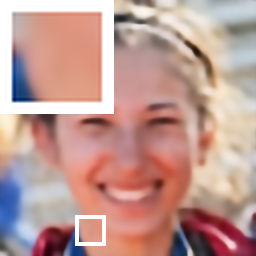} &  
\includegraphics[width=0.14\linewidth]{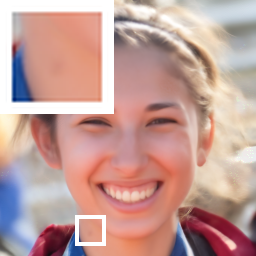}& \includegraphics[width=0.14\linewidth]{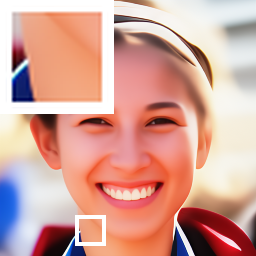} &  
\includegraphics[width=0.14\linewidth]{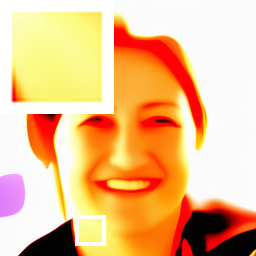}\\
\includegraphics[width=0.14\linewidth]{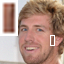} & \includegraphics[width=0.14\linewidth]{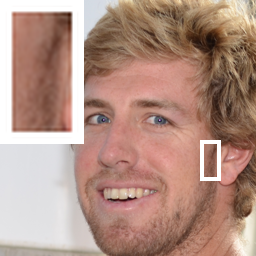}  & \includegraphics[width=0.14\linewidth]{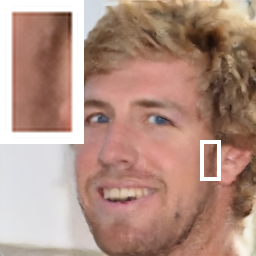} &
\includegraphics[width=0.14\linewidth]{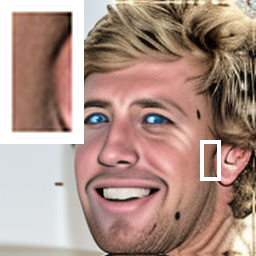} & \includegraphics[width=0.14\linewidth]{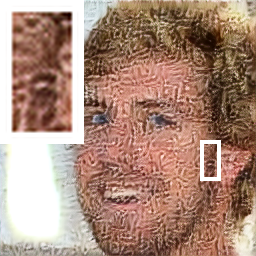} &
\includegraphics[width=0.14\linewidth]{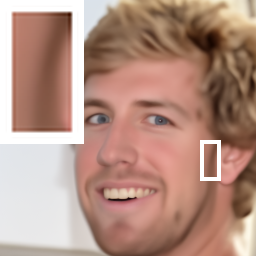} &  \includegraphics[width=0.14\linewidth]{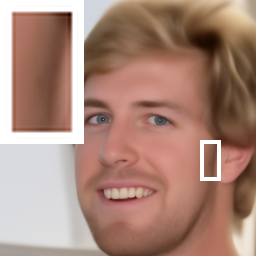} &  \includegraphics[width=0.14\linewidth]{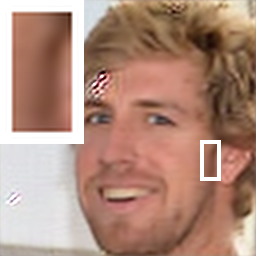} &  \includegraphics[width=0.14\linewidth]{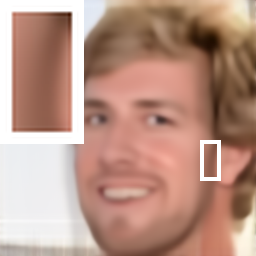}& \includegraphics[width=0.14\linewidth]{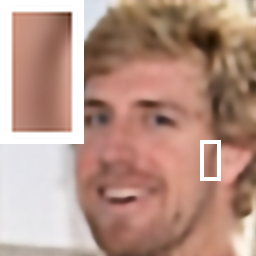} &  
\includegraphics[width=0.14\linewidth]{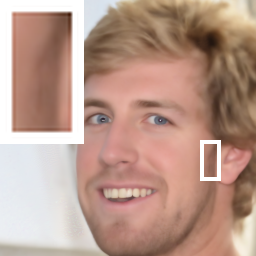}& \includegraphics[width=0.14\linewidth]{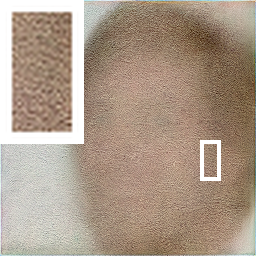} &  
\includegraphics[width=0.14\linewidth]{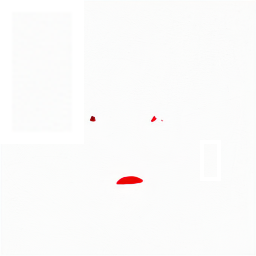}
\end{tabular}
}}
\vspace{5pt}
\subfloat[\label{fig:4c} Qualitative comparisons of image deblurring task. ]{
\resizebox{1\textwidth}{!}{
% \scriptsize
\setlength{\tabcolsep}{0.05cm} % 调整列间距
\renewcommand{\arraystretch}{0.5}  % 调整行距
\begin{tabular}{ccccccccccccc}
 Condition & Target & {\bf DICT} & DCDP* & PSLD & FPS-SMC$^*$ & SITCOM$^*$ & DMAP & FlowDPS & FlowChef & DOC$^*$ & TFG$^*$ & FreeDom$^*$\\
\includegraphics[width=0.14\linewidth]{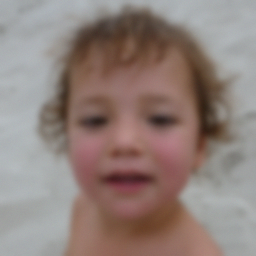} & \includegraphics[width=0.14\linewidth]{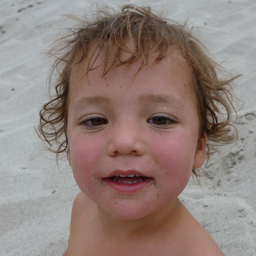}  & \includegraphics[width=0.14\linewidth]{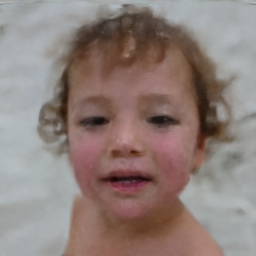} &
\includegraphics[width=0.14\linewidth]{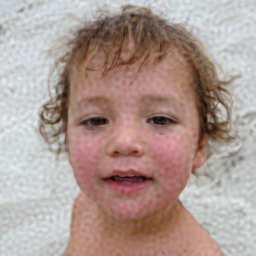} & \includegraphics[width=0.14\linewidth]{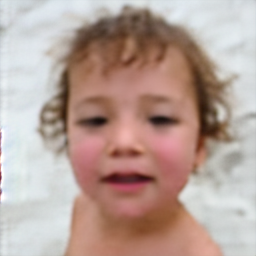} &
\includegraphics[width=0.14\linewidth]{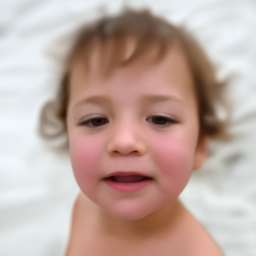} &  \includegraphics[width=0.14\linewidth]{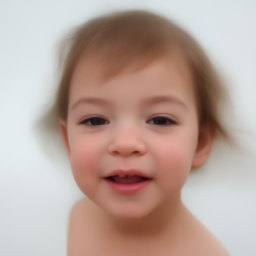} &  \includegraphics[width=0.14\linewidth]{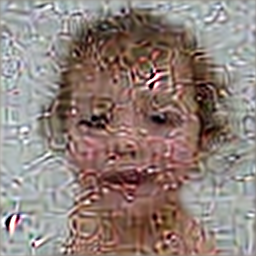} &  \includegraphics[width=0.14\linewidth]{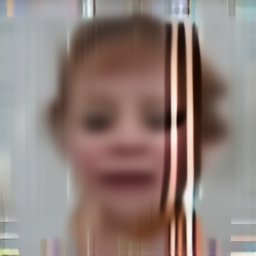}& \includegraphics[width=0.14\linewidth]{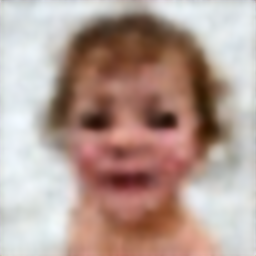} &  
\includegraphics[width=0.14\linewidth]{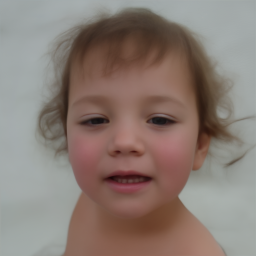} & \includegraphics[width=0.14\linewidth]{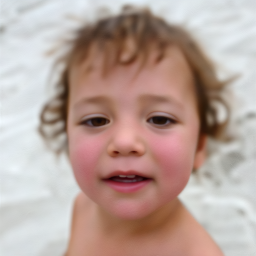} &  
\includegraphics[width=0.14\linewidth]{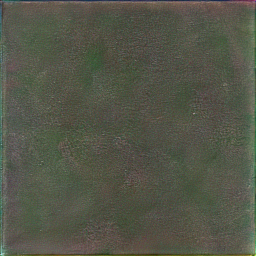}\\
\includegraphics[width=0.14\linewidth]{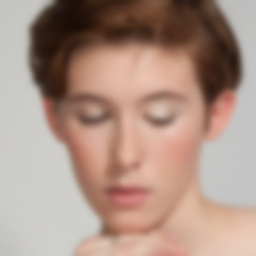} & \includegraphics[width=0.14\linewidth]{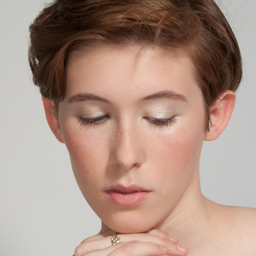}  & \includegraphics[width=0.14\linewidth]{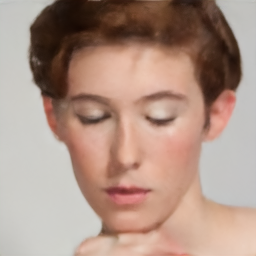} &
\includegraphics[width=0.14\linewidth]{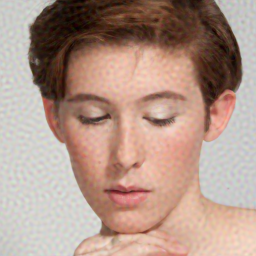} & \includegraphics[width=0.14\linewidth]{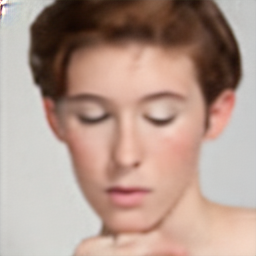} &
\includegraphics[width=0.14\linewidth]{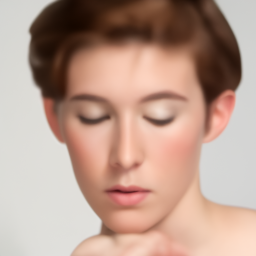} &  \includegraphics[width=0.14\linewidth]{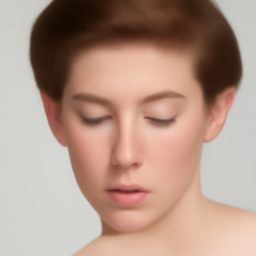} &  \includegraphics[width=0.14\linewidth]{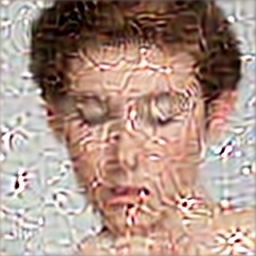} &  \includegraphics[width=0.14\linewidth]{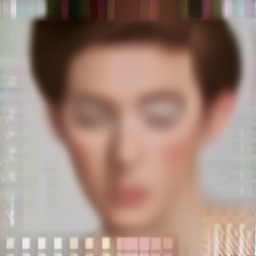}& \includegraphics[width=0.14\linewidth]{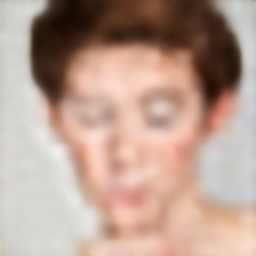} &  
\includegraphics[width=0.14\linewidth]{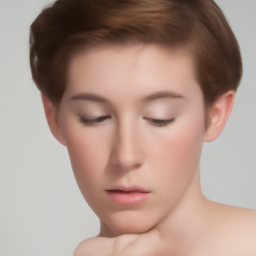} & \includegraphics[width=0.14\linewidth]{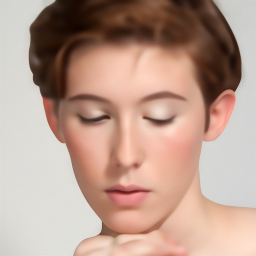} &  
\includegraphics[width=0.14\linewidth]{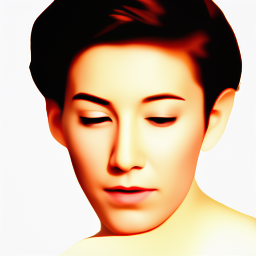}\\
\includegraphics[width=0.14\linewidth]{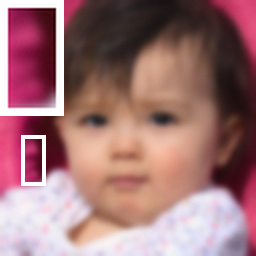} & \includegraphics[width=0.14\linewidth]{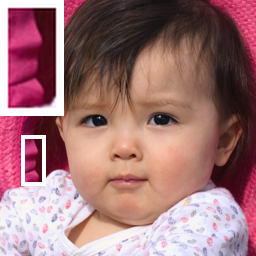}  & \includegraphics[width=0.14\linewidth]{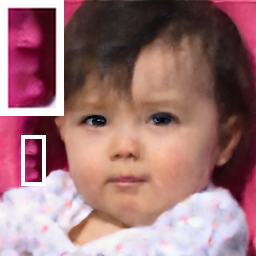} &
\includegraphics[width=0.14\linewidth]{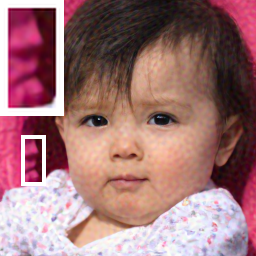} & \includegraphics[width=0.14\linewidth]{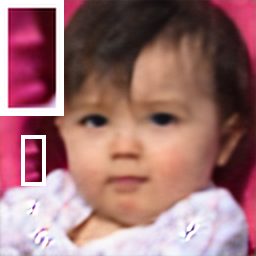} &
\includegraphics[width=0.14\linewidth]{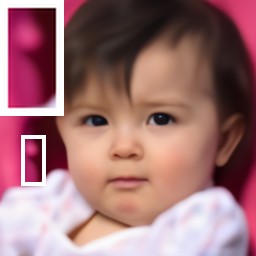} &  \includegraphics[width=0.14\linewidth]{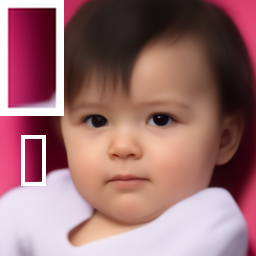} &  \includegraphics[width=0.14\linewidth]{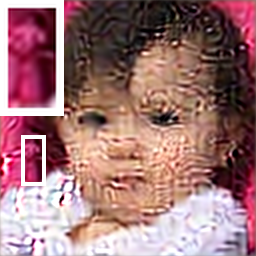} &  \includegraphics[width=0.14\linewidth]{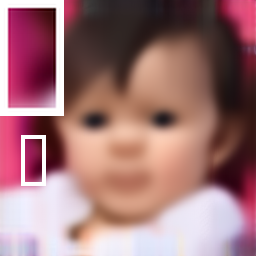}& \includegraphics[width=0.14\linewidth]{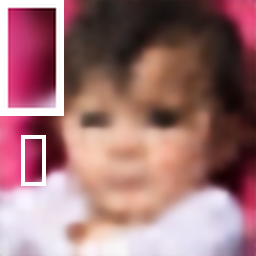} &  
\includegraphics[width=0.14\linewidth]{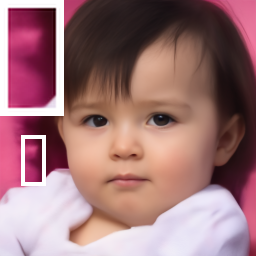} & \includegraphics[width=0.14\linewidth]{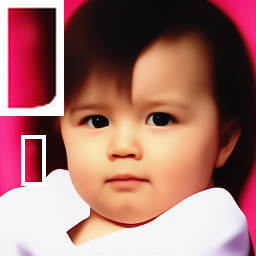} &  
\includegraphics[width=0.14\linewidth]{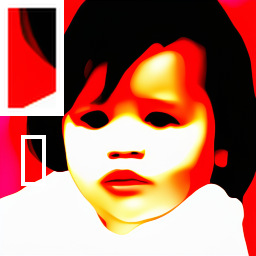}\\
\includegraphics[width=0.14\linewidth]{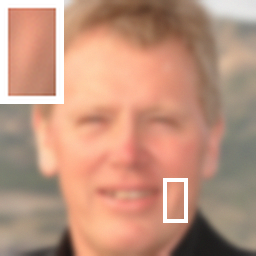} & \includegraphics[width=0.14\linewidth]{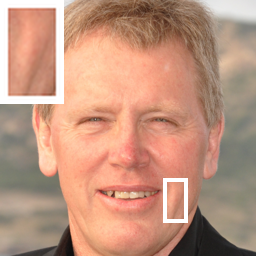}  & \includegraphics[width=0.14\linewidth]{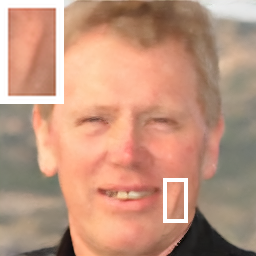} &
\includegraphics[width=0.14\linewidth]{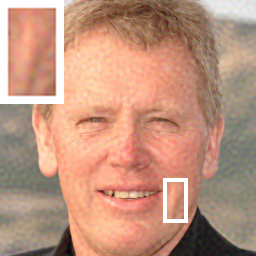} & \includegraphics[width=0.14\linewidth]{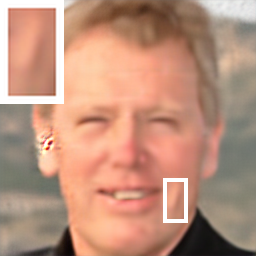} &
\includegraphics[width=0.14\linewidth]{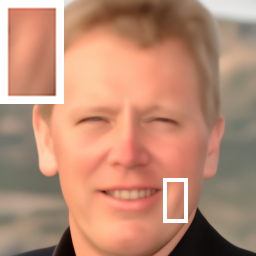} &  \includegraphics[width=0.14\linewidth]{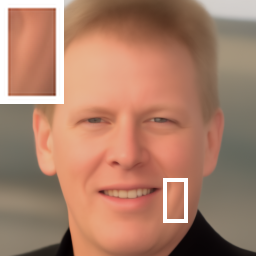} &  \includegraphics[width=0.14\linewidth]{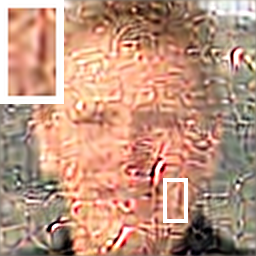} &  \includegraphics[width=0.14\linewidth]{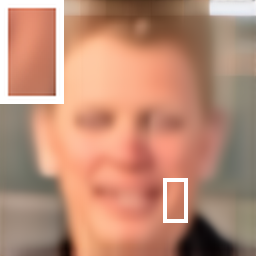}& \includegraphics[width=0.14\linewidth]{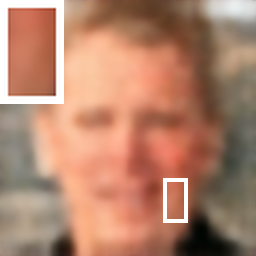} &  
\includegraphics[width=0.14\linewidth]{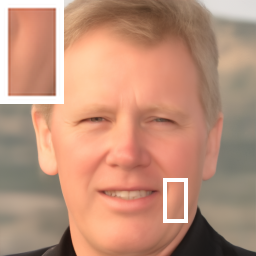} & \includegraphics[width=0.14\linewidth]{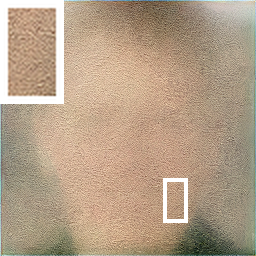} &  
\includegraphics[width=0.14\linewidth]{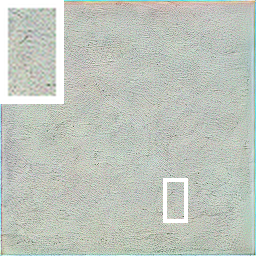}
\end{tabular}
}}

\caption{Qualitative comparisons. An asterisk ($*$) denotes models in the pixel space; those unmarked operate in the latent space. }
\label{fig:4}
\end{figure*}

\section{Experiments}
\subsection{Experiment Settings}

We study conditional image generation under three tasks: style transfer, super-resolution, and deblurring. For text-driven style transfer, we evaluate on 40,000 stylized images at $512\times512$ resolution, constructed by pairing 200 text prompts with 200 style images randomly sampled from WikiArt~\cite{phillips2011wiki}. For super-resolution and deblurring, we evaluate on 1,000 randomly selected FFHQ images~\cite{karras2019style}. In super-resolution, we downsample each image by a factor of $4\times$ and add Gaussian noise with standard deviation 0.01, and generate outputs at $256\times256$ resolution. In deblurring, we apply a Gaussian blur kernel of size 61 with standard deviation 3.0, add Gaussian noise with standard deviation 0.01, and generate outputs at $256\times256$ resolution. Hyperparameter settings for different tasks of DICT are in Tab.~\ref{tab:hy}. Detailed settings are in Sec.~B of the Supplementary Materials.

\newcommand{\mm}{0.9}
\newcommand{\mmm}{1.8}
\newcommand{\rrr}{0.45}

\begin{table*}[t!]
\caption{Quantitative comparisons of the conditional image generation. The best results are in \textbf{bold}. The second best results are \underline{underlined}.}
\centering
\scalebox{0.66}{
\begin{tabular}{c|ccccccccccc}

\multicolumn{12}{c}{\large (a) Quantitative comparisons of text-to-image style transfer task.  }\\
\multicolumn{12}{c}{}\\
\hline

& {\bf DICT} & StyleShot & StyleStudio & StyleCrafter & DEADiff & InstantStyle & StyleAlign & CSGO & StyleDrop & TFG & FreeDom\\
\hline
Text Score $\uparrow$ & \underline{0.2952} & 0.2740 & 0.2852 & 0.2582 & 0.2907 & 0.1920 & 0.2553 & 0.2688 & 0.2780 & {\bf 0.3092} & 0.2933\\
Style Loss $\downarrow$ & {\bf 0.6313} & \underline{0.6747} & 0.9996 & 0.7444 & 1.7697 & 1.2691 & 0.8252 & 0.6793 & 1.3341 & 2.1120 & 2.7492\\
CLIP Loss $\downarrow$ & {\bf 4.2334 } & \underline{6.0415} & 7.7899 & 7.0266 & 9.3485 & 8.4820 & 6.9670 & 6.5944 & 8.5434 & 7.7799 & 11.1131\\

\hline
\multicolumn{12}{c}{}\\
\multicolumn{12}{c}{\large (b) Quantitative comparisons of image super-resolution task.} \\
\multicolumn{12}{c}{}\\
\hline
& {\bf DICT} & InvSR & PSLD  & FPS-SMC$^*$ & SITCOM$^*$ & DMAP & FlowDPS & FlowChef & DOC$^*$ & TFG$^*$ & FreeDom$^*$\\
\hline
PSNR Score $\uparrow$ & {\bf 28.8600} & 21.5851 & 24.9210 & \underline{27.6305} & 25.3733 & 26.3395 & 25.0459 & 23.1086 & 26.7617 & 26.3441 & 10.7963 \\
SSIM Score $\uparrow$ & \underline{0.8233} & 0.6713 & 0.6397 & {\bf 0.8283} &  0.7805 & 0.7840 & 0.7438 & 0.7458 & 0.8167 & 0.7979 & 0.2546 \\
LPIPS Loss $\downarrow$ & {\bf 0.1573} & 0.2374 & 0.2560 & 0.2029 & 0.2460 & 0.2499 & 0.3114 & 0.2868 & \underline{0.1717} & 0.2106 & 0.7190\\
% Preference $\uparrow$ &  \\
\hline
\multicolumn{12}{c}{}\\
\multicolumn{12}{c}{\large (c) Quantitative comparisons of image deblurring task.}\\
\multicolumn{12}{c}{}\\
\hline
& {\bf DICT} & DCDP$^*$ & PSLD & FPS-SMC$^*$ & SITCOM$^*$ & DMAP & FlowDPS & FlowChef & DOC$^*$  & TFG$^*$ & FreeDom$^*$ \\
\hline
PSNR Score $\uparrow$ & \underline{27.5794} & {\bf 27.9110} & 25.8065 & 25.7486 & 23.0995 & 18.8024 & 21.1828 & 20.5104 & 25.2189 & 22.6209 & 12.3003\\
SSIM Score $\uparrow$ & {\bf 0.7736} & 0.7384 & 0.7566 & \underline{0.7665} & 0.7082 & 0.3096 & 0.5987 & 0.6067 & 0.7362  & 0.7149 & 0.3105\\
LPIPS Loss $\downarrow$ & {\bf 0.2236} & \underline{0.2325} & 0.2675 & 0.2540 & 0.3100 & 0.5541 & 0.4887 & 0.4934 & 0.2448 & 0.2869 & 0.6764\\
\hline

\end{tabular}
}
\label{tab:1}
\end{table*}

\subsection{Comparison}

For style transfer, we primarily compare against latent-space methods (including StyleShot~\cite{gao2025styleshot}, StyleStudio~\cite{lei2025stylestudio}, StyleCrafter~\cite{liu2024stylecrafter}, DEADiff~\cite{qi2024deadiff}, InstantStyle~\cite{wang2024instantstyle}, StyleAlign~\cite{hertz2024style}, CSGO~\cite{xing2024csgo}, StyleDrop~\cite{sohn2023styledrop}, TFG~\cite{ye2024tfg}, and FreeDom~\cite{yu2023freedom}), as they align better with generation conditioned on text and images. For super-resolution and deblurring, prior work spans two mainstream paradigms: pixel-space approaches (including FPS-SMC~\cite{dou2024diffusion}, SITCOM~\cite{alkhouri2025sitcom}, DOC~\cite{li2024solving}, TFG~\cite{ye2024tfg}, FreeDom~\cite{yu2023freedom}, and DCDP~\cite{li2024decoupled}), which often excel in pixel-level fidelity metrics, and latent-space approaches (including InvSR~\cite{yue2025arbitrary}, PSLD~\cite{rout2023solving}, DMAP~\cite{xu2025rethinking}, FlowDPS~\cite{kim2025flowdps}, and FlowChef~\cite{patel2024steering}), which are typically more efficient and flexible. Although our method is implemented in the latent space, we include both pixel-space and latent-space baselines for a comprehensive evaluation.

{\bf Qualitative Comparisons.} We present a qualitative comparison.

In Fig.~\ref{fig:4}(a), prior methods often exhibit weak stylization (e.g., DEADiff, StyleDrop, TFG, and FreeDom), inaccurate prompt adherence (e.g., StyleStudio, StyleCrafter, and CSGO), or content leakage from the reference style image (e.g., StyleShot, InstantStyle, and StyleAlign). In contrast, our DICT better preserves text semantics (e.g., snow in the $2nd$ row and a white towel in the $3rd$ row) while achieving stronger stylization (e.g., the dog in the $1st$ row).

In Fig.~\ref{fig:4}(b), competing approaches frequently produce artifacts (e.g., PSLD and DMAP), reduced photorealism (e.g., InvSR, FPS-SMC, and SITCOM), or overly smooth results with missing details (e.g., FlowDPS, FlowChef, and DOC), and in some cases deviate from the data distribution (e.g., TFG and FreeDom). DICT avoids noticeable artifacts and yields clearer, more realistic, and visually richer outputs (e.g., the $1st$ and $2nd$ rows), while recovering fine-grained details such as the mole in the $3rd$ row and the temple hair in the $4th$ row.

In Fig.~\ref{fig:4}(c), prior methods often exhibit distribution bias (e.g., DCDP, PSLD, FPS-SMC, SITCOM, FlowChef, and DOC), artifacts (e.g., DCDP, DMAP, and FlowDPS), and blurriness (e.g., FlowChef), and some results deviate from the data distribution (e.g., TFG and FreeDom). By comparison, DICT produces outputs that better match the target distribution (e.g., the background color in the $1st$ row) and restores fine details (e.g., the folds of the hat in the $3rd$ row).

Overall, DICT synergistically integrates Data Injection to preserve spatial information richness and Contrastive Trajectory Refinement to ensure temporal trajectory coherence. This framework enables high-fidelity conditional image generation that consistently outperforms task-specific and loss-guided baselines.

{\bf Quantitative Comparisons.} We also conduct quantitative comparisons.

For text-driven style transfer, we evaluate using Text Score~\cite{sohn2023styledrop} (prompt-image CLIP similarity), Style Loss~\cite{huang2017arbitrary} (VGG feature statistics discrepancy), and CLIP Loss~\cite{ye2024tfg} (CLIP Gram loss). As shown in Tab.~\ref{tab:1}(a), our method achieves the second-highest Text Score while obtaining the lowest Style Loss and CLIP Loss, indicating strong prompt fidelity and the most consistent stylization. Although TFG achieves a marginal lead in Text Score, it incurs substantially higher Style and CLIP losses, suggesting weaker style consistency.

 For super-resolution and deblurring, we evaluate PSNR~\cite{alkhouri2025sitcom}, SSIM~\cite{patel2024steering}, and LPIPS~\cite{xu2025rethinking}. PSNR and SSIM measure pixel-level fidelity and structural similarity, respectively, while LPIPS quantifies perceptual distance. In Tab.~\ref{tab:1}(b), DICT attains the highest PSNR and the lowest LPIPS; its SSIM is slightly below FPS-SMC, which exhibits a much higher LPIPS. In Tab.~\ref{tab:1}(c), DICT achieves the highest SSIM and the lowest LPIPS, with PSNR marginally below DCDP.

Overall, DICT integrates Data Injection and Contrastive Trajectory Refinement to achieve high-fidelity and precise conditional image generation.

\begin{figure*}[t!]
\centering

% ==================== 第一行：Style Transfer ====================
\begin{minipage}[c]{0.5\textwidth}
    \centering
    \resizebox{0.9\linewidth}{!}{
    \setlength{\tabcolsep}{0.06cm}
    \begin{tabular}{ccccc}
        { Text} & { Style} & {\bf (I) DICT} & { (II) w/o DI} & { (III) w/o CT} \\
        \includegraphics[width=0.3\linewidth]{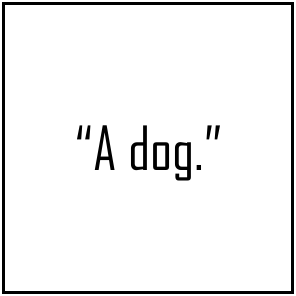} & 
        \includegraphics[width=0.3\linewidth]{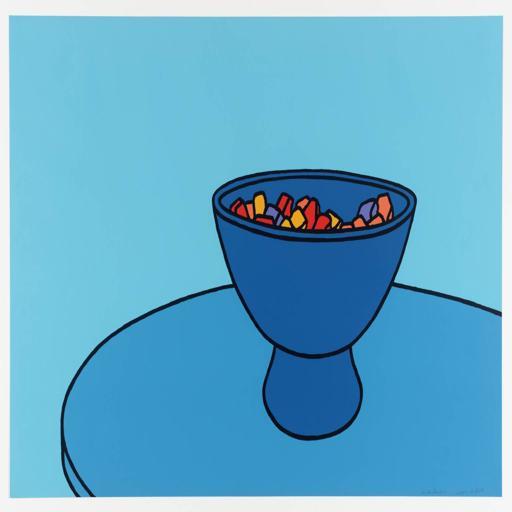} & 
        \includegraphics[width=0.3\linewidth]{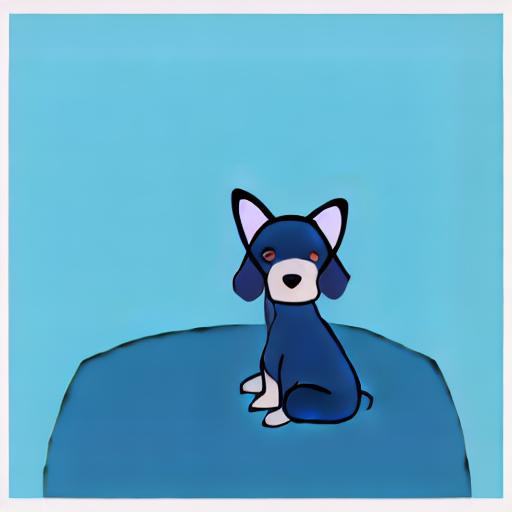} &
        \includegraphics[width=0.3\linewidth]{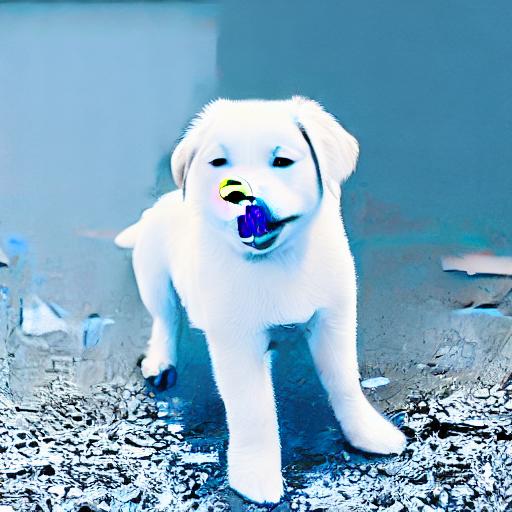} & 
        \includegraphics[width=0.3\linewidth]{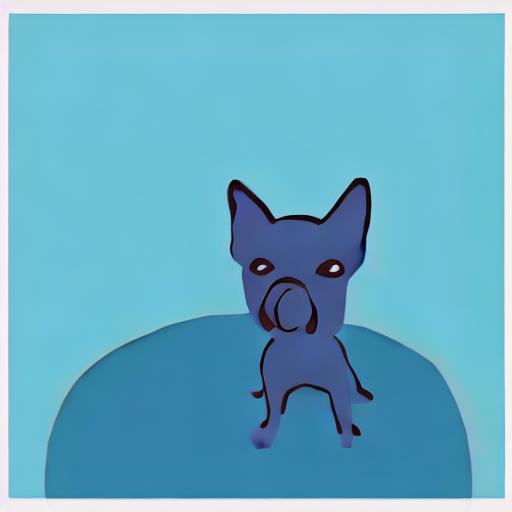} \\
        \includegraphics[width=0.3\linewidth]{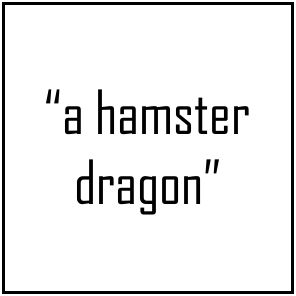} &
        \includegraphics[width=0.3\linewidth]{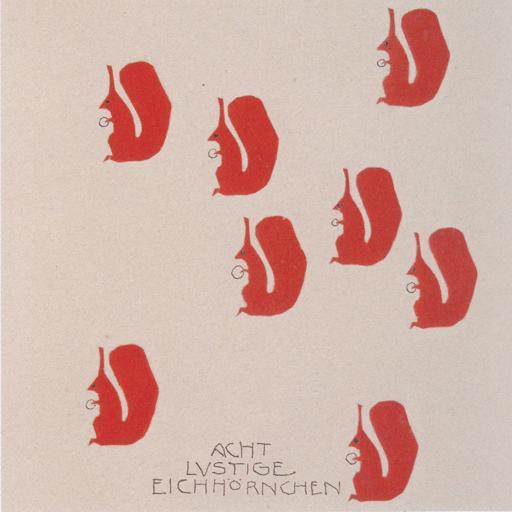} & 
        \includegraphics[width=0.3\linewidth]{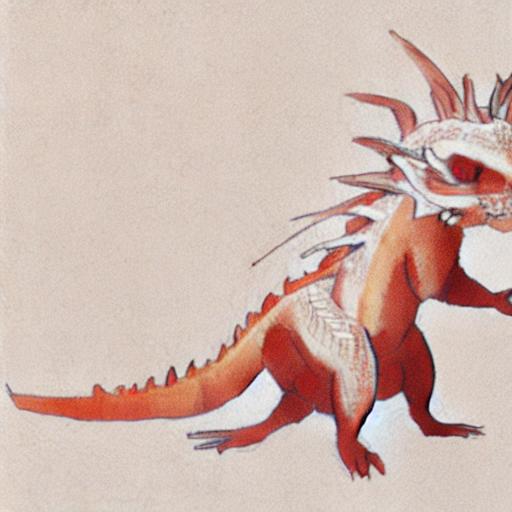} &
        \includegraphics[width=0.3\linewidth]{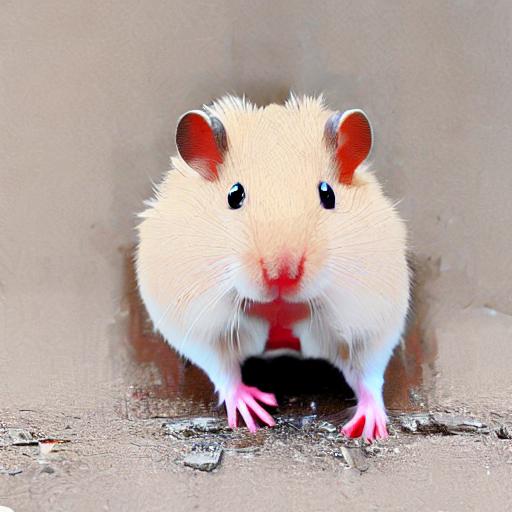} & 
        \includegraphics[width=0.3\linewidth]{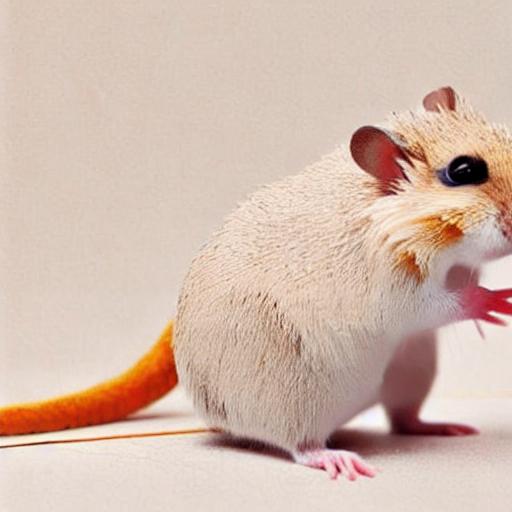} \\

    \end{tabular}}
    
\end{minipage}
\hspace{-10pt} % 极小的间距，让表格向左靠
\begin{minipage}[c]{0.5\textwidth}
    \centering
    \scalebox{0.8}{ % 进一步缩小表格以适配空间
    \setlength{\tabcolsep}{5pt}
    \begin{tabular}{cccc}
        \hline
        Metric & {\bf (I) DICT} & (II) w/o DI & (III) w/o CT \\
        \hline
        Text Score $\uparrow$ & 0.2952 & 0.2943 & {\bf 0.3008} \\
        Style Loss $\downarrow$ & {\bf0.6313} & 0.8098 & 0.9201 \\
        CLIP Loss $\downarrow$ & {\bf4.2334} & 4.7909 & 5.2108 \\
        \hline
    \end{tabular}}
\end{minipage}
\\
\vspace{2pt}
\centerline{\fontsize{8pt}{8pt}\selectfont (a) Style transfer ablation study.}
\vspace{0.3cm}

% ==================== 第二行：Super-resolution ====================
\begin{minipage}[c]{0.5\textwidth}
    \centering
    \resizebox{0.9\linewidth}{!}{
    \setlength{\tabcolsep}{0.06cm}
    \begin{tabular}{ccccc}
        { Condition} & { Target} & {\bf (I) DICT} & { (II) w/o DI} & { (III) w/o CT} \\
        \includegraphics[width=0.3\linewidth]{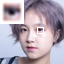}  & 
        \includegraphics[width=0.3\linewidth]{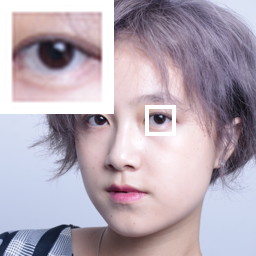} & 
        \includegraphics[width=0.3\linewidth]{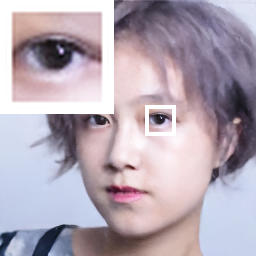} &
        \includegraphics[width=0.3\linewidth]{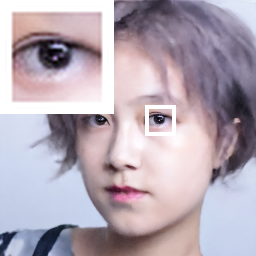} &
        \includegraphics[width=0.3\linewidth]{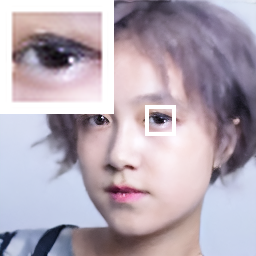} \\
        \includegraphics[width=0.3\linewidth]{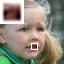} &
        \includegraphics[width=0.3\linewidth]{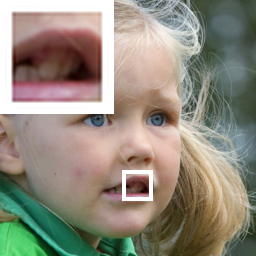} &
        \includegraphics[width=0.3\linewidth]{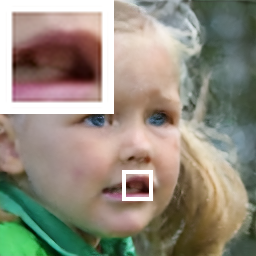} & 
        \includegraphics[width=0.3\linewidth]{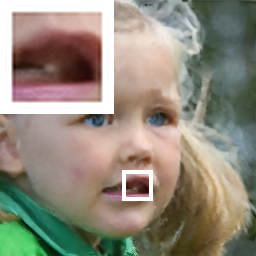} &
        \includegraphics[width=0.3\linewidth]{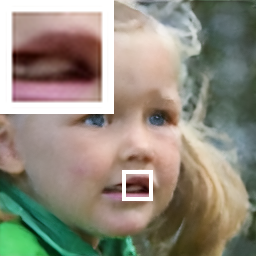} \\
    \end{tabular}}
\end{minipage}
\hspace{-10pt}
\begin{minipage}[c]{0.5\textwidth}
    \centering
    \scalebox{0.8}{
    \setlength{\tabcolsep}{5pt}
    \begin{tabular}{cccc}
        \hline
        Metric & {\bf (I) DICT} & (II) w/o DI & (III) w/o CT \\
        \hline
        PSNR $\uparrow$  & {\bf 28.8600} & 28.8155 & 28.7759 \\
        SSIM $\uparrow$  & {\bf 0.8233}  & 0.8224  & 0.8148 \\
        LPIPS $\downarrow$ & {\bf 0.1573}  & 0.1574  & 0.1818 \\
        \hline
    \end{tabular}}
\end{minipage}
\\
\vspace{2pt}
\centerline{\fontsize{8pt}{8pt}\selectfont (b) Super-resolution ablation study.}
\vspace{0.3cm}

% ==================== 第三行：Deblurring ====================
\begin{minipage}[c]{0.5\textwidth}
    \centering
    \resizebox{0.9\linewidth}{!}{
    \setlength{\tabcolsep}{0.06cm}
    \begin{tabular}{ccccc}
        {Condition} & { Target} & {\bf (I) DICT} & { (II) w/o DI} & { (III) w/o CT} \\
        \includegraphics[width=0.3\linewidth]{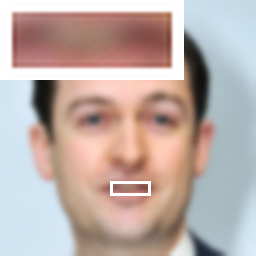} & 
        \includegraphics[width=0.3\linewidth]{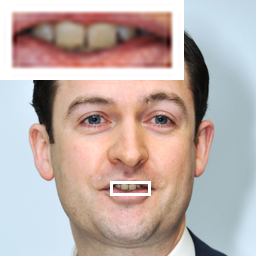}&
        \includegraphics[width=0.3\linewidth]{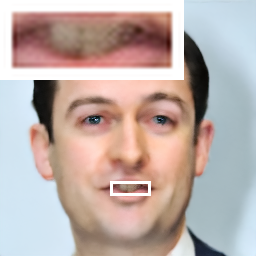} &
        \includegraphics[width=0.3\linewidth]{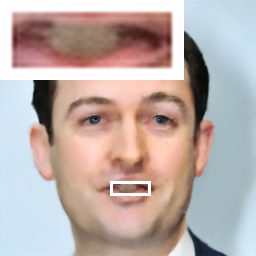} &
        \includegraphics[width=0.3\linewidth]{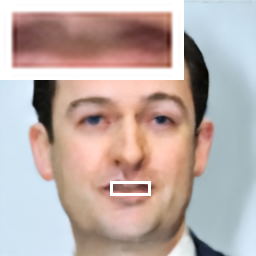} \\
        \includegraphics[width=0.3\linewidth]{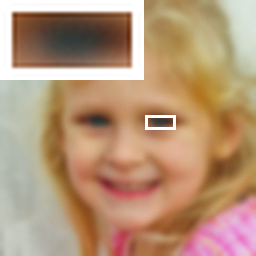} &
        \includegraphics[width=0.3\linewidth]{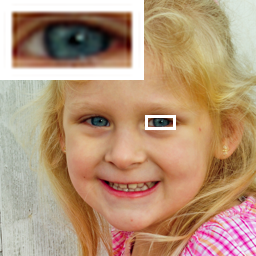} &
        \includegraphics[width=0.3\linewidth]{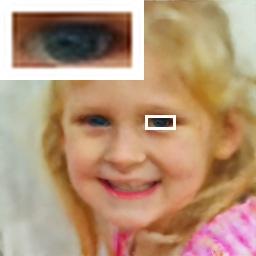} & 
        \includegraphics[width=0.3\linewidth]{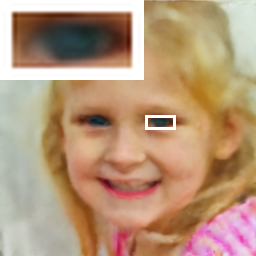} &
        \includegraphics[width=0.3\linewidth]{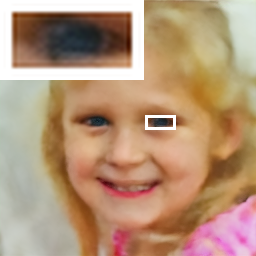} \\
    \end{tabular}}
\end{minipage}
\hspace{-10pt}
\begin{minipage}[c]{0.5\textwidth}
    \centering
    \scalebox{0.8}{
    \setlength{\tabcolsep}{5pt}
    \begin{tabular}{cccc}
    \hline
    Metric & {\bf (I) DICT} &  (II) w/o DI & (III) w/o CT \\
    \hline
    PSNR $\uparrow$  & {\bf 27.5794} & 27.5188 & 26.8616 \\
    SSIM $\uparrow$  & {\bf 0.7736}  & 0.7711  & 0.7496 \\
    LPIPS $\downarrow$ & {\bf 0.2236}  & 0.2241  & 0.2590 \\
    \hline
    \end{tabular}}
\end{minipage}
\\
\vspace{2pt}
\centerline{\fontsize{8pt}{8pt}\selectfont (c) Deblurring ablation study.}

\caption{Qualitative and quantitative ablation results. ``w/o DI'' and ``w/o CT'' denote without Data Injection and Contrastive Trajectory Refinement, respectively.}
\label{fig:ablation_all}
\end{figure*}

\subsection{Ablation Study}

To provide a more comprehensive validation of our framework, we present ablation studies on Data Injection and Contrastive Trajectory Refinement. Additional ablation results are provided in Sec.~D of the Supplementary Materials.

{\bf Data Injection.} We conduct an ablation study on Data Injection to evaluate its effectiveness. As illustrated in columns (\uppercase\expandafter{\romannumeral1}) and (\uppercase\expandafter{\romannumeral2}) of the left figure in Fig.~\ref{fig:ablation_all}(a), removing data conditioning causes the stylized results in (\uppercase\expandafter{\romannumeral2}) to exhibit a stronger inherent model bias (e.g., the overly white dog in the first row). Quantitatively, for style transfer, while omitting Data Injection leads to a marginal improvement in Text Score, it results in a substantial surge in Style Loss and CLIP Loss. This suggests that while the model may follow the text prompt slightly more easily without the Data Injection, it fails to capture the essential style, leading to a significant drop in overall fidelity. For super-resolution and deblurring, the absence of Data Injection leads to a consistent degradation across all metrics compared to the full DICT. This is also reflected in column (\uppercase\expandafter{\romannumeral2}) of the qualitative comparisons: specifically, the super-resolution results in the first row of Fig.~\ref{fig:ablation_all}(b) suffer from a loss of fine details (e.g., double eyelids), while the deblurring outputs in the second row of Fig.~\ref{fig:ablation_all}(c) fail to recover clear structures, such as eye regions. These observations further corroborate that Data Injection is indispensable for maintaining high-fidelity conditional alignment.

{\bf Contrastive Trajectory Refinement.} We evaluate the impact of Contrastive Trajectory Refinement by comparing the full DICT against a variant without it (column (\uppercase\expandafter{\romannumeral3})). Visually, omitting Contrastive Trajectory Refinement leads to noticeable stylistic discrepancies and structural artifacts. For style transfer, results in column (\uppercase\expandafter{\romannumeral3}) exhibit pronounced color shifts and inconsistent stylization, such as the unnatural hue of the dog in the $1st$ row and the distorted textures of the ``hamster dragon'' in the $2nd$ row in Fig.~\ref{fig:ablation_all}(a). In low-level vision tasks, the absence of Contrastive Trajectory Refinement compromises local structural fidelity. For super-resolution, outputs in the second row of Fig.~\ref{fig:ablation_all}(b) suffer from inaccurate local details, particularly around the teeth region; for deblurring, the recovery of fine textures is significantly constrained, as seen in the blurred eye regions in the second row of Fig.~\ref{fig:ablation_all}(c). These qualitative shortcomings are reflected in the quantitative results in Fig.~\ref{fig:ablation_all}(b) and Fig.~\ref{fig:ablation_all}(c), where removing Contrastive Trajectory Refinement consistently leads to inferior metrics, thereby validating its essential role for higher-quality generation.

\begin{figure}[t]
\centering
\resizebox{0.9\textwidth}{!}{
\setlength{\tabcolsep}{0.000cm} % 调整列间距
\renewcommand{\arraystretch}{0.00}  % 调整行距
\begin{tabular}{>{\centering\arraybackslash}m{1.0cm} 
 >{\centering\arraybackslash}m{2cm} >{\centering\arraybackslash}m{2cm} >{\centering\arraybackslash}m{1cm} >{\centering\arraybackslash}m{2cm} >{\centering\arraybackslash}m{2cm} >{\centering\arraybackslash}m{1cm} >{\centering\arraybackslash}m{2cm} >{\centering\arraybackslash}m{2cm}}
 \\
  Inputs & PixArt & DICT & Inputs  &  PixArt &  DICT & Inputs  &  PixArt &  DICT
\\
\includegraphics[width=1\linewidth]{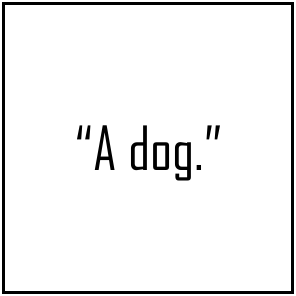} 
\includegraphics[width=1\linewidth]{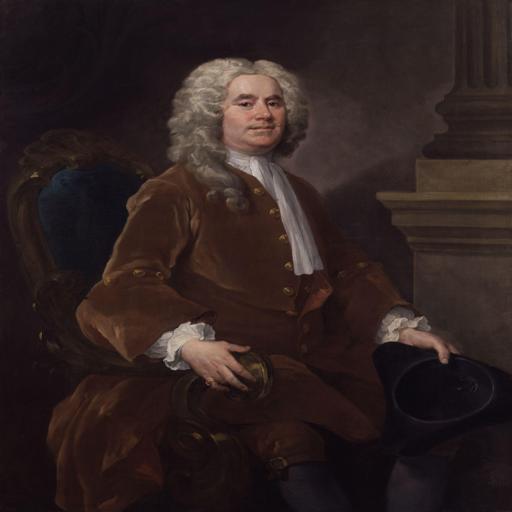} & 
\includegraphics[width=1\linewidth]{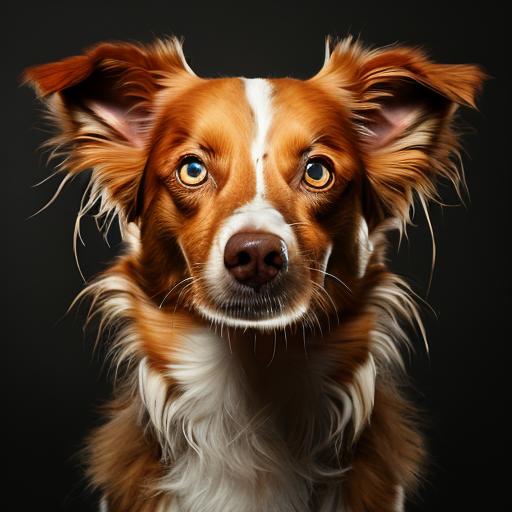}  &
\includegraphics[width=1\linewidth]{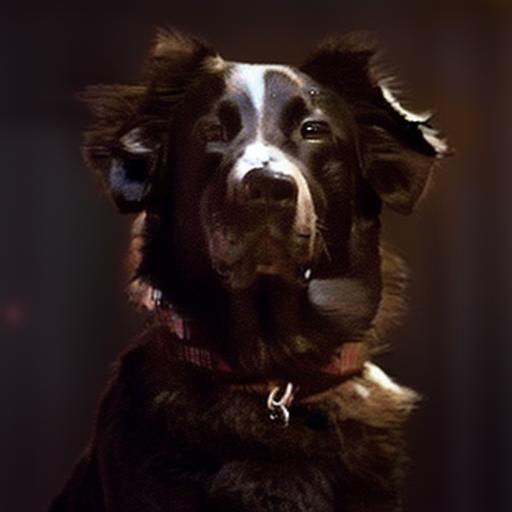}  & 
\includegraphics[width=1\linewidth]{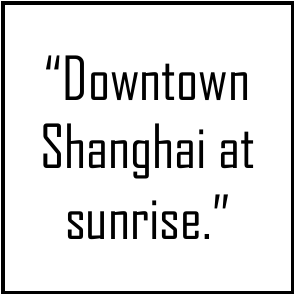} 
\includegraphics[width=1\linewidth]{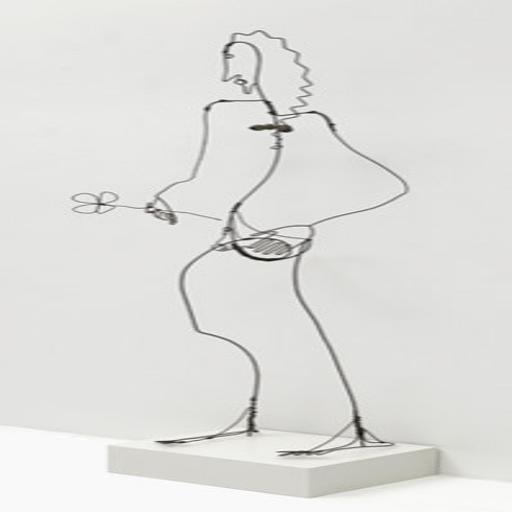} &
\includegraphics[width=1\linewidth]{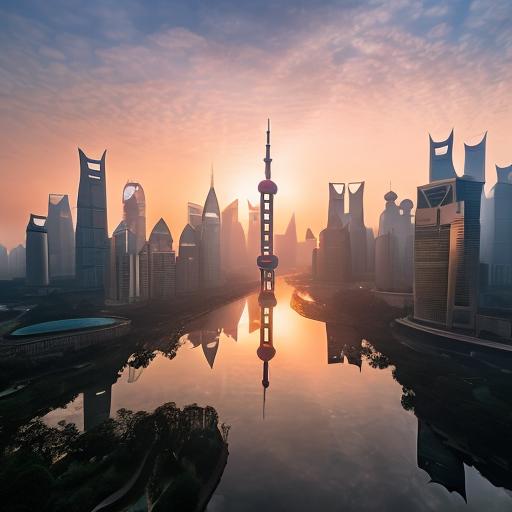} &  
\includegraphics[width=1\linewidth]{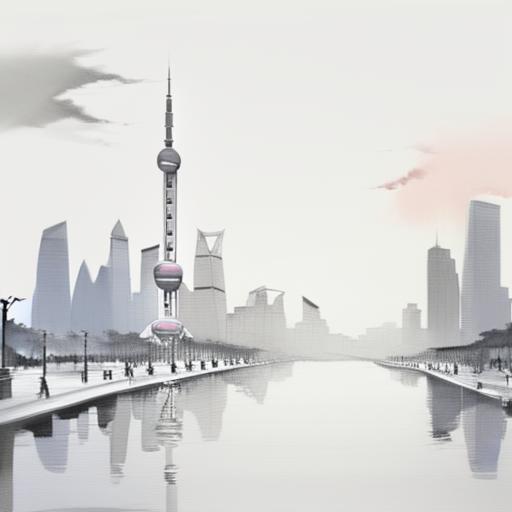} &
\includegraphics[width=1\linewidth]{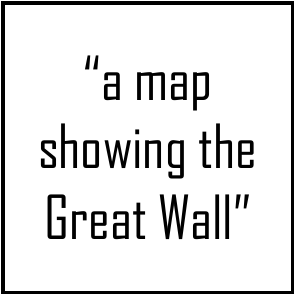} 
\includegraphics[width=1\linewidth]{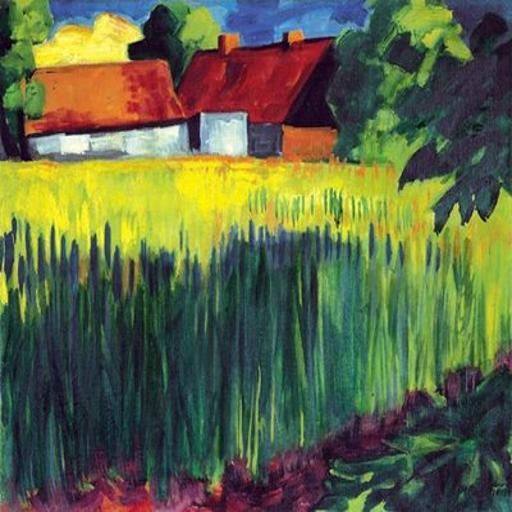} &
\includegraphics[width=1\linewidth]{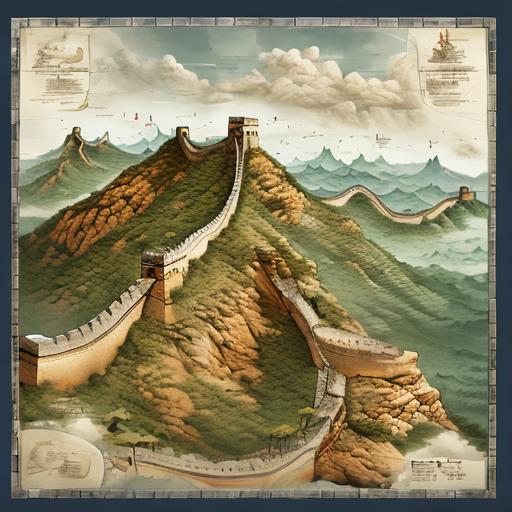} &  
\includegraphics[width=1\linewidth]{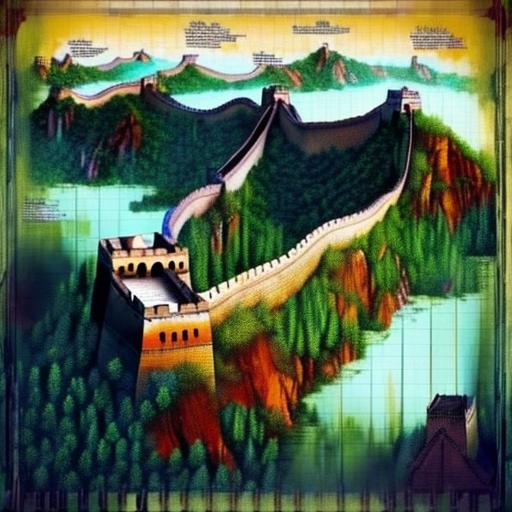}

\end{tabular}}

\caption{Qualitative DiT-based stylization results.}
\label{fig:dit}

\end{figure}

\section{Additional Experiments}

To demonstrate its backbone-agnostic generalization, we integrate DICT into a Diffusion Transformer (DiT), specifically PixArt~\cite{chen2024pixart}. On the demanding text-driven style transfer task, DICT consistently yields high-fidelity results (Fig.~\ref{fig:dit}), confirming its architectural independence.

\section{Conclusion}

In this paper, we present a unified perspective on conditional image generation. We reveal that traditional loss-guided methods compress high-dimensional conditional cues into a single scalar objective, causing severe information bottlenecks and cumulative sampling errors. To overcome these challenges, we introduce DICT, a versatile inference framework that synergistically combines Data Injection with Contrastive Trajectory Refinement. By directly injecting noise-perturbed conditional signals and denoising them under guidance, DICT preserves the spatial richness lost in scalar-based methods and adaptively distills task-relevant information. Furthermore, its contrastive refinement mechanism ensures coherence between adjacent trajectory steps, progressively improving generation quality. Extensive evaluations across diverse tasks, including style transfer, super-resolution, and deblurring, demonstrate that DICT significantly enhances both fidelity and perceptual quality. Ultimately, DICT offers a robust, training-free, and unified solution for high-quality conditional image generation without requiring task-specific architectures.

\section*{Acknowledgements}
This work was supported by the National Key Research and Development Program of China under Grant Nos. 2024YFF0907802 and 2024YFF0907803, and the Research Fund for International Scientists of the National Natural Science Foundation of China under Grant No. 72350710798.

\clearpage
\hypersetup{
    colorlinks=true,
    urlcolor=magenta, 
}

% Support for ORCID icon
% \usepackage{orcidlink}

\renewcommand{\dbltopfraction}{0.95} 
\renewcommand{\textfraction}{0.05}  
\renewcommand{\floatpagefraction}{0.8}

% \usepackage{marvosym}
% \usepackage{array}
% \newcommand{\corrauth}{\textsuperscript{\textrm{\Letter}}}
% \begin{document}
\appendix
% ---------------------------------------------------------------
% TODO REVIEW: Replace with your title
% \title{DICT: Data Injection and Contrastive Trajectory Refinement for Conditional Image Generation with Diffusion Models\\
% -- Supplementary Materials --} 

% TODO REVIEW: If the paper title is too long for the running head, you can set
% an abbreviated paper title here. If not, comment out.
\titlerunning{DICT}

% TODO FINAL: Replace with your author list. 
% Include the authors' OCRID for the camera-ready version, if at all possible.
% \author{Chunnan Shang \and
% Xin Zhang \and 
% Zhizhong Wang \corrauth \and
% Hongwei Wang \corrauth }

% TODO FINAL: Replace with an abbreviated list of authors.
\authorrunning{C.~Shang et al.}
% First names are abbreviated in the running head.
% If there are more than two authors, 'et al.' is used.

% TODO FINAL: Replace with your institution list.
% \institute{Zhejiang University, Hangzhou, China\\
% \email{ \{chunnan.22, hongweiwang\}@intl.zju.edu.cn \\
% \{22421264, endywon\}@zju.edu.cn} \\
% \url{https://github.com/scn-00/DICT}
% }

% \maketitle
% \let\thefootnote\relax\footnotetext{\corrauth Corresponding authors.}

\section{Theoretical Intuitions}

While the empirical success of DICT is demonstrated across various conditional generation tasks, fully characterizing its underlying mathematical dynamics remains an open challenge. In this section, we provide a conceptual framework and theoretical intuition to help explain why our method is effective. We offer these perspectives as a foundational hypothesis to inspire future theoretical discussions and more rigorous analyses.

\subsection{Data Injection}

Data Injection as a Non-gradient Score Proxy. Existing training-free methods typically rely on a scalar-guidance paradigm, which creates a severe information bottleneck by compressing high-dimensional conditional cues into a single scalar loss objective. Because scalar gradients fundamentally struggle to encapsulate complex spatial textures and fine-grained contents, the generative model is often forced to hallucinate missing details. To conceptualize how our method overcomes this, our Data Injection mechanism can be viewed as a non-gradient score proxy through a latent spatial displacement. Through the formulation $c_t = \alpha_{data} z_t + (1-\alpha_{data}) \hat{c}_t$, we dynamically fuse the current latent $z_t$ with the processed condition variable $\hat{c}_t$. Under this perspective, $\hat{c}_t$ acts similarly to an empirical Dirac distribution, serving as a robust anchor to guide the generative trajectory. We hypothesize that by performing guided denoising on these injected signals, DICT implicitly approximates the maximization of the high-dimensional conditional likelihood $p(x|z_t)$, bypassing the dimensional compression of scalar gradients. This proxy preserves complex structural priors and enables the model to adaptively distill target data manifolds.

\subsection{Contrastive Trajectory Refinement}

Contrastive Trajectory Refinement as a Discrete Lyapunov Stabilizer. Furthermore, relying solely on a single-step scalar loss $\mathcal{L}_1$ provides only a static optimization target, largely ignoring the temporal structure of the reverse diffusion trajectory. During the iterative denoising process, numerical solvers inevitably introduce Local Truncation Errors (LTE, $\mathcal{O}(\Delta t^{k+1})$). Without an explicit trajectory-aware correction mechanism, we observe that these minor step-wise inaccuracies can compound, leading to tangential drift and degraded sample quality. We posit that our Contrastive Trajectory Refinement objective ($\mathcal{L}_2$) functions conceptually as a discrete Lyapunov stabilizer. By establishing a pairwise contrastive constraint, formulated as $\mathcal{L}_2 = \max(\mathcal{L}_1(z_{0|t-1}, y) - \mathcal{L}_1(z_{0|t}, y) + \alpha_{margin}, 0)$, we enforce a condition that shares similarities with strict monotonic energy decay ($\Delta \mathcal{L}_1 \le -\alpha_{margin}$) in dynamical systems. It requires the intermediate prediction $z_{0|t-1}$ to be closer to the condition target than the preceding prediction $z_{0|t}$ by a safety margin of $\alpha_{margin}$. From a manifold learning standpoint, this design can be understood as a soft Manifold Projection Operator, generating a normal constraint force that helps neutralize orthogonal deviations caused by LTE, thereby promoting trajectory coherence and progressive refinement.

\section{Detailed Experiment Settings}
\label{sec:Detailed Experiment Settings}

\subsection{Style Transfer} 

{\bf Condition Preprocessing.} In style transfer, the objective is to capture the stylistic essence of a reference image. Since style is a highly abstract concept that is challenging to represent explicitly or initialize effectively, we do not apply any additional operations to the label, i.e., $M=None$.

{\bf Loss Function.} To faithfully capture and transfer the stylistic attributes from the reference style image $x$, we employ a multi-level style alignment strategy. This objective is achieved through two complementary components: a VGG-based statistical loss and a CLIP-based semantic style loss.

{\it VGG Style Loss.} Following the established practices in~\cite{gatys2016image,Huang2017ArbitraryST}, we also leverage channel-wise statistics to represent stylistic characteristics. By aligning the mean and standard deviation of feature maps, the model captures the color distribution and texture motifs of the reference style image $x$ without being constrained by its original spatial structure. Specifically, the style loss is formulated as:
\begin{equation}
\begin{aligned}
\label{eq:A1}
&\mathcal{L}_{VGG}=\sum_{i=1}^5(||\mu(\phi_{i}(D(z_{0|t})))-\mu(\phi_{i}(x))||_2\\
&\qquad\qquad+||\sigma(\phi_{i}(D(z_{0|t})))-\sigma(\phi_{i}(x))||_2),\\
\end{aligned}
\end{equation}
where $z_{0|t}$ is the predicted clean stylized image at time step $t$. $\phi_{i}$ denotes the feature extraction mapping corresponding to the $Relui\_1$ ($i \in \left\{1,\dots, 5\right\}$) layers of a pre-trained VGG-19 network~\cite{simonyan2014very}. $D$ represents the decoder of the stable diffusion models~\cite{rombach2022high}. $\mu$ and $\sigma$ denote the channel-wise mean and standard deviation, respectively.

{\it CLIP Style Loss.} While VGG features effectively capture mid-level textures, CLIP~\cite{radford2021learning} encapsulates high-level semantic abstractions that are invariant to low-level pixel fluctuations. To align the global artistic atmosphere, we compute the Gram matrix~\cite{gatys2016image} on the CLIP image embeddings. By capturing the second-order correlations between high-level semantic features, the model is able to represent and transfer complex stylistic patterns and global aesthetic motifs that transcend simple texture matching. The CLIP style loss is formulated as:
\begin{equation}
\begin{aligned}
\label{eq:A2}
\mathcal{L}_{CLIP}=||G(CLIP(D(z_{0|t})))-G(CLIP(x))||_2,
\end{aligned}
\end{equation}
where $G(\cdot)$ denotes the Gram matrix computation operation~\cite{gatys2016image}. 

{\it Total Loss.} The final objective for stylistic alignment is formulated as:
\begin{equation}
\begin{aligned}
\label{eq:A3}
\mathcal{L}_{1}(z_{0|t},x)=\mathcal{L}_{CLIP}+\alpha_{loss} \times \mathcal{L}_{VGG},
\end{aligned}
\end{equation}
where $\alpha_{loss}$ is a hyperparameter balancing the two style components. We set $\alpha_{loss}=0.02$ in our experiments.

\begin{table}[t]
\centering
\caption{Hyperparameter settings for different tasks in DICT.}
\label{tab: suppl1}
\setlength{\tabcolsep}{3pt}
\scalebox{0.78}{
\begin{tabular}{lccc}
\toprule
Hyperparameters & Style Transfer & Image Super-resolution & Image Deblurring \\ 
\midrule
Base Model   & SD v1.4   & LDM  & LDM  \\
Sampling Steps   & 50 & 200 & 200   \\
Latent Weight ($\alpha_{data}$) & 0.88 & 0.88  & 0.88   \\
Noise Weight ($\gamma_{data}$)  & 0.2  & 0.1   & 0.1 \\
Iterative Updates ($N_{iter1}$, $N_{iter2}$) & 3, 3 & 3, 3  & 3, 4  \\
Initial Update Rate ($\eta_{1}$, $\eta_{2}$)    & 2$^\diamond$, 2$^\diamond$ & [0.08, 0.015, 0.006]$^{\dagger}$, 0.02   & [0.13, 0.015, 0.006]$^*$, 0.02  \\
Data Injection Steps ($T_1$)   & 1-18  & 1-8  & 1-50   \\
Contrastive Trajectory Margin ($\alpha_{margin}$)& 1.0  & 1.0  & 1.0 \\ 
\bottomrule
\multicolumn{4}{l}{\footnotesize $^{\diamond}$ This task employs a varying parameter schedule~\cite{ye2024tfg} for $\eta_1$ and $\eta_2$.}\\
\multicolumn{4}{l}{\footnotesize $^{\dagger}$ This task employs a three-stage schedule for $\eta_1$: steps 1--130, 131--160, and 161--200.}\\
\multicolumn{4}{l}{\footnotesize $^{*}$ This task employs a three-stage schedule for $\eta_1$: steps 1--130, 131--150, and 151--200.}\\

\end{tabular}
}
\end{table}

{\bf Implementation Details.} Our DICT framework is built upon the Stable Diffusion v1.4~\cite{rombach2022high} architecture. We employ 50 sampling steps for the inference process to balance generation quality and efficiency. Regarding other hyperparameters, $\alpha_{data}$ regulates the blending intensity of noise-perturbed stylistic information within the input latent space. Meanwhile, $\gamma_{data}$ serves as a modulation weight to determine the usage ratio of the adaptively distilled stylistic cues. These cues are formulated within the style-conditioned branch, where the model automatically identifies and extracts informative stylistic components to constitute the final noise estimate. To strike a balance between generation quality and computational efficiency, the number of iterative updates for refining clean data at both the current and subsequent time steps, denoted as $N_{iter1}$ and $N_{iter2}$  are carefully configured. We adopt a decaying parameter strategy~\cite{ye2024tfg} to schedule the iterative update rates $\eta_1$ and $\eta_2$  starting from a predefined initial value to ensure stable convergence. To effectively incorporate stylistic attributes while mitigating potential content leakage from the reference image, Data Injection is restricted to the early stages of the reverse diffusion process; this allows for the synthesis of highly stylized images that remain semantically consistent with the text prompt. Furthermore, for the distillation of process knowledge, the margin parameter $\alpha_{margin}$ is employed to calibrate the perceptual distance between the style reference $x$, the positive sample $z_{0|t-1}$ and the negative sample $z_{0|t}$. Detailed hyperparameter settings for each task are provided in Tab.~\ref{tab: suppl1}.

\subsection{Image Super-resolution}

{\bf Condition Preprocessing.} Since the degraded observation $x$ is provided at a lower resolution, we employ a standard interpolation function~\cite{paszke2017automatic} to upsample it to the target output resolution, i.e., $M=upsampling$. This ensures spatial consistency between the synthesized results and the intended target output, effectively accommodating the inherent downsampling and upsampling operations within the diffusion model's encoder-decoder architecture. Crucially, this pre-alignment allows the latent representation of the condition to be directly integrated with the generative latents, as they now share the same spatial manifold. Such a configuration is essential for preserving fine-grained structural alignment during the subsequent data injection process. While more sophisticated upsampling techniques could theoretically further enhance the performance of our framework, we utilize simple interpolation to maintain computational efficiency during the initialization of the conditional signal.

{\bf Loss Function.} Following the formulations in\cite{zhang2025improving,chung2022diffusion}, we employ a Euclidean loss to enforce consistency between the generated latent and the degraded observation $x$. The loss is defined as:
\begin{equation}
\begin{aligned}
\label{eq:A4}
\mathcal{L}_{1}(z_{0|t},x)=||\mathcal{A}(D(z_{0|t}))-x||_2,
\end{aligned}
\end{equation}
where $D(z_{0|t})$ denotes the decoded clean image estimate at time step $t$, $\mathcal{A}$ represents the downsampling degradation operator used to project the high-resolution estimate back into the observation space.

{\bf Implementation Details.} We evaluate our DICT framework on the FFHQ dataset~\cite{karras2019style}, utilizing the pre-trained weights provided by~\cite{rout2023solving} without any further fine-tuning or task-specific optimization. We configure the sampling process to 200 steps to achieve an optimal balance between super-resolution quality and computational efficiency. Regarding the hyperparameters, $\alpha_{data}$ regulates the blending intensity of noise-perturbed structural priors from the degraded observation $x$ within the input latent space. Meanwhile, $\gamma_{data}$ serves as a modulation weight to determine the usage ratio of the adaptively distilled content details. These details are formulated within the condition-guided branch, where the model automatically identifies and extracts informative high-frequency components to constitute the final noise estimate, thereby ensuring the fidelity of the restored image. To optimize the iterative refinement of clean data estimates, we perform $N_{iter1}$ and $N_{iter2}$ updates per step using the AdamW~\cite{loshchilov2017decoupled} optimizer. To ensure stable and efficient convergence, the update rate $\eta_1$ follows a three-stage piecewise constant schedule: the initial phase accelerates convergence, the intermediate phase stabilizes the refinement trajectory, and the final phase suppresses late-stage oscillations. The update rate $\eta_2$  is fixed to ensure a steady evolution of the sampling path. Furthermore, as the degraded condition $x$ primarily provides coarse semantic skeletons and lacks fine-grained details, Data Injection is restricted to the very early stage of the reverse diffusion. This strategy facilitates rapid semantic grounding and structural initialization early in the denoising process without introducing low-quality artifacts from the observation. Finally, the margin parameter $\alpha_{margin}$ is employed to calibrate the perceptual distance between the condition image $x$ and the refined samples ($z_{0|t-1}$ and $z_{0|t}$). Detailed hyperparameter values and stage-wise step ranges for the super-resolution task are provided in Tab.~\ref{tab: suppl1}.

\subsection{Image Deblurring}

{\bf Condition Preprocessing.} Since the condition image $x$ is corrupted by a Gaussian blur, we apply a Wiener filter~\cite{wiener1964extrapolation} to obtain a preliminarily deblurred version, denoted as $M=WienerDeblur$. Unlike the interpolation used in super-resolution for spatial alignment, this preprocessing step is specifically designed to recover latent high-frequency cues and provide a more informative initialization for the subsequent Data Injection. By enhancing the signal saliency of the condition image, we enable the diffusion model to more effectively distill task-relevant structural features from the noise-perturbed signals during the early denoising stages. While more sophisticated deblurring techniques could theoretically further enhance the performance of our framework, we utilize the Wiener filter to maintain an optimal balance between signal enhancement and computational efficiency.

{\bf Loss Function.} Following the formulations in~\cite{zhang2025improving,chung2022diffusion}, we employ a Euclidean loss to enforce data-fidelity between the generated latent and the condition image $x$. The loss is defined as:
\begin{equation}
\begin{aligned}
\label{eq:A5}
\mathcal{L}{1}(z{0|t},x)=||\mathcal{A}(D(z_{0|t}))-x||_2,
\end{aligned}
\end{equation}
where $D(z_{0|t})$ denotes the decoded clean image estimate at time step $t$, and $\mathcal{A}$ represents the blurring degradation operator used to project the high-fidelity estimate back into the space of the condition image.

{\bf Implementation Details.} We evaluate our DICT framework on the FFHQ dataset~\cite{karras2019style}, utilizing the pre-trained weights provided by~\cite{rout2023solving} without any further fine-tuning or task-specific optimization. The sampling process is configured to 200 steps to achieve an optimal trade-off between deblurring quality and inference speed. Regarding the hyperparameters, $\alpha_{data}$ regulates the blending intensity of noise-perturbed structural information from the condition image $x$ within the input latent space. Meanwhile, $\gamma_{data}$ serves as a modulation weight to determine the usage ratio of the adaptively distilled content details. These details are formulated within the condition-guided branch, where the model automatically identifies and extracts informative high-frequency components to constitute the final noise estimate, thereby ensuring the fidelity of the generated image. To optimize the iterative refinement of clean data estimates, we perform $N_{iter1}$ and $N_{iter2}$ updates per step using the AdamW~\cite{loshchilov2017decoupled} optimizer. To ensure stable and efficient convergence, the update rate $\eta_1$ follows a three-stage piecewise constant schedule: the initial phase accelerates convergence, the intermediate phase stabilizes the refinement trajectory, and the final phase suppresses late-stage oscillations. The update rate $\eta_2$ is fixed to ensure a steady evolution of the sampling path. Furthermore, as the blurred condition image $x$ primarily provides coarse semantic skeletons and lacks fine-grained textures, Data Injection is restricted to the early stage of the reverse diffusion process. This strategy facilitates rapid semantic grounding and provides a more informative initialization for the denoising process without introducing blurred artifacts from the observation. Finally, the margin parameter $\alpha_{margin}$ is employed to calibrate the perceptual distance between the condition image $x$ and the refined samples ($z_{0|t-1}$ and $z_{0|t}$). Detailed hyperparameter values and stage-wise step ranges for the image deblurring task are provided in Tab.~\ref{tab: suppl1}.

\section{Parameter Selection Strategy}
\label{sec:Parameter Selection Strategy}

\begin{figure}[t]
\centering

\resizebox{0.98\textwidth}{!}{

\setlength{\tabcolsep}{0.05cm} 
\renewcommand{\arraystretch}{0.5}  
\begin{tabular}{cccccc}
Text & Style & $T_{1}=0$ & $T_{1}=9$ & $T_{1}=18$ & $T_{1}=27$ \\
\includegraphics[width=0.14\linewidth]{st_images/A_dog.pdf} & 
\includegraphics[width=0.14\linewidth]{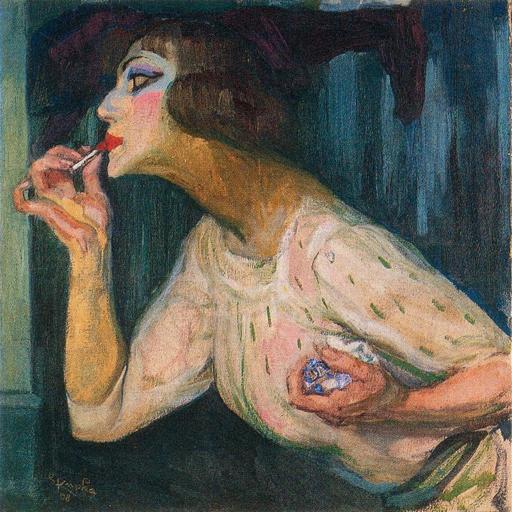} & 
\includegraphics[width=0.14\linewidth]{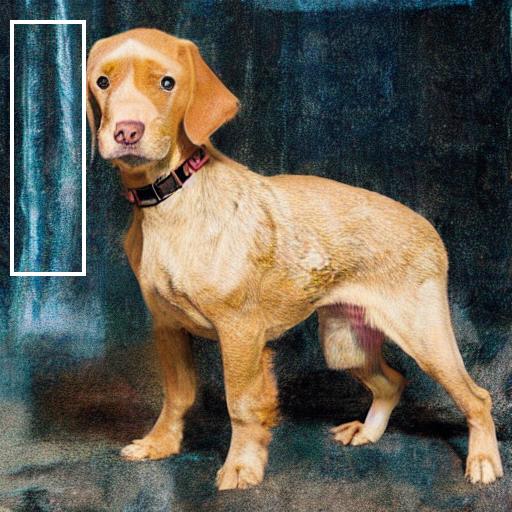} & 
\includegraphics[width=0.14\linewidth]{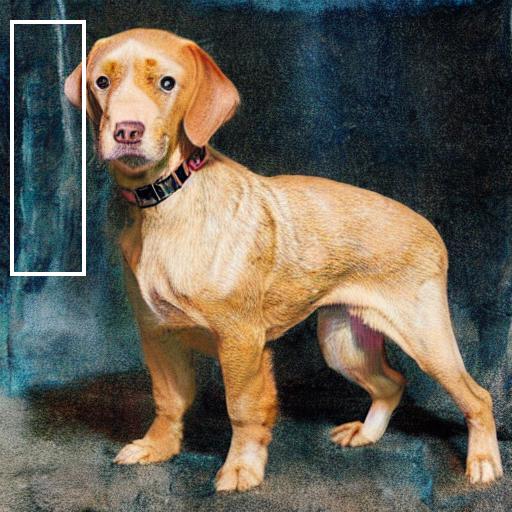} & 
\includegraphics[width=0.14\linewidth]{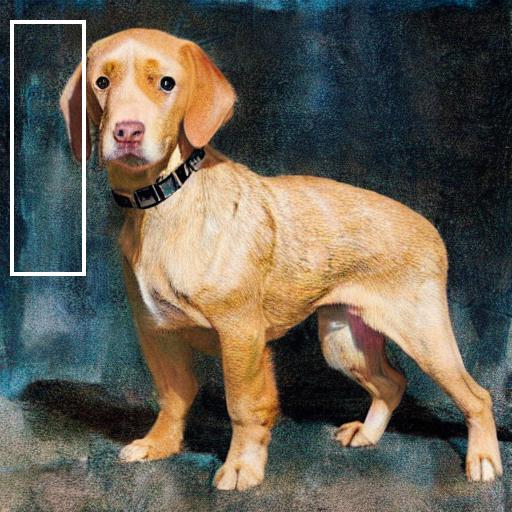} & 
\includegraphics[width=0.14\linewidth]{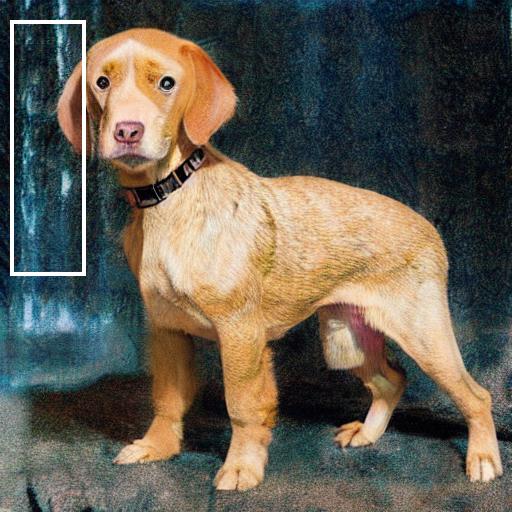} 

\end{tabular}
}

\caption{Qualitative results showing the effect of parameter $T_1$ on the style transfer task.}
\label{fig:sm1}
\end{figure}

$\boldsymbol{T_{1}}$: This parameter specifies the number of early reverse diffusion steps used for injecting task-relevant data priors. Its choice depends on both the downstream task and the fidelity of the information preserved in the processed condition. For style transfer, an optimal $T_1$ (e.g., $T_1=18$ in Fig.~\ref{fig:sm1}) is required to achieve high-quality stylistic fusion. At lower values of $T_1$ (e.g., $T_1=0$ and $T_1=9$ ), the model fails to sufficiently capture the intricate stylistic cues of the reference image, particularly within the background regions, where the artistic textures remain noticeably underdeveloped. Conversely, an overly large $T_1$ disrupts the generation process. As the model approaches the data manifold, it prioritizes high-fidelity reconstruction over stylistic guidance. Consequently, injecting perturbed conditions during this sensitive phase introduces disruptive noise that the model cannot reconcile with the established structural prior. Lacking sufficient residual steps to correct these anomalies, the injected signals manifest as conspicuous artifacts rather than coherent stylistic textures. Likewise, for image super-resolution, a large $T_{1}$ may incorporate excessive blurred structures into the generation process, thereby compromising the model’s ability to recover high-frequency details. In contrast, a small $T_{1}$ provides only limited prior guidance and may consequently restrict super-resolution performance. For image deblurring, a large $T_{1}$ may transfer biases from conventional deblurring algorithms into the sampling process, causing the generated results to inherit these artifacts. Conversely, a small $T_{1}$ may underexploit structural cues, leading to suboptimal deblurring quality. Therefore, across these three tasks, the choice of $T_{1}$ is made by balancing prior injection against potential task-specific side effects, guided by both theoretical considerations and empirical observations.

\begin{figure}
\centering
\resizebox{0.55\textwidth}{!}{
\setlength{\tabcolsep}{0.02cm} 
\renewcommand{\arraystretch}{0.5} 
\begin{tabular}{
 >{\centering\arraybackslash}m{0.4cm} >{\centering\arraybackslash}m{2.0cm} >{\centering\arraybackslash}m{2.0cm} >{\centering\arraybackslash}m{2.0cm} >{\centering\arraybackslash}m{2.0cm}}
  &&  Text & Style &
 \\
 &&  \includegraphics[width=1.0\linewidth]{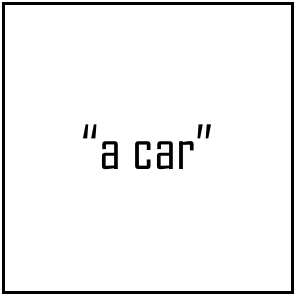} &  \includegraphics[width=1.0\linewidth]{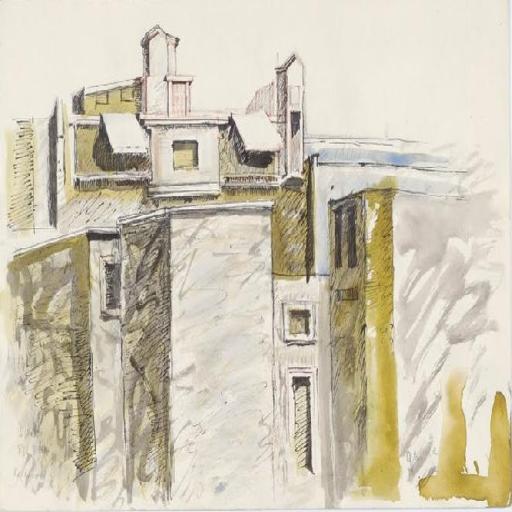} &
 \\
  & $ N_{iter2} = 1$ &  $2$ &  $3$ &  $4$ 
 \\
\rotatebox[origin=rb]{270}{$N_{iter1} = 1$} & 
 \includegraphics[width=1.0\linewidth]{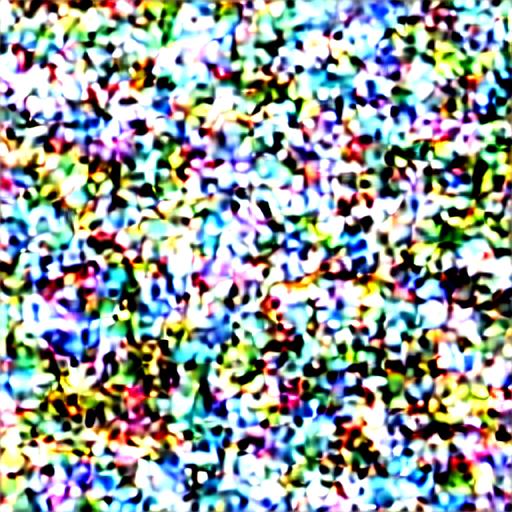} & \includegraphics[width=1.0\linewidth]{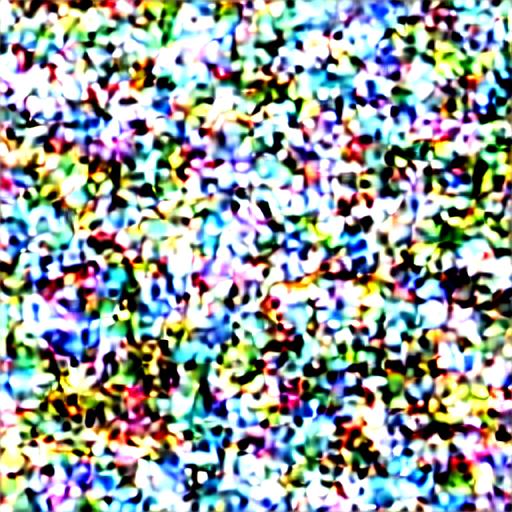} & \includegraphics[width=1.0\linewidth]{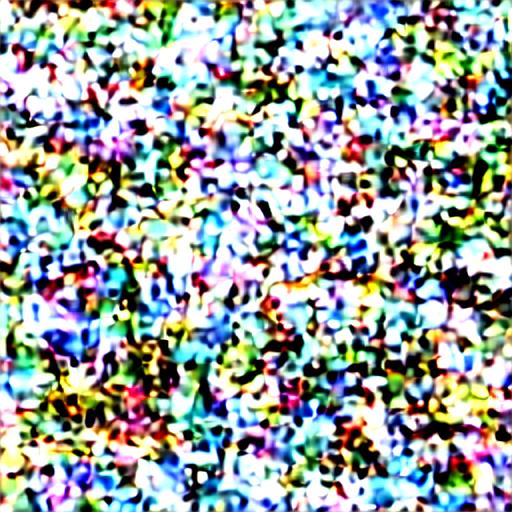} & \includegraphics[width=1.0\linewidth]{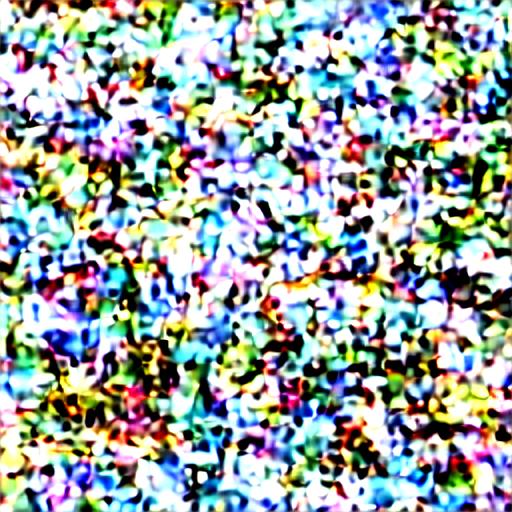}  \\
 \rotatebox[origin=rb]{270}{$2$} &
 \includegraphics[width=1.0\linewidth]{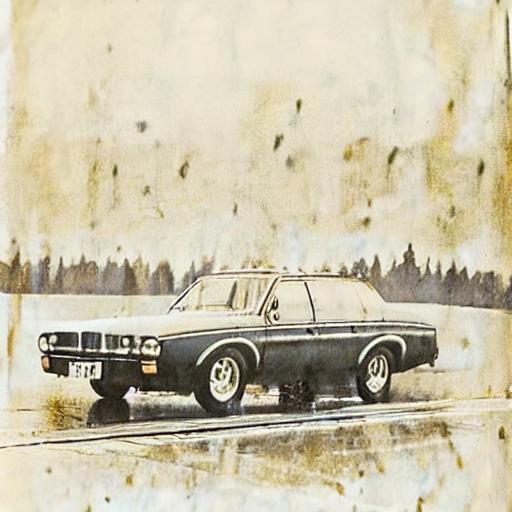} & \includegraphics[width=1.0\linewidth]{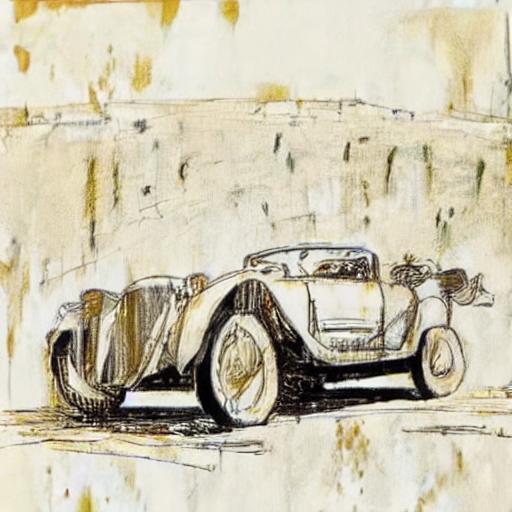} & \includegraphics[width=1.0\linewidth]{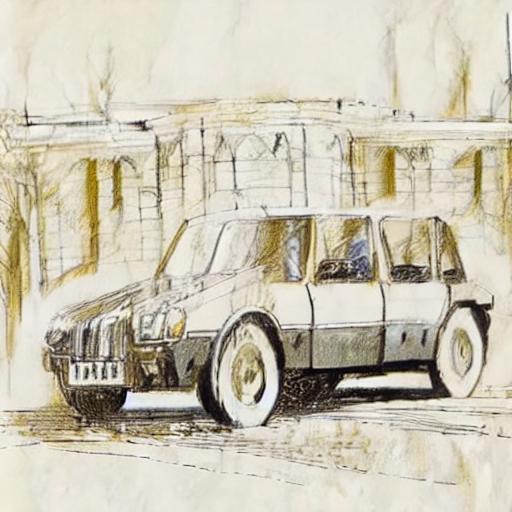} & \includegraphics[width=1.0\linewidth]{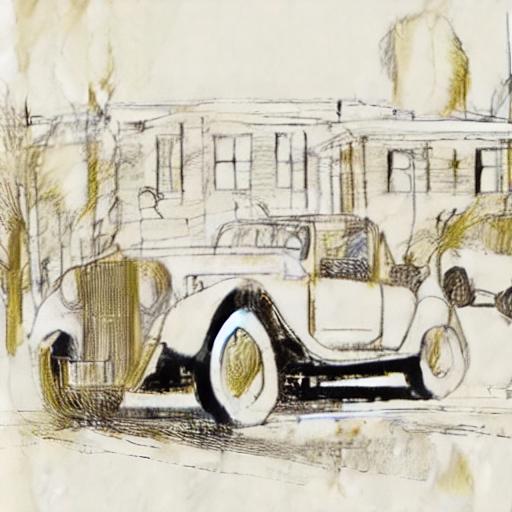}  \\
  \rotatebox[origin=rb]{270}{$3$} & 
 \includegraphics[width=1.0\linewidth]{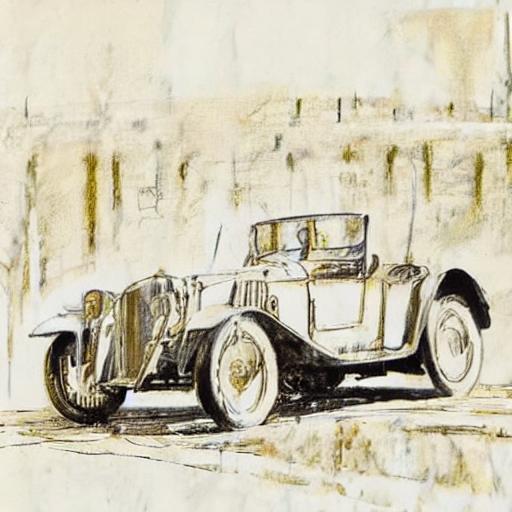} & \includegraphics[width=1.0\linewidth]{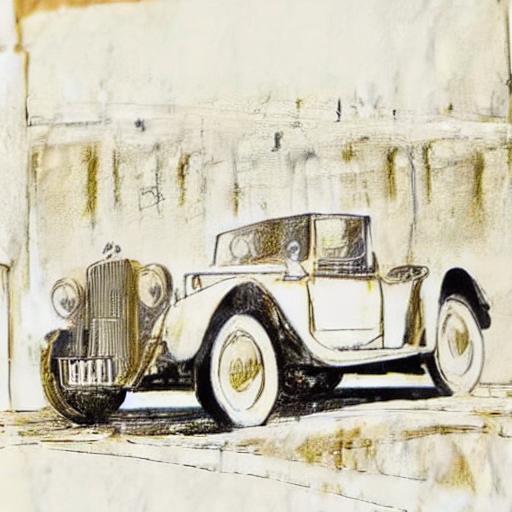} & \includegraphics[width=1.0\linewidth]{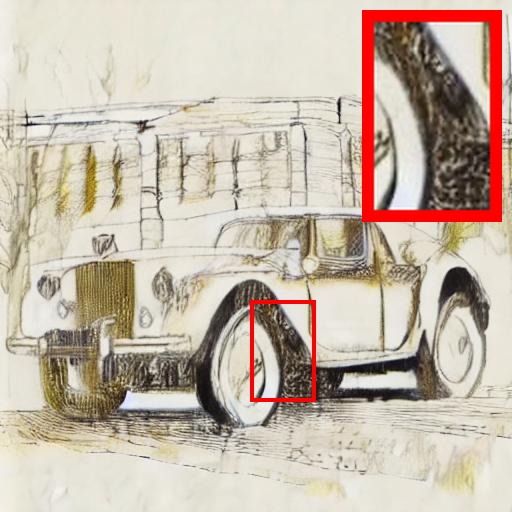} & \includegraphics[width=1.0\linewidth]{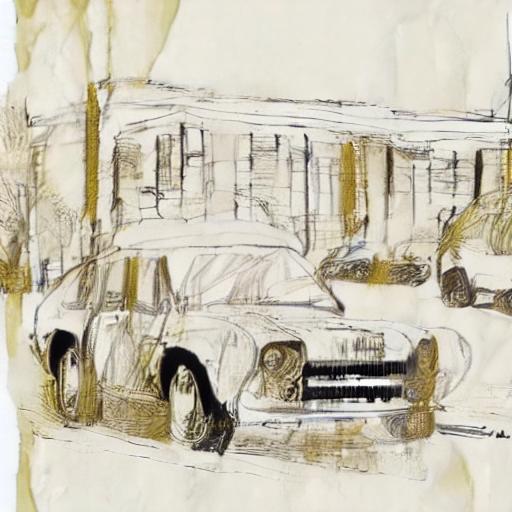}  \\
    \rotatebox[origin=rb]{270}{$4$} &
 \includegraphics[width=1.0\linewidth]{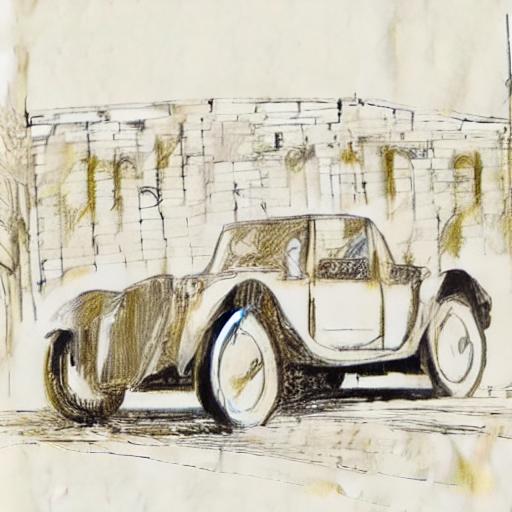} & \includegraphics[width=1.0\linewidth]{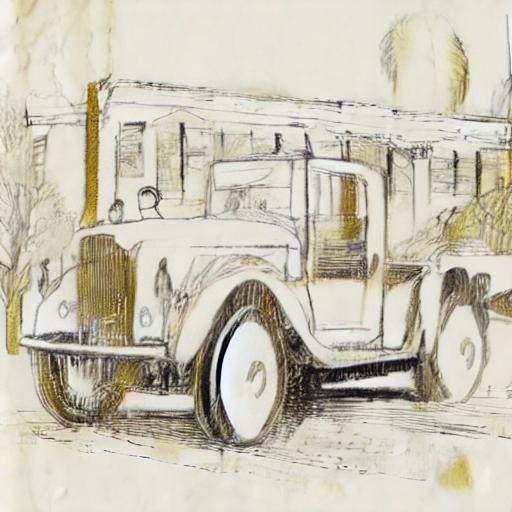} & \includegraphics[width=1.0\linewidth]{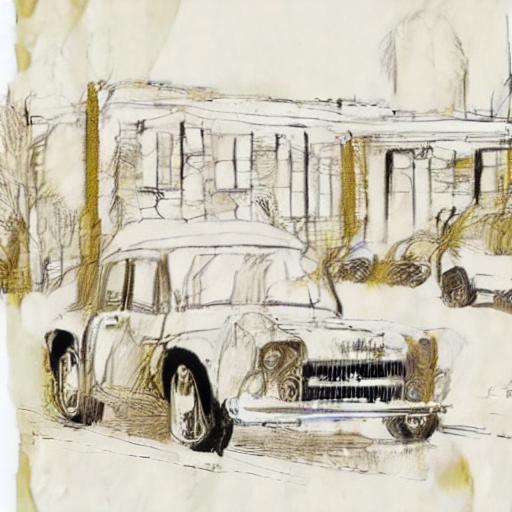} & \includegraphics[width=1.0\linewidth]{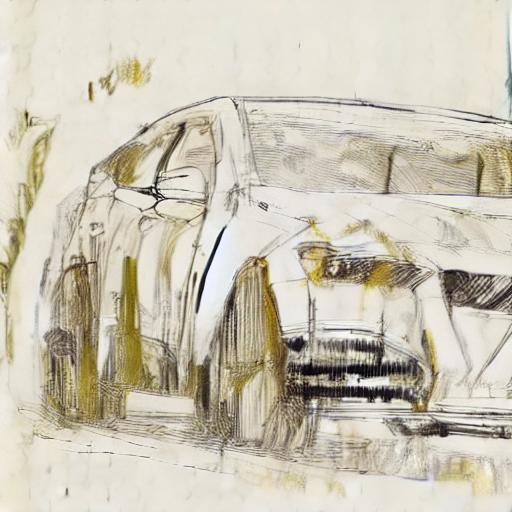} 

\end{tabular}}
\caption{Qualitative results of style transfer with varying parameter $N_{iter1}$ and $N_{iter2}$.}
\label{fig:sm2}
\end{figure}

$\boldsymbol{N_{iter1}}$: This parameter represents the number of iterative updates applied to the predicted data at the current timestep during the reverse diffusion process. The choice of ${N_{iter1}}$ depends on the specific task, whether the updated data remains within the model's expected data distribution, and the overall effectiveness of the updates. For instance, in style transfer, since we primarily learn the overall style and detailed textures rather than the exact content of the style image, we can appropriately reduce the number of iterations. Conversely, in super-resolution and deblurring tasks, the model needs to recover specific semantics, layouts, and content details from the labels, so we can increase the number of iterations to improve performance. It's important to note that excessively increasing the number of iterations is akin to an optimization process. This is extremely time-consuming. Therefore, our selection of $N_{iter1}$ is a balance between theoretical considerations and practical experience. As shown in Fig.~\ref{fig:sm2}, when $N_{iter1}=1$, the output suffers from noise as the data fails to adequately align with the model's expected data distribution. Increasing $N_{iter1}$ enhances stylistic fidelity and realigns the data distribution to mitigate the model’s inherent bias toward generic object colors; specifically, it allows the wheels to adopt the stylistic tones of the reference image rather than defaulting to a standard black. This increase also facilitates the suppression of sky artifacts. Balancing visual performance and efficiency, we set $N_{iter1}=3$.

$\boldsymbol{N_{iter2}}$: The parameter represents the number of iterative updates applied to the predicted data for the next timestep during the reverse diffusion process. The selection of $N_{iter2}$ depends on the effectiveness of the updates and whether the updated data distribution remains within the model's valid range. The primary goal of this update step is to widen the loss gap between the predicted data and the label data for the current timestep and the previous step. In other words, we want the loss for the current timestep to be significantly smaller than the loss for the before one. For this reason, we can appropriately increase the number of iterations. Therefore, based on the theoretical analysis and practical experience mentioned above, we have made a reasonable choice for the $N_{iter2}$ value across the three tasks, balancing both efficiency and effectiveness. For example, as shown in Fig.~\ref{fig:sm2}, increasing $N_{iter2}$ enhances stylistic alignment with the reference image. Like $N_{iter1}$, this increased stylistic intensity helps mitigate the model's inherent bias toward generic object representations. However, excessive optimization can compromise the structural integrity of the objects. Balancing stylistic fidelity and structural preservation, we set $N_{iter2}=3$.

\begin{figure}
\centering
\resizebox{0.55\textwidth}{!}{
\setlength{\tabcolsep}{0.02cm} % 调整列间距
\renewcommand{\arraystretch}{0.5}  % 调整行距
\begin{tabular}{
 >{\centering\arraybackslash}m{0.4cm} >{\centering\arraybackslash}m{2.0cm} >{\centering\arraybackslash}m{2.0cm} >{\centering\arraybackslash}m{2.0cm} >{\centering\arraybackslash}m{2.0cm}}
  &&  Text & Style &
 \\
 &&  \includegraphics[width=1.0\linewidth]{st_images/A_dog.pdf} &  \includegraphics[width=1.0\linewidth]{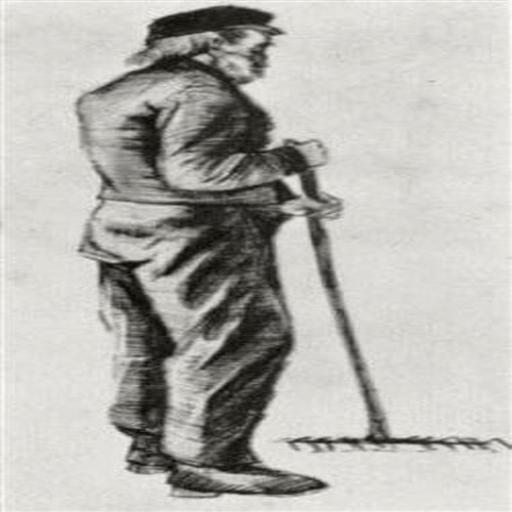} &
 \\
  & $ \eta_{2} = 1$ &  $2$ &  $3$ &  $4$ 
 \\
\rotatebox[origin=rb]{270}{$\eta_{1} = 1$} & 
 \includegraphics[width=1.0\linewidth]{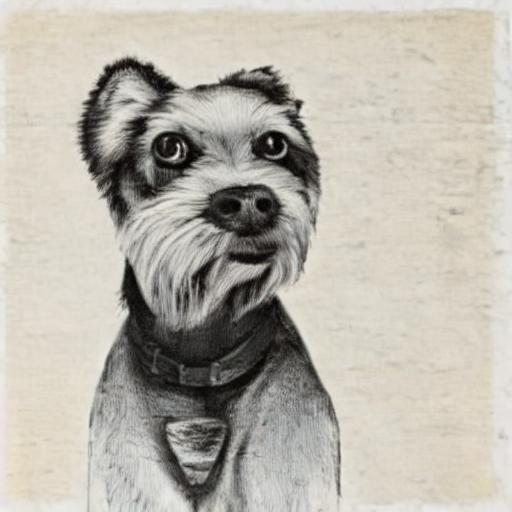} & \includegraphics[width=1.0\linewidth]{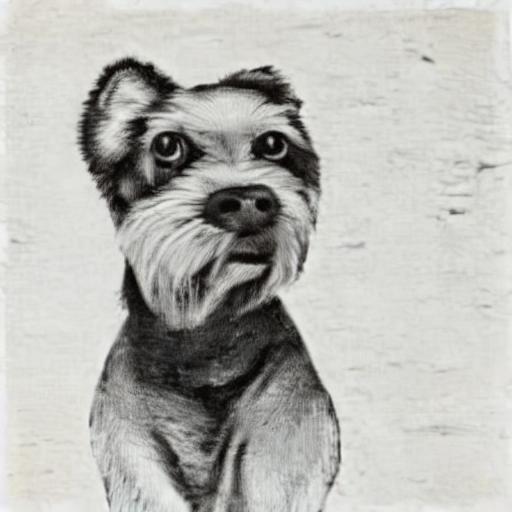} & \includegraphics[width=1.0\linewidth]{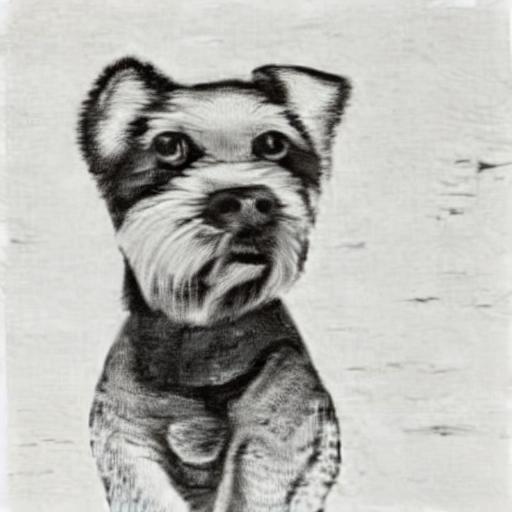} & \includegraphics[width=1.0\linewidth]{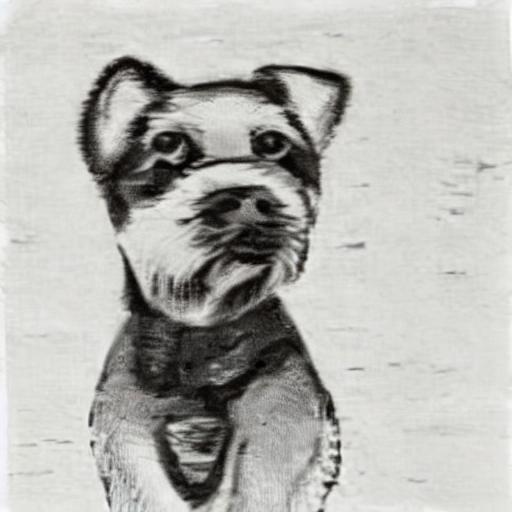}  \\
 \rotatebox[origin=rb]{270}{$2$} &
 \includegraphics[width=1.0\linewidth]{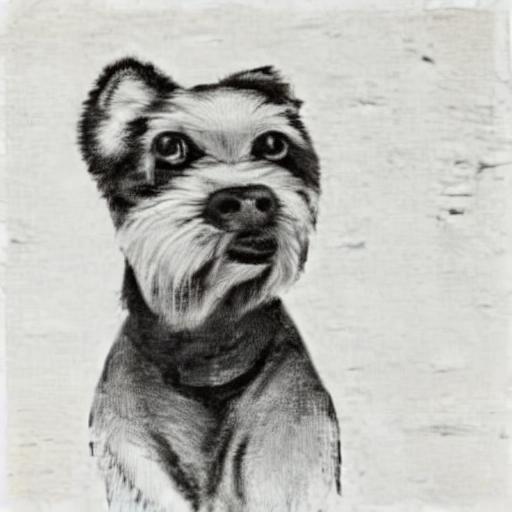} & \includegraphics[width=1.0\linewidth]{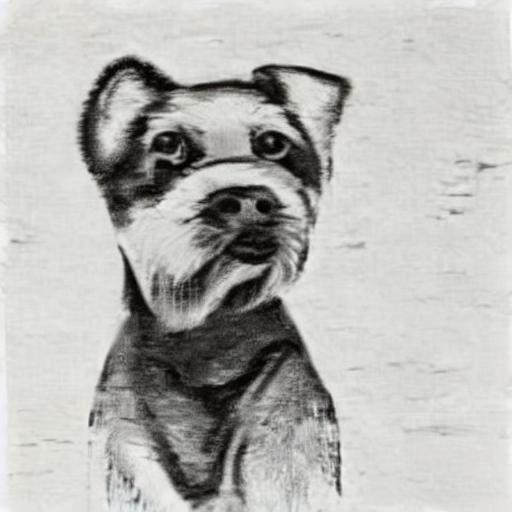} & \includegraphics[width=1.0\linewidth]{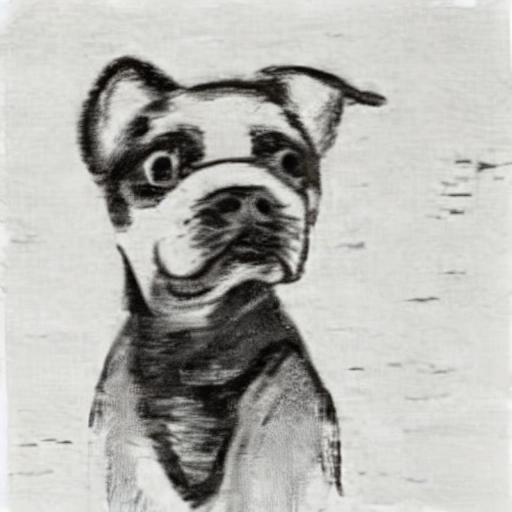} & \includegraphics[width=1.0\linewidth]{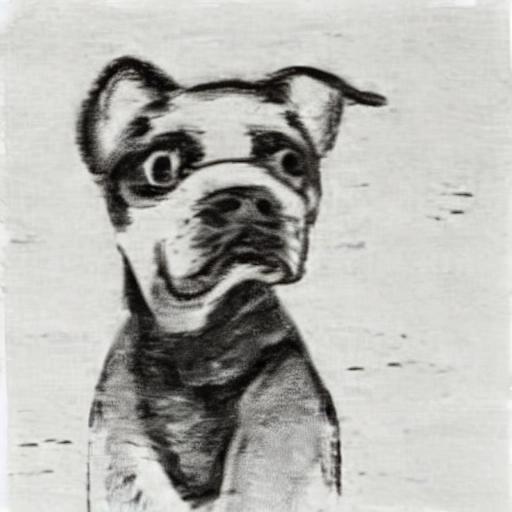}  \\
  \rotatebox[origin=rb]{270}{$3$} & 
 \includegraphics[width=1.0\linewidth]{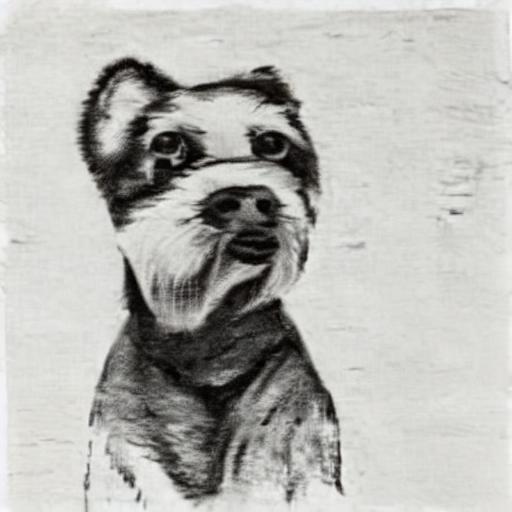} & \includegraphics[width=1.0\linewidth]{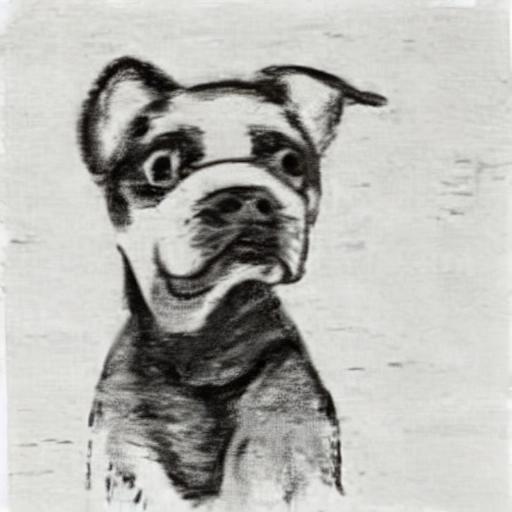} & \includegraphics[width=1.0\linewidth]{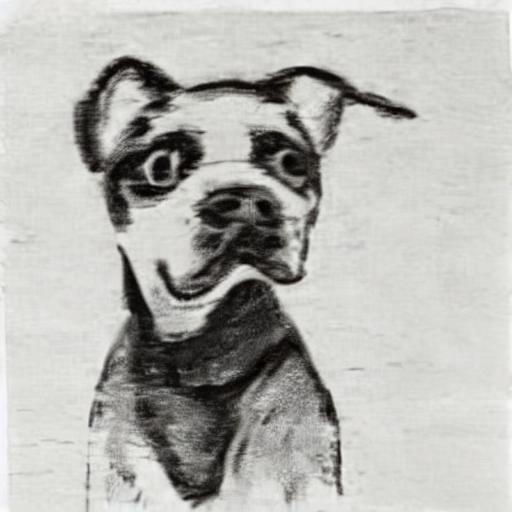} & \includegraphics[width=1.0\linewidth]{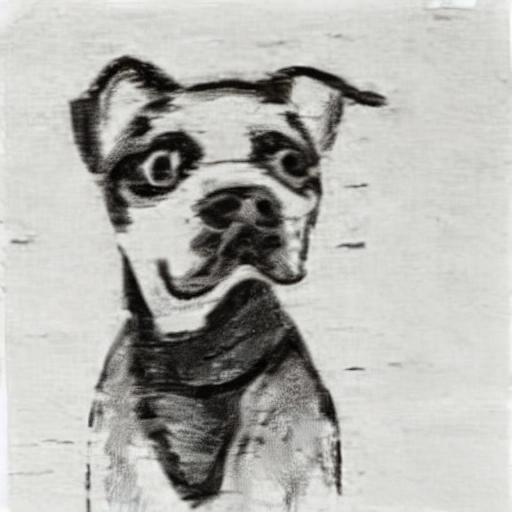}  \\
    \rotatebox[origin=rb]{270}{$4$} &
 \includegraphics[width=1.0\linewidth]{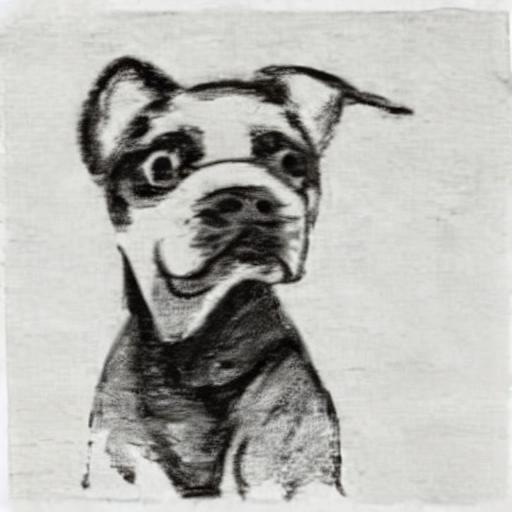} & \includegraphics[width=1.0\linewidth]{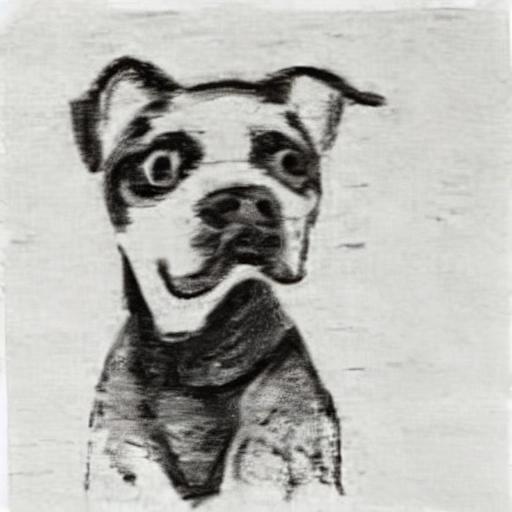} & \includegraphics[width=1.0\linewidth]{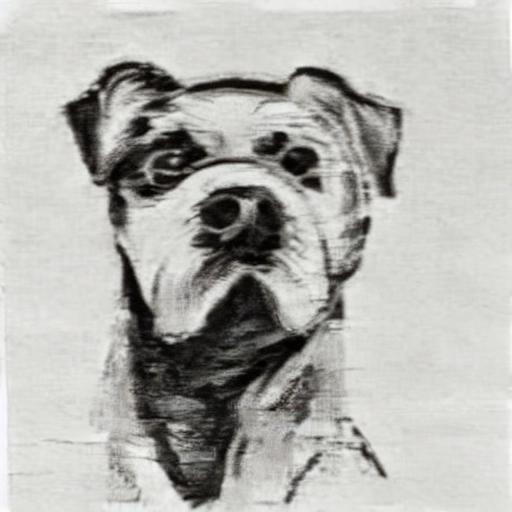} & \includegraphics[width=1.0\linewidth]{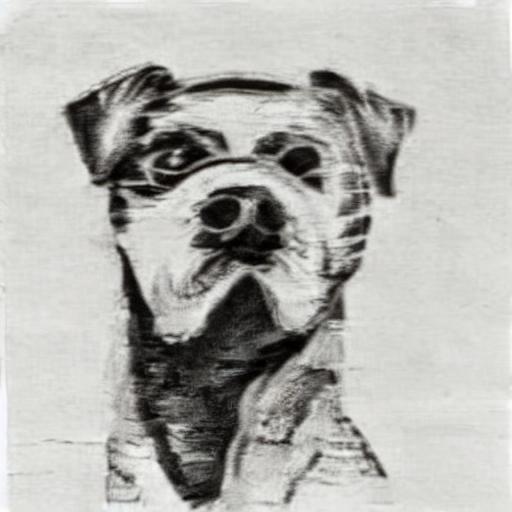} 

\end{tabular}}
\caption{Qualitative results of style transfer with varying parameter $\eta_{1}$ and $\eta_{2}$.}
\label{fig:sm6}
\end{figure}

$\boldsymbol{\eta_{1}}$: The parameter controls the speed of iterative updates to the predicted data at the current timestep during the reverse diffusion process. The magnitude of the learning rate depends on the required optimization granularity and the stability of the optimization for a specific task. For instance, in style transfer, since we are concerned with the overall feature distribution rather than the semantic content of the style image, the optimization is relatively coarse. Thus, we can use a higher learning rate. However, an excessively high learning rate prioritizes stylistic transfer at the expense of structural integrity, leading to the degradation of content-related features. As demonstrated in Fig.~\ref{fig:sm6}, an excessively low  $\eta_{1}$ leads to insufficient style transfer (e.g., incomplete transfer of the background color), whereas an excessively high value induces optimization oscillations, causing a significant loss of structural fidelity (e.g., distortion of the dog's shape). Conversely, for super-resolution and deblurring, where optimization requires attention to specific layouts and structural details, a moderate learning rate is more appropriate. A learning rate that is too high can cause the optimization to oscillate, while one that is too low may not be sufficient. Furthermore, because optimization for super-resolution and deblurring initially focuses on the overall layout and then on specific content and structural details, our learning rate should be gradually decreased. This allows for a more stable refinement of fine details in later stages. If the $\eta_{1}$ remains too high and causes oscillations at this point, the generated details could be significantly different. Based on these theoretical insights and empirical observations, we utilize task-specific schedules for $\eta_{1}$ to ensure both stylistic expression and structural fidelity.

$\boldsymbol{\eta_{2}}$: The parameter controls the speed of iterative updates for the predicted data at the next timestep during the reverse diffusion process. The learning rate is determined by the stability of the optimization and whether the updated data distribution remains within the model's valid range. Our objective for this optimization step is to ensure that the loss between the next timestep's predicted data and the label is smaller than the loss from the current timestep. We also want the optimization to remain stable and the updated data to stay within a reasonable range. Consequently, we must fine-tune the next timestep's predicted data within a narrow, sensible range, as it cannot be set too high. An excessively high $\eta_{2}$ can cause the optimization to overshoot the optimal solution, leading to oscillations where the loss value bounces erratically instead of steadily decreasing. This is analogous to taking giant, clumsy steps toward a destination, often overshooting the mark and having to backtrack. Such instability can prevent the model from converging to a valid solution, or it might cause the updated data to fall outside the model's expected distribution, generating nonsensical or artefactual results. Conversely, a learning rate that is too low can lead to under-optimization. The model's updates would be too small and gradual, making the convergence process exceedingly slow and potentially preventing the model from ever reaching an optimal state within a reasonable timeframe. For example, as illustrated in Fig.~\ref{fig:sm6} for style transfer, an overly low initial value ($\eta_{2}=1$) results in insufficient stylization, while an excessively high value ($\eta_{2}=4$) compromises structural integrity, leading to significant distortion. Consequently, we determine the optimal initial value of $\eta_{2}=2$ through empirical analysis to strike a balance between optimization stability and performance.

\begin{figure}[t]
\centering

\resizebox{0.98\textwidth}{!}{

\setlength{\tabcolsep}{0.05cm}
\renewcommand{\arraystretch}{0.5}  
\begin{tabular}{cccccc}
Text & Style & $\alpha_{data}=0$ & $\alpha_{data}=0.44$ & $\alpha_{data}=0.88$ & $\alpha_{data}=1$ \\
\includegraphics[width=0.14\linewidth]{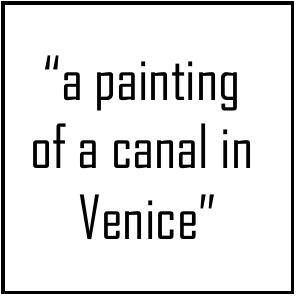} & 
\includegraphics[width=0.14\linewidth]{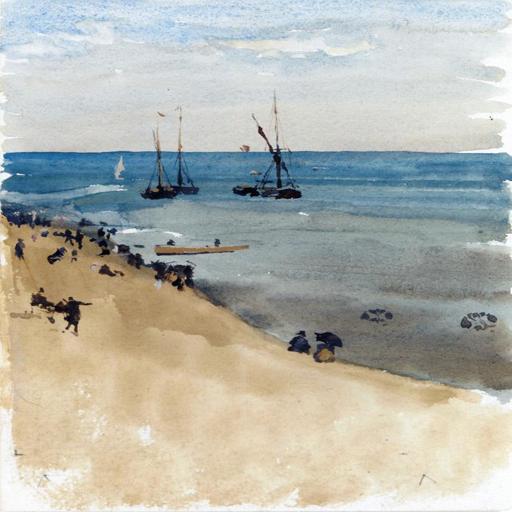} & 
\includegraphics[width=0.14\linewidth]{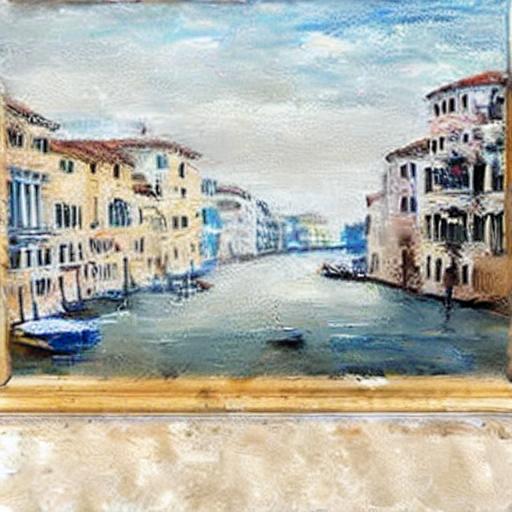} & 
\includegraphics[width=0.14\linewidth]{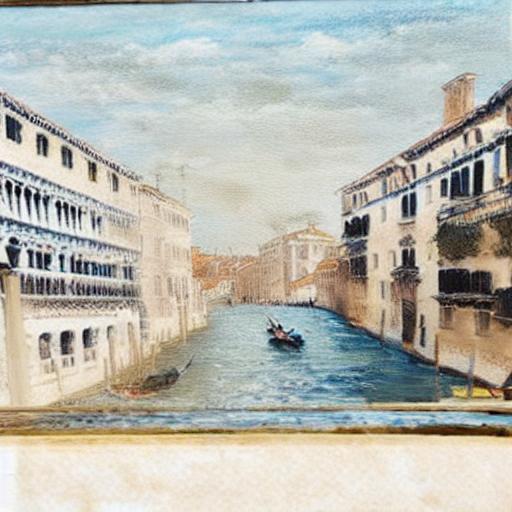} & 
\includegraphics[width=0.14\linewidth]{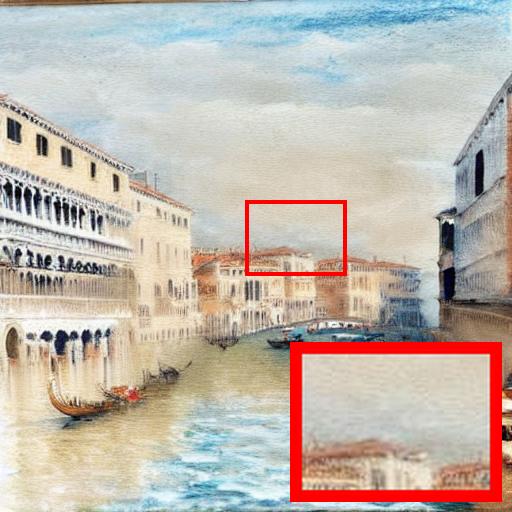} & 
\includegraphics[width=0.14\linewidth]{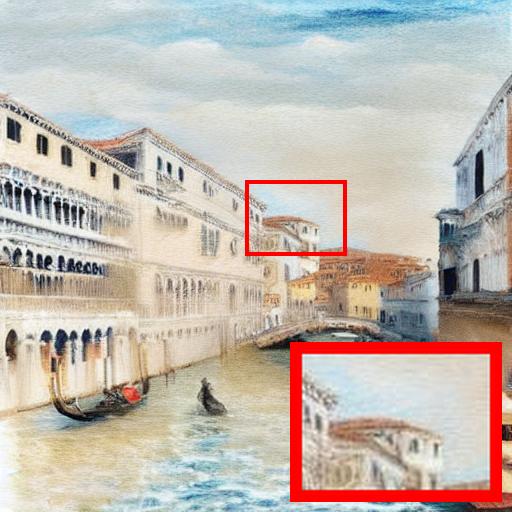} 

\end{tabular}
}

\caption{Qualitative results showing the effect of parameter $\alpha_{data}$ for style transfer.}
\label{fig:sm3}
\end{figure}

\begin{figure}[t]
\centering

\resizebox{0.98\textwidth}{!}{

\setlength{\tabcolsep}{0.05cm} % 调整列间距
\renewcommand{\arraystretch}{0.5}  % 调整行距
\begin{tabular}{cccccc}
Text & Style & $\gamma_{data}=0$ & $\gamma_{data}=0.2$ & $\gamma_{data}=0.4$ & $\gamma_{data}=1$ \\
\includegraphics[width=0.14\linewidth]{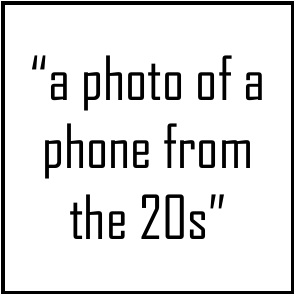} & 
\includegraphics[width=0.14\linewidth]{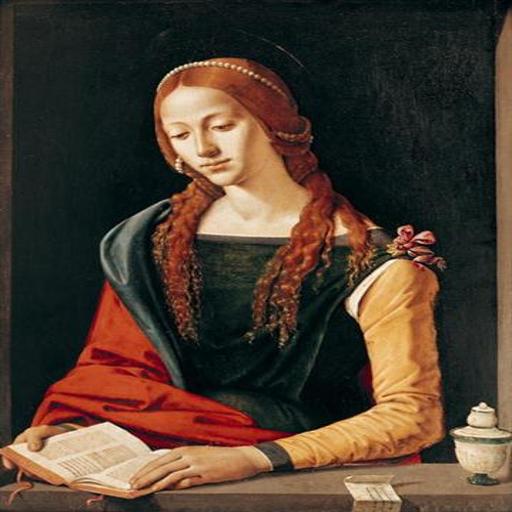} & 
\includegraphics[width=0.14\linewidth]{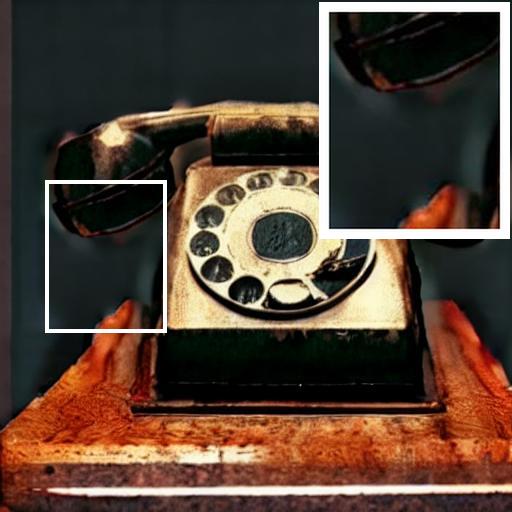} & 
\includegraphics[width=0.14\linewidth]{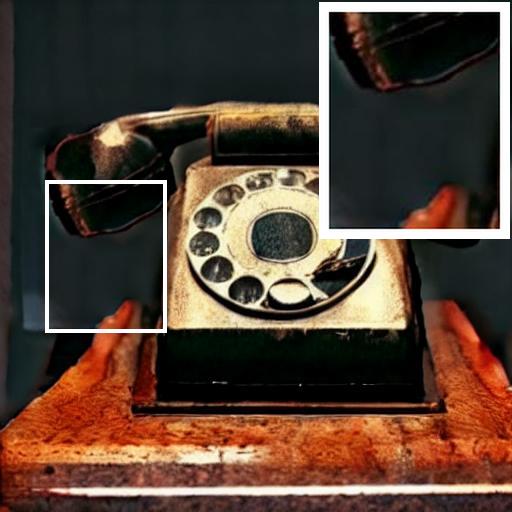} & 
\includegraphics[width=0.14\linewidth]{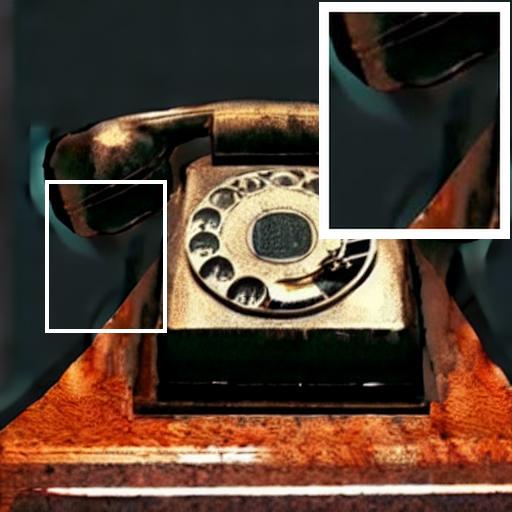} & 
\includegraphics[width=0.14\linewidth]{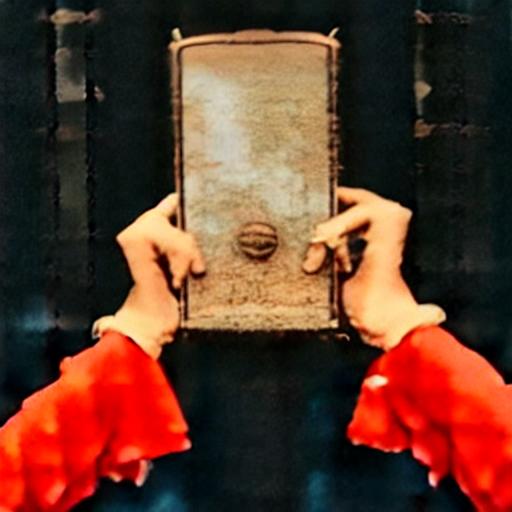} 

\end{tabular}
}

\caption{Qualitative results showing the effect of parameter $\gamma_{data}$ for style transfer.}
\label{fig:sm4}
\end{figure}

\begin{figure}[t]
\centering

\resizebox{0.98\textwidth}{!}{

\setlength{\tabcolsep}{0.05cm} % 调整列间距
\renewcommand{\arraystretch}{0.5}  % 调整行距
\begin{tabular}{cccccc}
Text & Style & $\alpha_{margin}=0$ & $\alpha_{margin}=1$ & $\alpha_{margin}=3$ & $\alpha_{margin}=4$ \\
\includegraphics[width=0.14\linewidth]{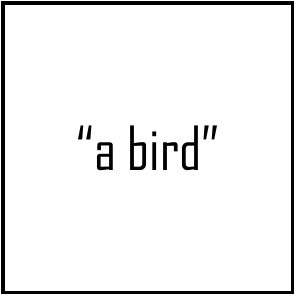} & 
\includegraphics[width=0.14\linewidth]{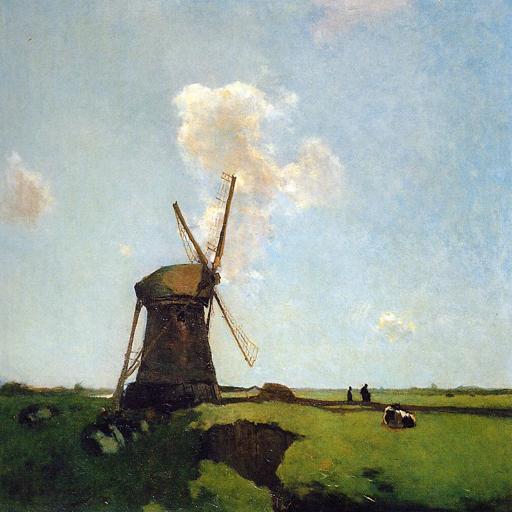} & 
\includegraphics[width=0.14\linewidth]{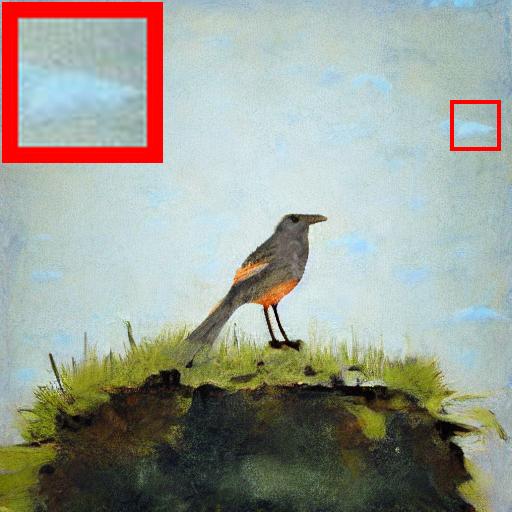} & 
\includegraphics[width=0.14\linewidth]{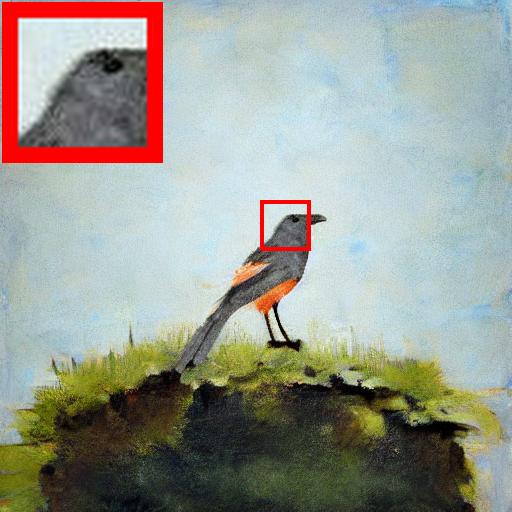} & 
\includegraphics[width=0.14\linewidth]{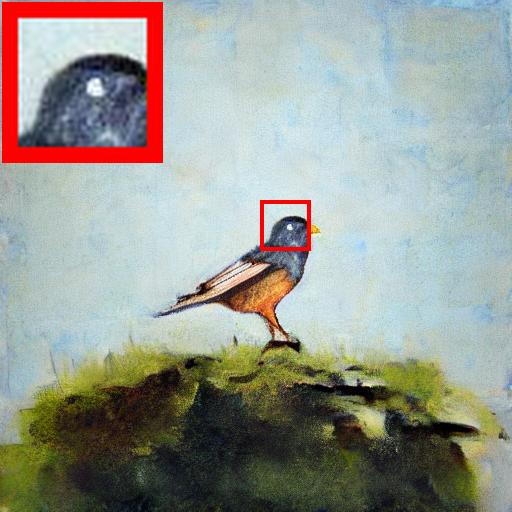} & 
\includegraphics[width=0.14\linewidth]{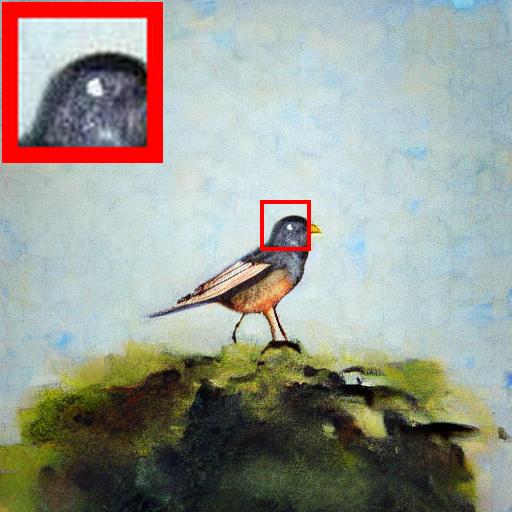} 

\end{tabular}
}

\caption{Qualitative results showing the effect of parameter $\alpha_{margin}$ for style transfer.}
\label{fig:sm5}
\end{figure}

$\boldsymbol{\alpha_{data}}$: The parameter governs the degree of mixing between the data condition and the current timestep's features. The selection of $\alpha_{data}$ depends on the specific task and the reliability of the information provided by the conditioned data. Specifically, for style transfer tasks, our focus is on the style of the image rather than its semantic content. Consequently, we increase $\alpha_{data}$ to suppress the influence of the reference image's content, thereby mitigating the risk of content leakage. In contrast, for super-resolution and deblurring, the choice depends on the quality of the processed labels: a lower $\alpha_{data}$ allows the model to learn richer semantics and fine details when high-quality labels are available. However, if the labels are of poor quality, a low mixing degree may lead to the injection of detrimental noise, ultimately compromising the fidelity of the generated output. Thus, selecting $\alpha_{data}$ is a critical balancing act to ensure that the model internalizes relevant information while remaining resilient to noise. As illustrated in Fig.~\ref{fig:sm3}, the impact of this parameter is evident in style transfer. Consistent with our analysis, a small $\alpha_{data}$ leads to visible content leakage, where elements from the reference image (e.g., the beach) appear in the generated output. As $\alpha_{data}$ increases, content leakage is effectively suppressed. However, when $\alpha_{data}=1$ is excessively high, the influence of the condition diminishes, resulting in suboptimal style transfer—observed as stylistic discontinuities (e.g., a difference in color between the sky adjacent to the building and the surrounding sky). Therefore, we determined $\alpha_{data}=0.88$ to be the optimal value, effectively balancing the content and style of the generated image.

$\boldsymbol{\gamma_{data}}$: This parameter regulates the guidance scale of the data condition, representing the weight of the condition-injected noise relative to the unconditional noise. This mechanism is analogous to classifier-free guidance in conditional diffusion models; however, rather than using text-based prompts, we employ noise predicted with and without a specific data condition to guide the generative process. The optimal value of $\gamma_{data}$ depends on the specific task requirements and the quality of the preprocessed labels. For style transfer, where the primary objective is to transfer stylistic attributes while preserving the original semantic content, $\gamma_{data}$ must be carefully constrained. If the value is too large, it risks introducing unintended content leakage from the style reference image. Conversely, for super-resolution and deblurring, higher-quality preprocessed condition images allow for a stronger conditioning signal, necessitating a larger $\gamma_{data}$ to enhance fidelity and detail reconstruction. Based on these theoretical considerations and empirical observations, we carefully calibrated $\gamma_{data}$ for each task. As illustrated in Fig.~\ref{fig:sm4}, the impact of $\gamma_{data}$ on style transfer is significant. When $\gamma_{data}$ is too small (e.g., $\gamma_{data}=0$), the stylization is insufficient, resulting in artifacts such as discontinuous edges. As $\gamma_{data}$ increases (e.g., $\gamma_{data}=1$), the model suffers from content leakage from the reference image. Consequently, we determined $\gamma_{data}=0.2$ to be the optimal value, striking a balance between achieving the desired degree of stylization and maintaining the integrity of the original content.

$\boldsymbol{\alpha_{margin}}$: This parameter controls the maximum permissible reduction in the loss between the predicted data and the conditional image at the current timestep relative to the previous one. The selection of $\alpha_{margin}$ is directly related to the required intensity of the optimization constraint. Specifically, a larger $\alpha_{margin}$ imposes a stricter requirement, forcing the current loss to be significantly lower than that of the previous timestep, thereby exerting stronger pressure on the optimization process. As shown in Fig.~\ref{fig:sm5}, the value of $\alpha_{margin}$ directly impacts the generation quality. When $\alpha_{margin}$ is too small (e.g., $\alpha_{margin}=0$), the optimization is insufficient, leading to the emergence of visual artifacts. Conversely, when $\alpha_{margin}$ is excessively large (e.g., $\alpha_{margin}=3$ or $4$), the optimization pressure becomes too intense, driving the data distribution into unrealistic regions and resulting in unnatural features (e.g., the white eyes shown in the figure). Based on these theoretical considerations and empirical observations, we have determined the optimal value for $\alpha_{margin}$ to ensure stable and high-quality performance across all three tasks.

\begin{algorithm*}[t]
\caption{ \small {\bf DICT}: Data Injection and Contrastive Trajectory Refinement for Conditional Image Generation with Diffusion Models}
\label{a:1}

\begin{algorithmic}[1]
    \State \textbf{Input:}  Diffusion model $\epsilon_\theta$, encoder $E$, decoder $D$, condition $x$, condition operation $M$, evaluation function $f_{loss}$, task condition $c_{task}$

    \State $\hat{x}=M(x)$, $\hat{z} = E(\hat{x})$

    \State $ 
        z_{T} = \sqrt{\bar{\alpha}_T}\hat{z} + \sqrt{1-\bar{\alpha}_T}\epsilon$,  where $\epsilon \sim \mathcal{N}(0, \mathbf{I})$
    \For {$t=T, \cdots, T - T_1$}
        \For{$i=1,\cdots, N_{iter1}$}
        \State $\hat{c}_{t} = \sqrt{\bar{\alpha}_t}\hat{z} + \sqrt{1-\bar{\alpha}_t}\epsilon$,   where $\epsilon \sim \mathcal{N}(0, \mathbf{I})$ when $i=1$

        \State $c_t  = \alpha_{data} \times z_t + (1-\alpha_{data}) \times \hat{c}_{t}$

        \State $\hat{\epsilon}_\theta(z_t, c_t, c_{task})  = \gamma_{data}   \times \epsilon_\theta(c_{t}, c_{task}) 
    + (1 - \gamma_{data} ) \times \epsilon_\theta(z_t, c_{task})$
        
        \State $\epsilon_\theta(t)  = \hat{\epsilon}_\theta(z_t, \hat{c}_t, c_{task})$

        \State $z_{0|t} = (z_t - \sqrt{1 - \bar{\alpha}_t} \, \epsilon_\theta(t))/{\sqrt{\bar{\alpha}_t}}$

        \State $z_{0|t} = z_{0|t}  - \eta_1 \times \nabla_{z_{0|t}}\mathcal{L}_{1}(z_{0|t},x)$, where $\mathcal{L}_{1}(z_{0|t},x)  =  f_{loss}(D(z_{0|t}), x)$ 
        
        \State $z_t = z_{0|t}  \sqrt{\bar{\alpha}_t} + \sqrt{1 - \bar{\alpha}_t}\epsilon_\theta(t) $, $\epsilon = \epsilon_\theta(t)$

        \EndFor

        \State $z_{t-1} = \sqrt{\bar{\alpha}_{t-1}} z_{0|t} + \sqrt{1 - \bar{\alpha}_{t-1}}\epsilon_\theta(t)$

        \For{$j=1, \cdots, N_{iter2}$}

        \State $\hat{c}_{t-1} = \sqrt{\bar{\alpha}_{t-1}}\hat{z} + \sqrt{1-\bar{\alpha}_{t-1}}\epsilon$,   where $\epsilon \sim \mathcal{N}(0, \mathbf{I})$ when $j=1$

        \State $c_{t-1}  = \alpha_{data} \times z_{t-1} + (1-\alpha_{data}) \times \hat{c}_{t-1}$

        \State $\hat{\epsilon}_\theta(z_{t-1}, c_{t-1}, c_{task})  = \gamma_{data}   \times \epsilon_\theta(c_{t-1}, c_{task}) 
    + (1 - \gamma_{data} ) \times \epsilon_\theta(z_{t-1}, c_{task})$
        
        \State $\epsilon_\theta(t-1)  = \hat{\epsilon}_\theta(z_{t-1}, \hat{c}_{t-1}, c_{task})$

        \State $z_{0|t-1} = (z_{t-1} - \sqrt{1 - \bar{\alpha}_{t-1}} \, \epsilon_\theta({t-1}))/{\sqrt{\bar{\alpha}_{t-1}}}$

        \State $\mathcal{L}_{2} = \max \big( \mathcal{L}_{1}(z_{0|t-1},x) - \mathcal{L}_{1}(z_{0|t},x) + \alpha_{margin}, 0 \big)$

        \State $z_{0|{t-1}} = z_{0|{t-1}}  - \eta_2 \times \nabla_{z_{0|{t-1}}}\mathcal{L}_{2}$
        
        \State $z_{t-1} = z_{0|{t-1}}  \sqrt{\bar{\alpha}_{t-1}} + \sqrt{1 - \bar{\alpha}_{t-1}}\epsilon_\theta(t) $, $\epsilon = \epsilon_\theta({t-1})$
        \EndFor
    \EndFor

    \For {$t=T - T_1-1, \cdots, 1$}
        \For{$i=1,\cdots, N_{iter1}$}

        \State $\epsilon_\theta(t)  = \hat{\epsilon}_\theta(z_t, c_{task})$, 

        \State $z_{0|t} = (z_t - \sqrt{1 - \bar{\alpha}_t} \, \epsilon_\theta(t))/{\sqrt{\bar{\alpha}_t}}$

        \State $z_{0|t} = z_{0|t}  - \eta_1 \times \nabla_{z_{0|t}}\mathcal{L}_{1}(z_{0|t},x)$, where $\mathcal{L}_{1}(z_{0|t},x)  =  f_{loss}(D(z_{0|t}), x)$ 
        
        \State $z_t = z_{0|t}  \sqrt{\bar{\alpha}_t} + \sqrt{1 - \bar{\alpha}_t}\epsilon_\theta(t) $

        \EndFor

        \State $z_{t-1} = \sqrt{\bar{\alpha}_{t-1}} z_{0|t} + \sqrt{1 - \bar{\alpha}_{t-1}}\epsilon_\theta(t)$

        \For{$j=1, \cdots, N_{iter2}$}

        \State $\epsilon_\theta(t-1)  = \hat{\epsilon}_\theta(z_{t-1}, c_{task})$,

        \State $z_{0|t-1} = (z_{t-1} - \sqrt{1 - \bar{\alpha}_{t-1}} \, \epsilon_\theta({t-1}))/{\sqrt{\bar{\alpha}_{t-1}}}$

        \State $\mathcal{L}_{2} = \max \big( \mathcal{L}_{1}(z_{0|t-1},x) - \mathcal{L}_{1}(z_{0|t},x) + \alpha_{margin}, 0 \big)$

        \State $z_{0|{t-1}} = z_{0|{t-1}}  - \eta_2 \times \nabla_{z_{0|{t-1}}}\mathcal{L}_{2}$
        
        \State $z_{t-1} = z_{0|{t-1}}  \sqrt{\bar{\alpha}_{t-1}} + \sqrt{1 - \bar{\alpha}_{t-1}}\epsilon_\theta(t) $
        \EndFor
    \EndFor

    \State $x_0 = D(z_0)$

    \State \textbf{Output:} Sample $y=x_0$

\end{algorithmic}

\end{algorithm*}

\begin{figure}[t]
\centering
\resizebox{1\textwidth}{!}{
\setlength{\tabcolsep}{0.000cm} 
\renewcommand{\arraystretch}{0.00}  
\begin{tabular}{>{\centering\arraybackslash}m{1.0cm} 
 >{\centering\arraybackslash}m{2cm} >{\centering\arraybackslash}m{2cm} >{\centering\arraybackslash}m{1cm} >{\centering\arraybackslash}m{2cm} >{\centering\arraybackslash}m{2cm} >{\centering\arraybackslash}m{1cm} >{\centering\arraybackslash}m{2cm} >{\centering\arraybackslash}m{2cm}}
 \\
  Inputs & PixArt & DICT & Inputs  &  PixArt &  DICT & Inputs  &  PixArt &  DICT
\\
\includegraphics[width=1\linewidth]{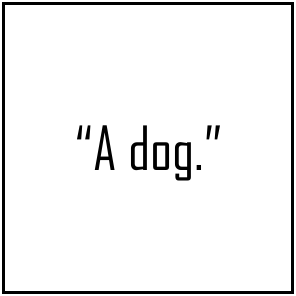} 
\includegraphics[width=1\linewidth]{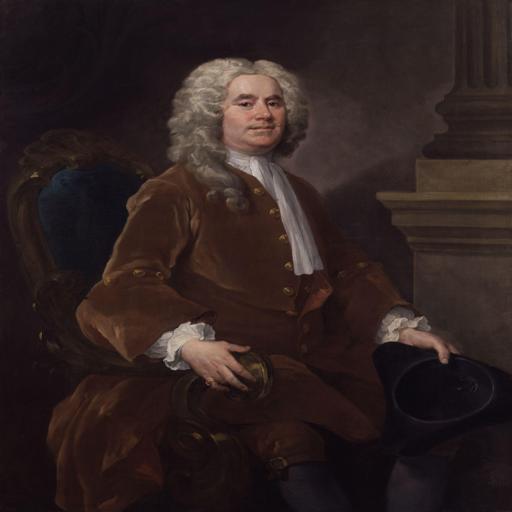}  & 
\includegraphics[width=1\linewidth]{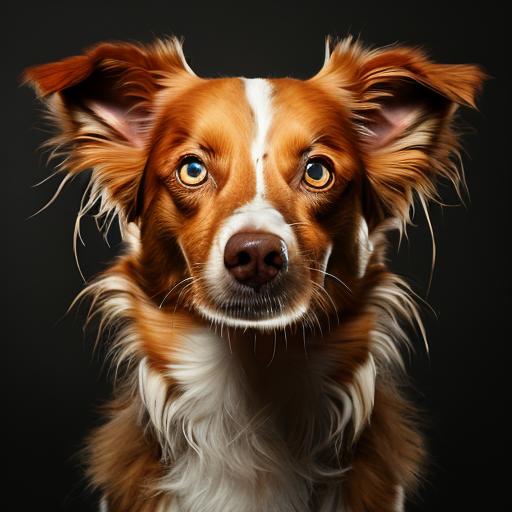}  &
\includegraphics[width=1\linewidth]{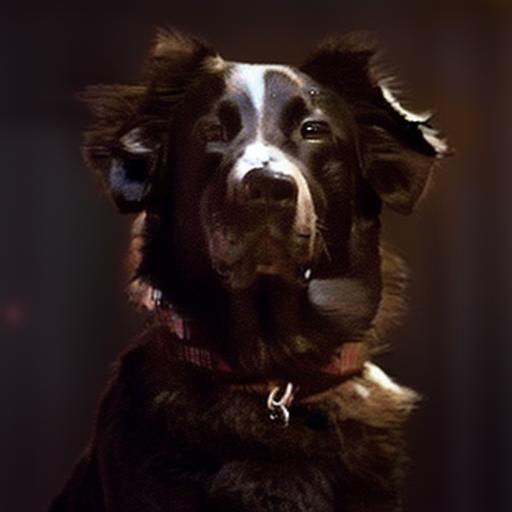}  & 
\includegraphics[width=1\linewidth]{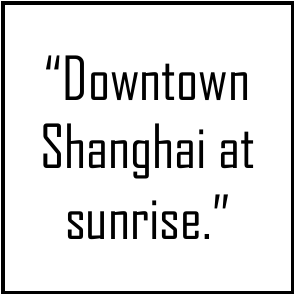} 
\includegraphics[width=1\linewidth]{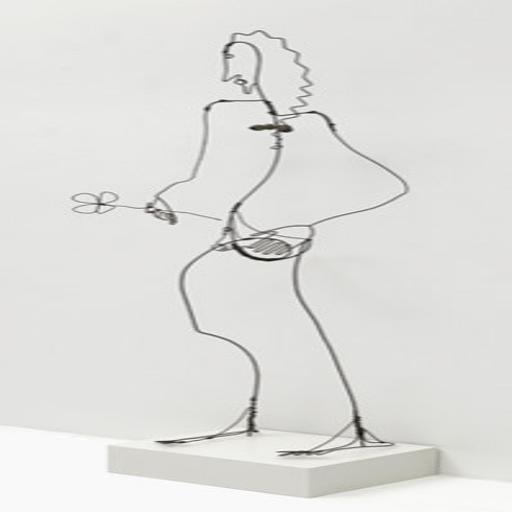} &
\includegraphics[width=1\linewidth]{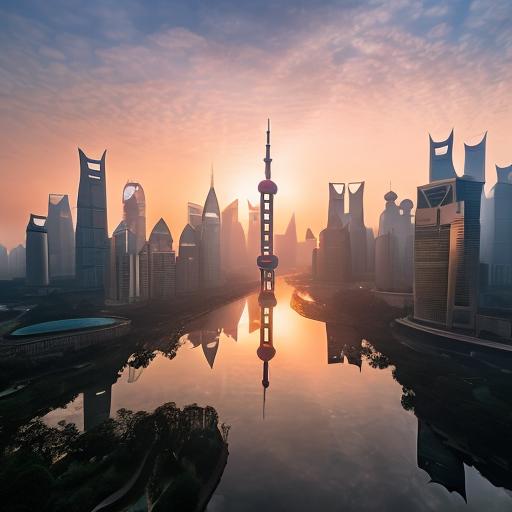} &  
\includegraphics[width=1\linewidth]{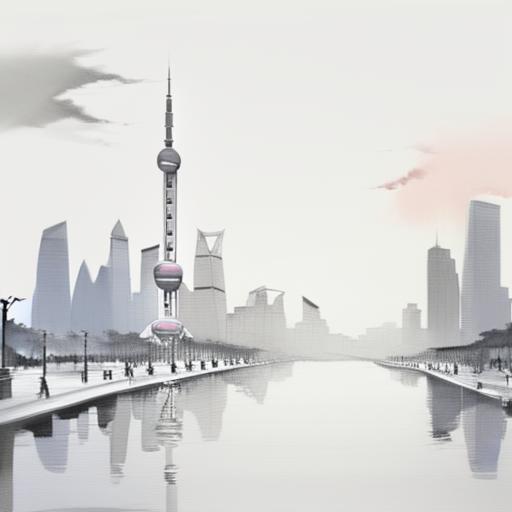} &
\includegraphics[width=1\linewidth]{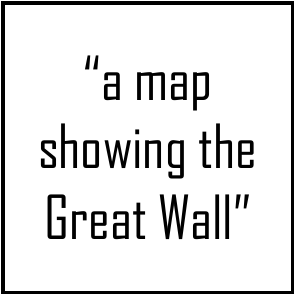} 
\includegraphics[width=1\linewidth]{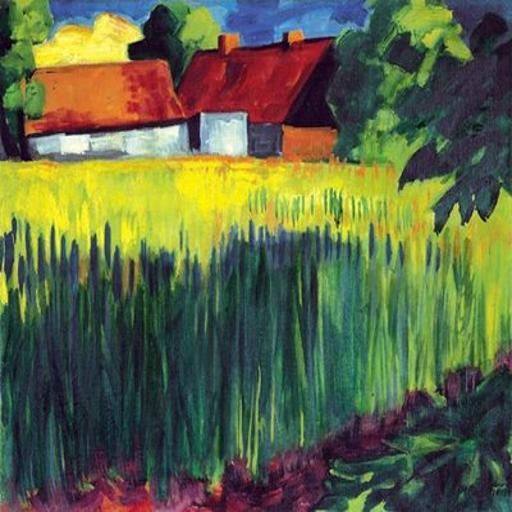} &
\includegraphics[width=1\linewidth]{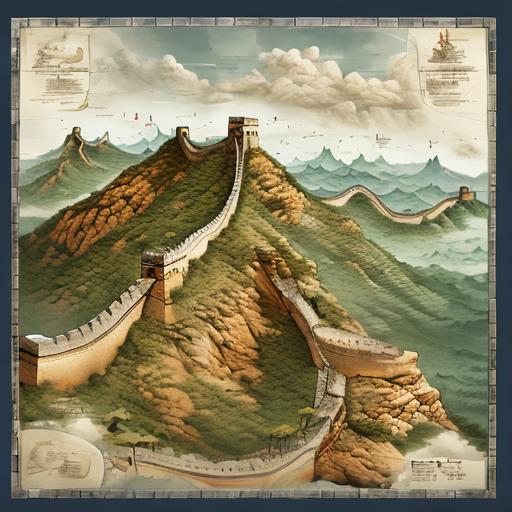} &  
\includegraphics[width=1\linewidth]{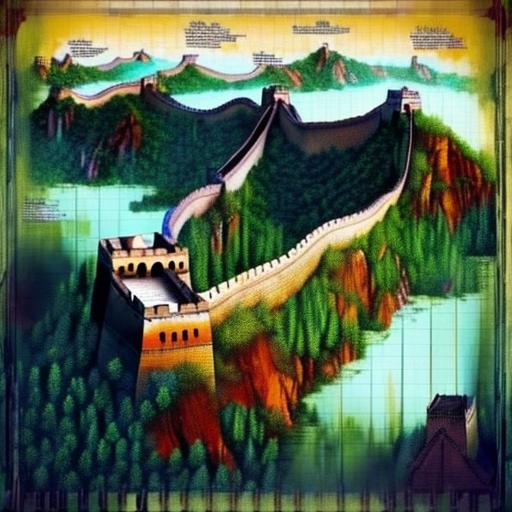}

\end{tabular}}

\caption{Qualitative DiT-based stylization results.}
\label{fig:suppldit}

\end{figure}

\begin{figure*}[t]
\centering
\resizebox{0.65\textwidth}{!}{
\setlength{\tabcolsep}{0.15cm} 
\renewcommand{\arraystretch}{0.00}  
\begin{tabular}{
>{\centering\arraybackslash}m{2.2cm} 
>{\centering\arraybackslash}m{2.2cm} 
>{\centering\arraybackslash}m{2.2cm} 
>{\centering\arraybackslash}m{2.2cm}
>{\centering\arraybackslash}m{2.2cm}}

% ------------------- 任务1：风格迁移 -------------------
Text  & Style & (1) DICT & (2) w/o DICT & (3) SDEdit \\

\includegraphics[width=1\linewidth]{rebuttal/fig/A_dog.pdf} &
\includegraphics[width=1\linewidth]{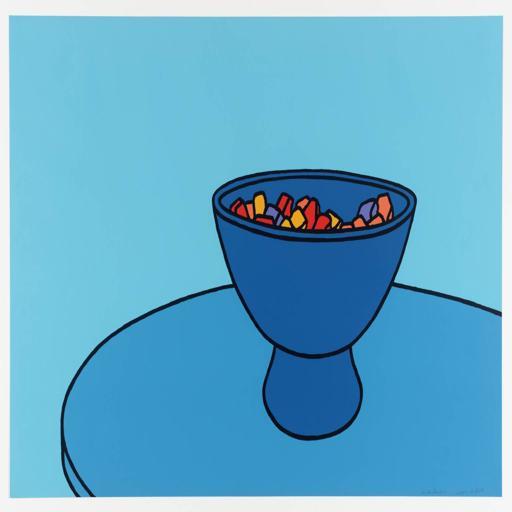}  &
\includegraphics[width=1\linewidth]{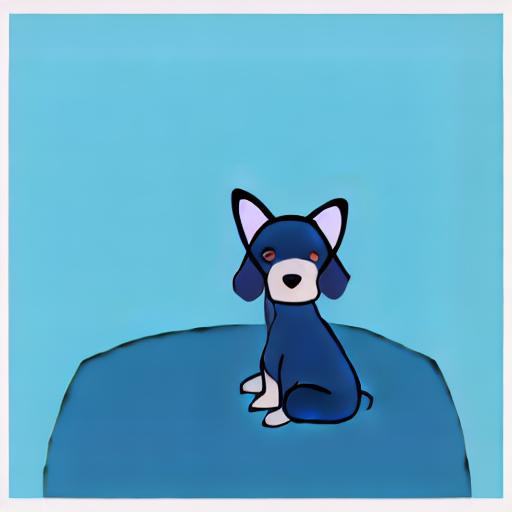}  &
\includegraphics[width=1\linewidth]{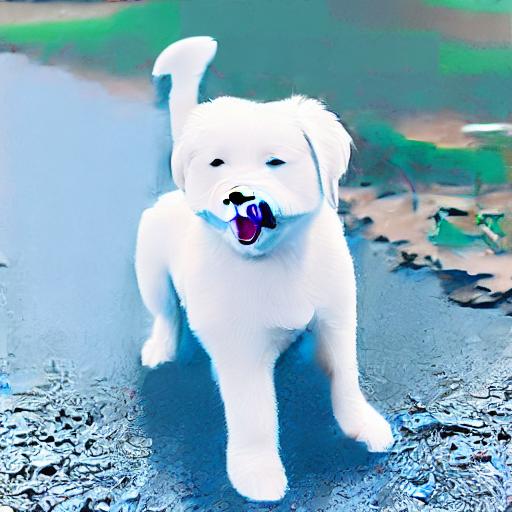}  & 
\includegraphics[width=1\linewidth]{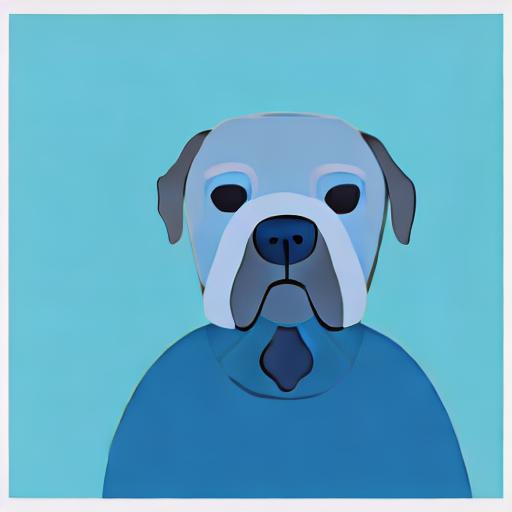}  \\
\noalign{\vspace{0.1cm}}
\multicolumn{2}{c}{Text Score $\uparrow$} & 0.264 & {\bf 0.270} & 0.258 \\
\noalign{\vspace{0.1cm}}\\
\multicolumn{2}{c}{Style Loss $\downarrow$} & {\bf 0.155} & 1.089 & 0.198 \\
\noalign{\vspace{0.1cm}}\\
\multicolumn{2}{c}{CLIP Loss $\downarrow$} & {\bf 2.306} & 8.220 & 2.617 \\
\noalign{\vspace{0.15cm}}
\multicolumn{5}{c}{\large (a) Comparisons of text-to-image style transfer task.} \\
\noalign{\vspace{0.35cm}}

% ------------------- 任务2：图像超分 -------------------
Condition & Target & (1) DICT & (2) w/o DICT & (3) SDEdit \\

\includegraphics[width=1\linewidth]{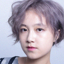} &
\includegraphics[width=1\linewidth]{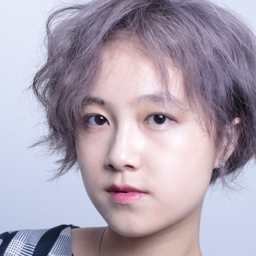} &
\includegraphics[width=1\linewidth]{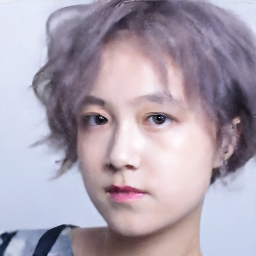} &  
\includegraphics[width=1\linewidth]{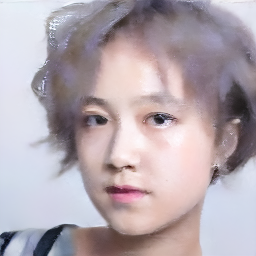} &
\includegraphics[width=1\linewidth]{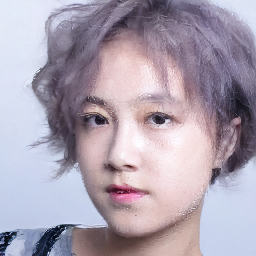} \\
\noalign{\vspace{0.1cm}}
\multicolumn{2}{c}{PSNR Score $\uparrow$} & {\bf 29.168} & 25.853 & 26.668 \\
\noalign{\vspace{0.1cm}}\\
\multicolumn{2}{c}{SSIM Score $\uparrow$} & {\bf 0.823} & 0.751 & 0.734 \\
\noalign{\vspace{0.1cm}}\\
\multicolumn{2}{c}{LPIPS Loss $\downarrow$} & {\bf 0.157} & 0.246 & 0.198 \\
\noalign{\vspace{0.15cm}}
\multicolumn{5}{c}{\large (b) Comparisons of image super-resolution task.} \\
\noalign{\vspace{0.35cm}} 

% ------------------- 任务3：图像去模糊 -------------------
Condition & Target & (1) DICT & (2) w/o DICT & (3) SDEdit \\

\includegraphics[width=1\linewidth]{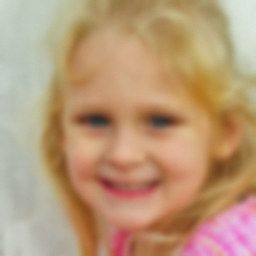} &
\includegraphics[width=1\linewidth]{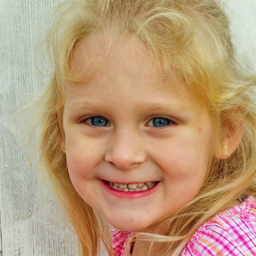} & 
\includegraphics[width=1\linewidth]{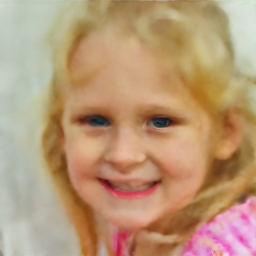} &  
\includegraphics[width=1\linewidth]{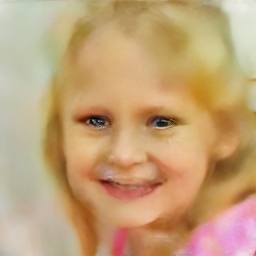} &
\includegraphics[width=1\linewidth]{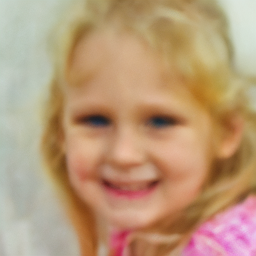} \\
\noalign{\vspace{0.1cm}}
\multicolumn{2}{c}{PSNR Score $\uparrow$} & {\bf 27.226} & 24.204 & 26.760 \\
\noalign{\vspace{0.1cm}}\\
\multicolumn{2}{c}{SSIM Score $\uparrow$} & {\bf 0.723} & 0.661 & 0.707 \\
\noalign{\vspace{0.1cm}}\\
\multicolumn{2}{c}{LPIPS Loss $\downarrow$} & {\bf 0.330} & 0.455 & 0.368 \\
\noalign{\vspace{0.15cm}}
\multicolumn{5}{c}{\large (c) Comparisons of image deblurring task.} \\

\end{tabular}}

\caption{Comprehensive comparisons between DICT, the base model (w/o DICT), and SDEdit. Both visual visualizations and numerical metrics are provided for (a) style transfer, (b) super-resolution, and (c) deblurring.}
\label{fig:ab}
\end{figure*}

\section{More Experiments}

\subsection{Generalization to Diffusion Transformers}

To demonstrate the backbone-agnostic nature of our proposed method, we extend our evaluation to the Diffusion Transformer (DiT) architecture. Specifically, we adopt text-driven style transfer as a representative task. For this experiment, we implement our approach on top of the PixArt~\cite{chen2024pixart}. As illustrated in Fig.~\ref{fig:suppldit}, our method consistently achieves high-quality stylization on the DiT backbone. These results provide strong empirical evidence that DICT generalizes well and is fundamentally independent of the underlying network architecture.

\subsection{More Ablation Studies and Comparisons}

To provide a more comprehensive validation of our framework, we present further ablations and comparative analyses in this section. Using text-driven style transfer as our testbed, we investigate the generation quality of DICT relative to the base model, alongside a direct comparison with the SDEdit method. Specifically, the ablation results comparing DICT against the base model are illustrated in Fig.~\ref{fig:ab} (1) and (2), while the comparative results with SDEdit are presented in Fig.~\ref{fig:ab} (1) and (3). The ensuing qualitative results strongly substantiate the underlying necessity of our core designs and clearly demonstrate our superiority

\section{Future Work}

\subsection{Adaptive Parameter Selection Strategy}

Based on our analysis of the parameter selection strategies in Sec.~\ref{sec:Parameter Selection Strategy}, we find that the choice of parameters is closely tied to the data distribution during optimization, the stability of the optimization process, and the specific task. Given that the entire optimization is driven by the loss between the predicted data and the label data, we can leverage this core mechanism. This insight leads us to propose a loss-adaptive parameter selection strategy. This strategy will dynamically adjust relevant parameters based on the changing loss between the predicted and label data. By adopting this approach, we will achieve optimal parameter configurations, leading to the generation of higher-quality and more accurate results. This method transforms parameter selection from a static, manual decision based on human experience into a dynamic, intelligent process guided by real-time loss feedback,  enhancing the model's performance and generalization ability.

\subsection{Accelerated Inference}

Our method involves certain optimization loops that are inherently determined by the iterative optimization process and the nature of the diffusion model in Alg.~\ref{a:1}. Firstly, when updating the prediction data using the loss between the predicted and label data, the step size must not be too large, to avoid exceeding the model's effective data distribution, which could lead to artificial artifacts. Therefore, in order to achieve optimal optimization results, a progressive loop-based optimization approach is necessary. In addition, during the reverse diffusion process, each prediction at every time step needs to be re-noised before entering the next time step for further computation and optimization. This repetitive process of adding noise can disturb the already partially optimized data, causing a loss in the preservation of the relevant features of the label data. As a result, re-optimization at each time step is required. These factors inevitably affect the speed of the optimization process to some extent. Future research will focus on developing more efficient optimization strategies and improved denoising sampling methods to enhance performance.

\section{DICT Algorithm}
\label{sec:algorithm}
To offer a clearer understanding of our approach, we provide a detailed algorithm in Alg.~\ref{a:1}. This algorithm formally outlines the operational flow of both the data injection and the contrastive trajectory refinement modules.

\section{More Qualitative Results}
We provide more qualitative results of our method in this section for deeper visual analysis and understanding. These qualitative results supplement the previously reported quantitative evaluation data, aiming to show the performance and effect of our model.

\begin{figure}[t]
\centering

\resizebox{0.98\textwidth}{!}{

\setlength{\tabcolsep}{0.05cm} % 调整列间距
\renewcommand{\arraystretch}{0.5}  % 调整行距
\begin{tabular}{ccccccc}
\includegraphics[width=0.14\linewidth]{st_images/A_dog.pdf} & 
\includegraphics[width=0.14\linewidth]{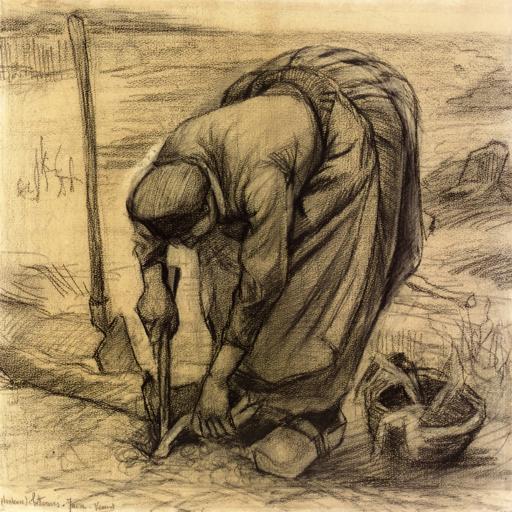} & \includegraphics[width=0.14\linewidth]{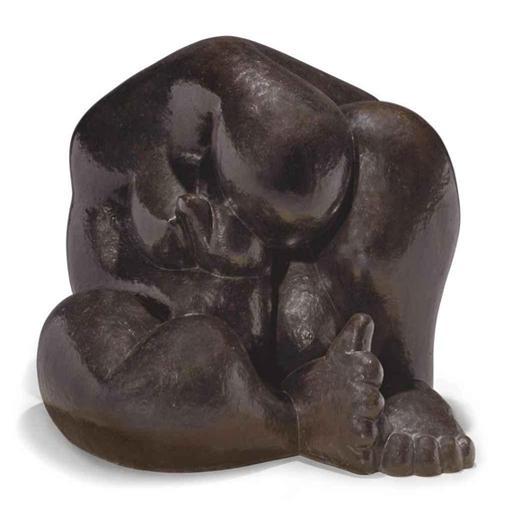}  & \includegraphics[width=0.14\linewidth]{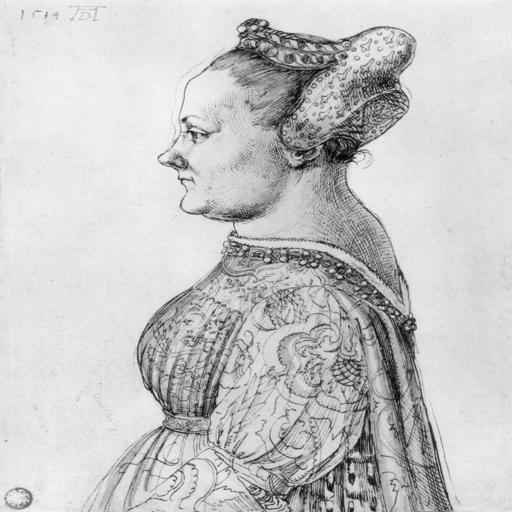} &
\includegraphics[width=0.14\linewidth]{appendix_images/st/1_43.jpg} & \includegraphics[width=0.14\linewidth]{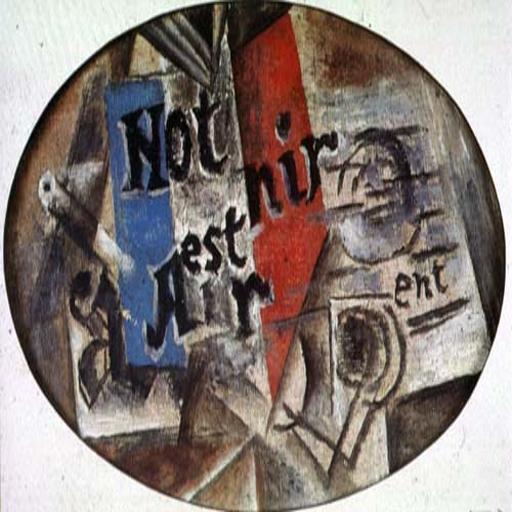} \\ &
\includegraphics[width=0.14\linewidth]{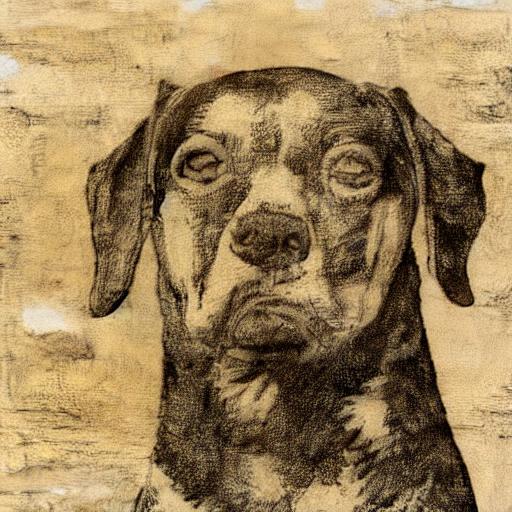} & \includegraphics[width=0.14\linewidth]{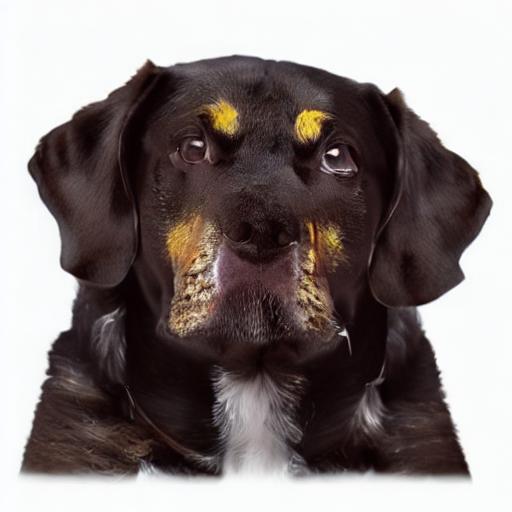}  & \includegraphics[width=0.14\linewidth]{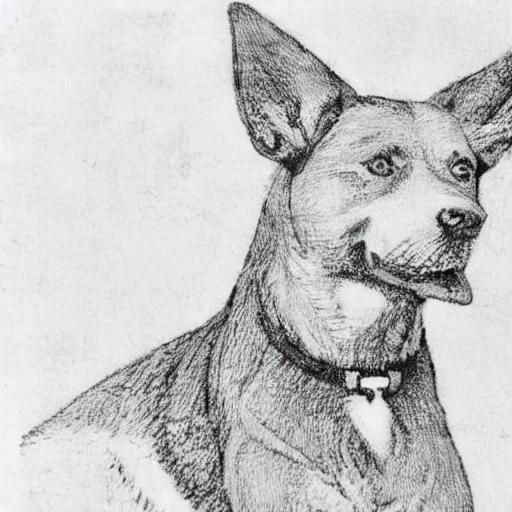} &
\includegraphics[width=0.14\linewidth]{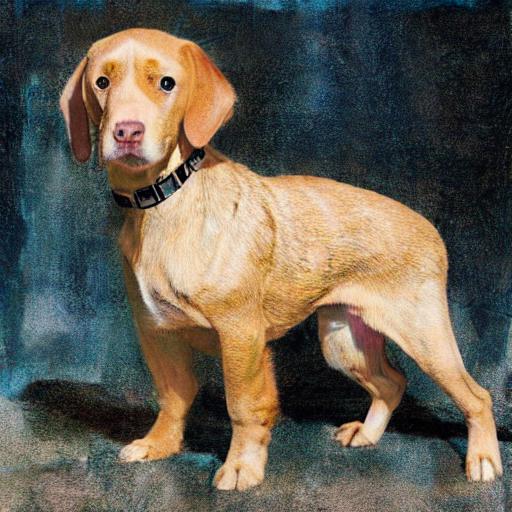} & \includegraphics[width=0.14\linewidth]{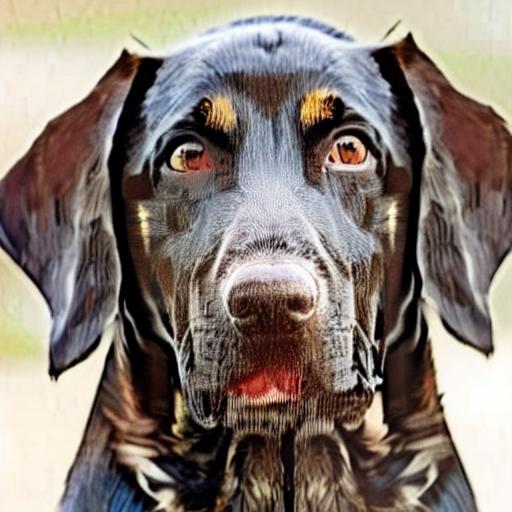} \\
\includegraphics[width=0.14\linewidth]{appendix_images/st/a_car.pdf} & 
\includegraphics[width=0.14\linewidth]{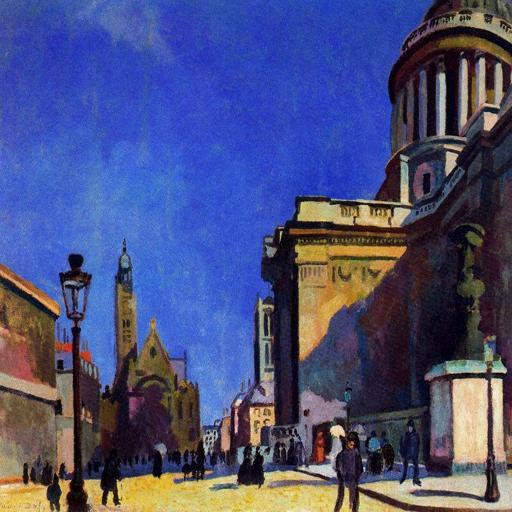} & \includegraphics[width=0.14\linewidth]{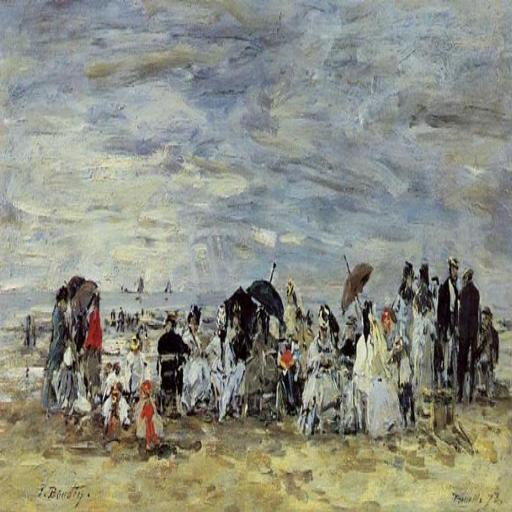}  & \includegraphics[width=0.14\linewidth]{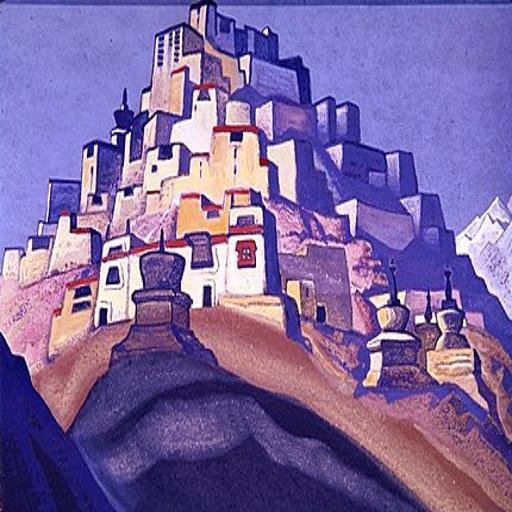} &
\includegraphics[width=0.14\linewidth]{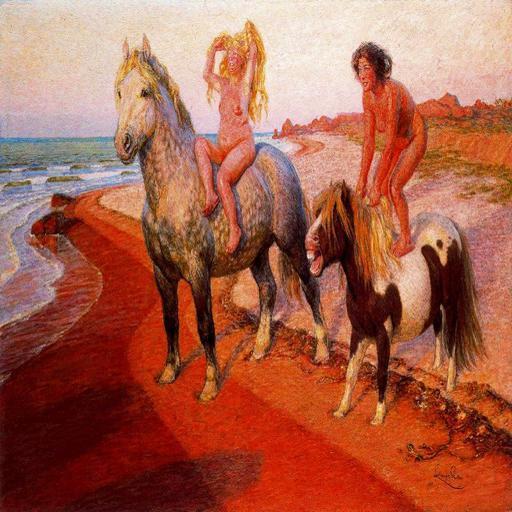} & \includegraphics[width=0.14\linewidth]{appendix_images/st/59_42.jpg} \\
& 
\includegraphics[width=0.14\linewidth]{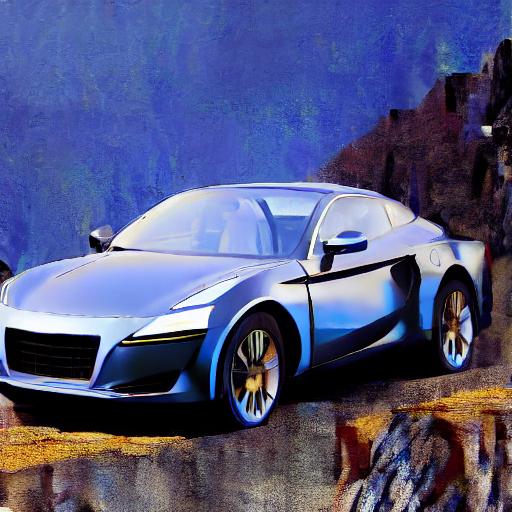} & \includegraphics[width=0.14\linewidth]{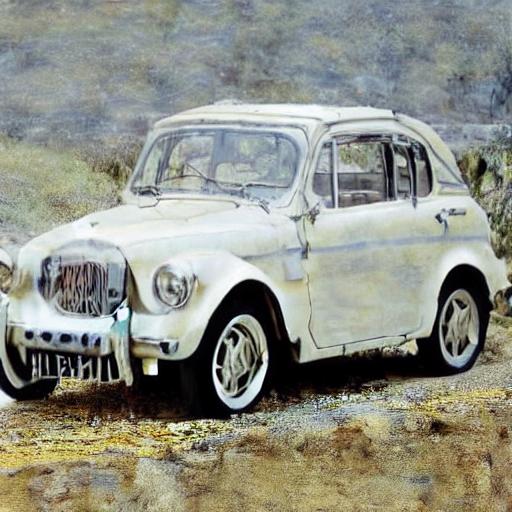}  & \includegraphics[width=0.14\linewidth]{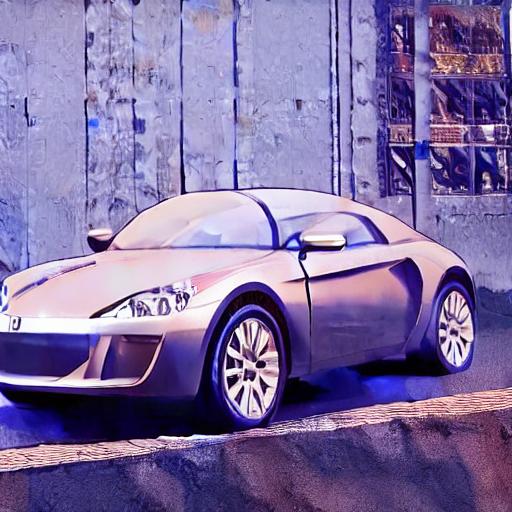} &
\includegraphics[width=0.14\linewidth]{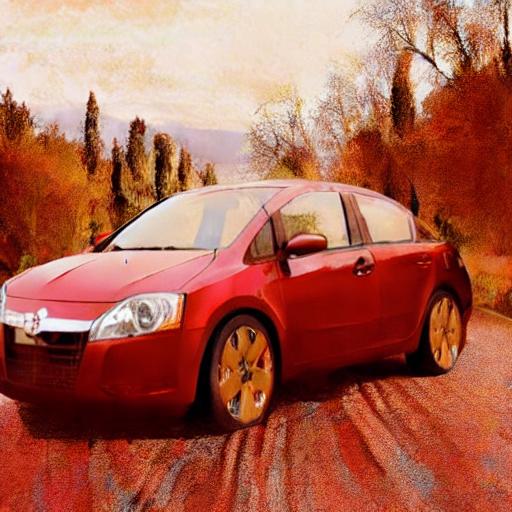} & \includegraphics[width=0.14\linewidth]{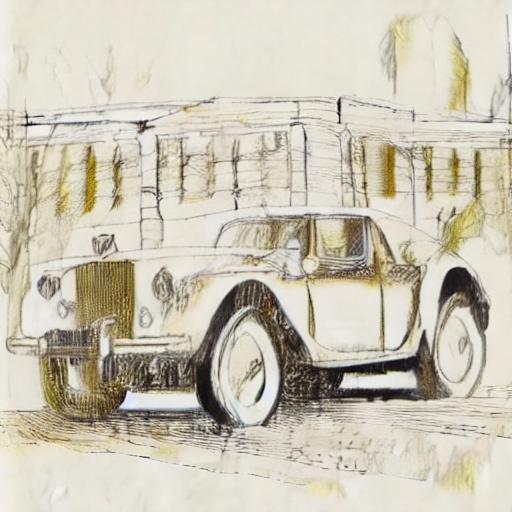} \\
\includegraphics[width=0.14\linewidth]{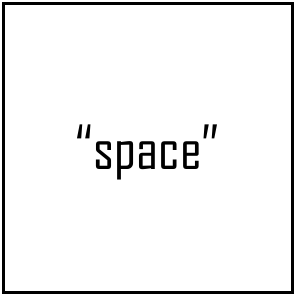} & 
\includegraphics[width=0.14\linewidth]{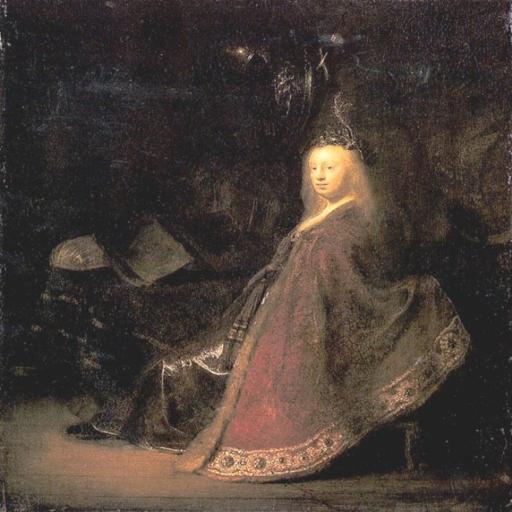} & \includegraphics[width=0.14\linewidth]{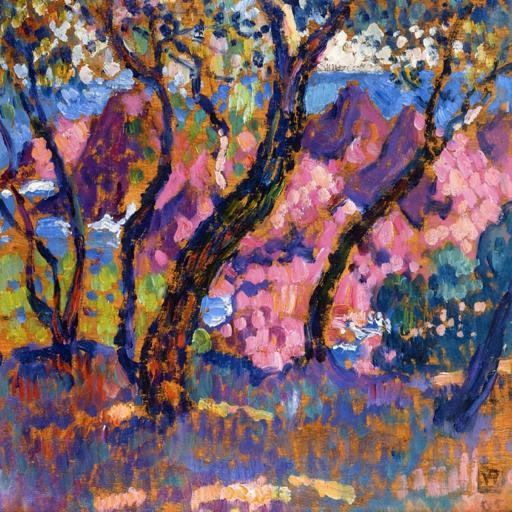}  & \includegraphics[width=0.14\linewidth]{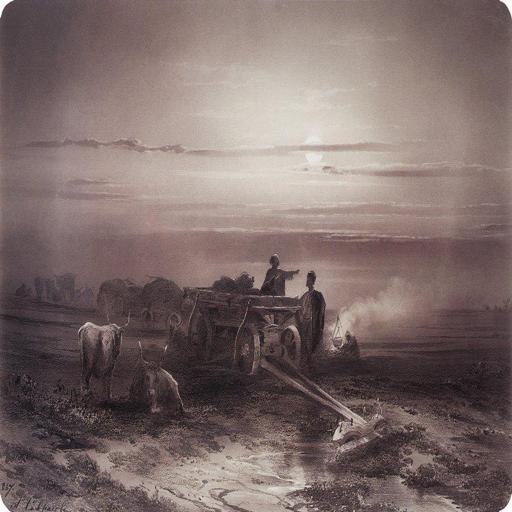} &
\includegraphics[width=0.14\linewidth]{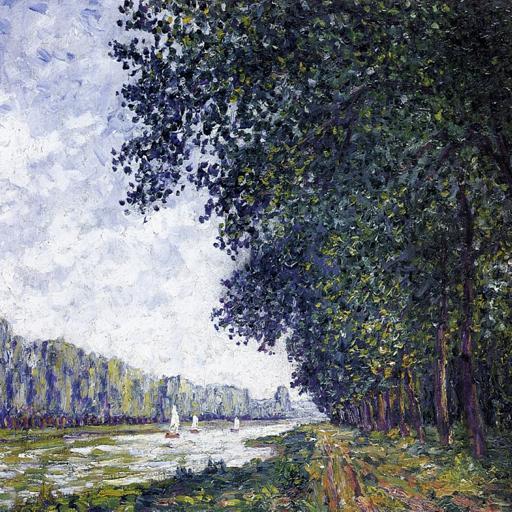} & \includegraphics[width=0.14\linewidth]{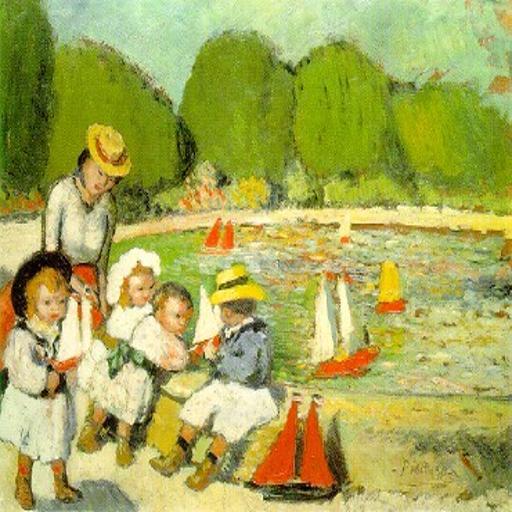} \\
 & 
\includegraphics[width=0.14\linewidth]{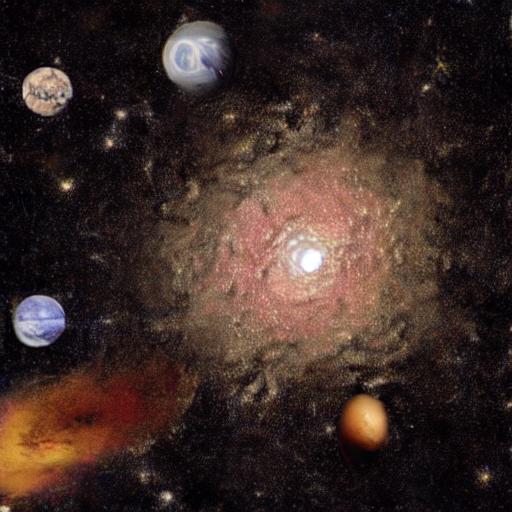} & \includegraphics[width=0.14\linewidth]{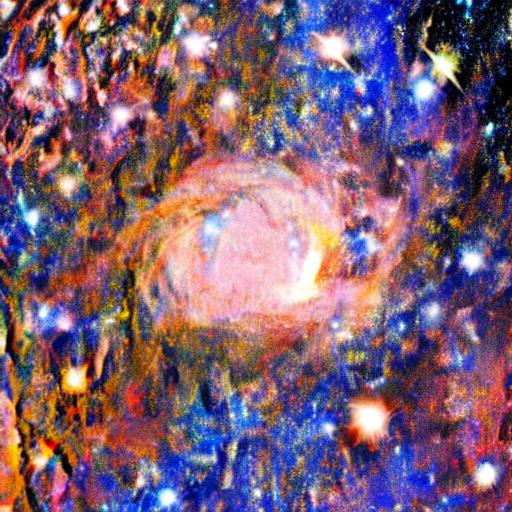}  & \includegraphics[width=0.14\linewidth]{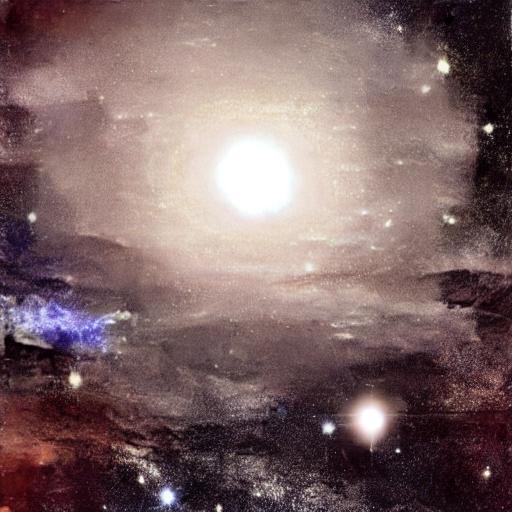} &
\includegraphics[width=0.14\linewidth]{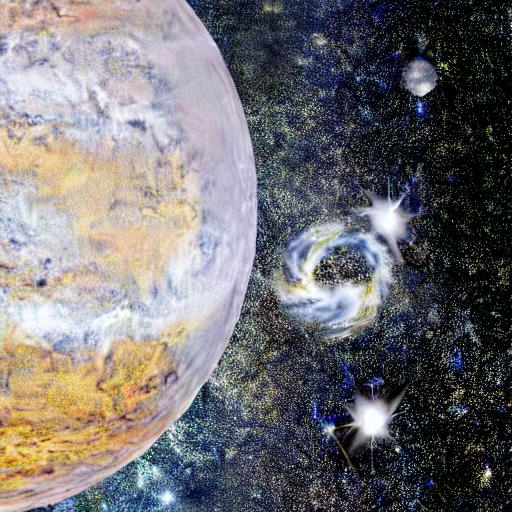} & \includegraphics[width=0.14\linewidth]{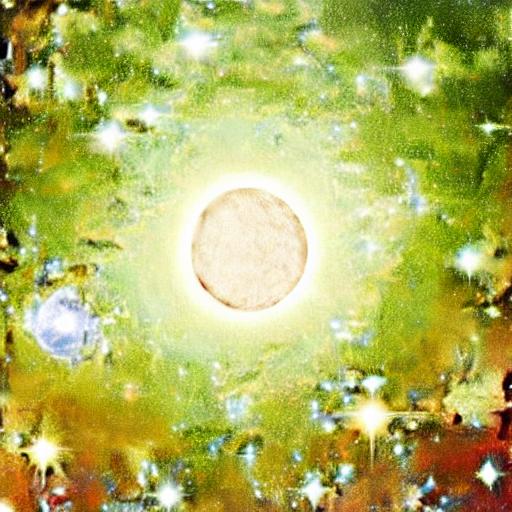} \\
\includegraphics[width=0.14\linewidth]{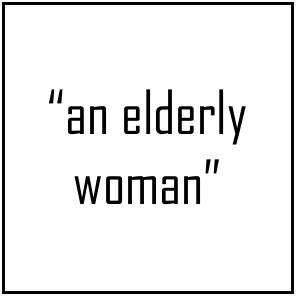} & 
\includegraphics[width=0.14\linewidth]{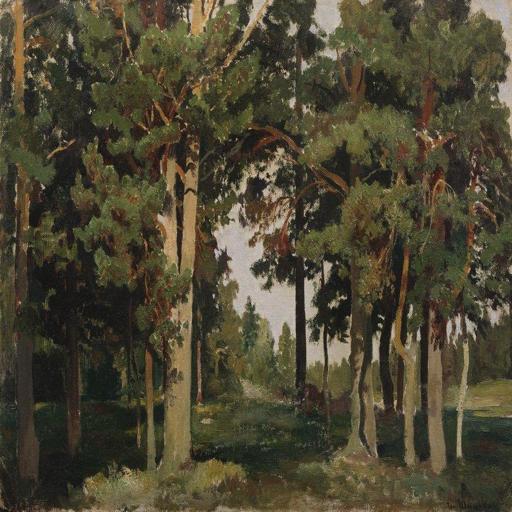} & \includegraphics[width=0.14\linewidth]{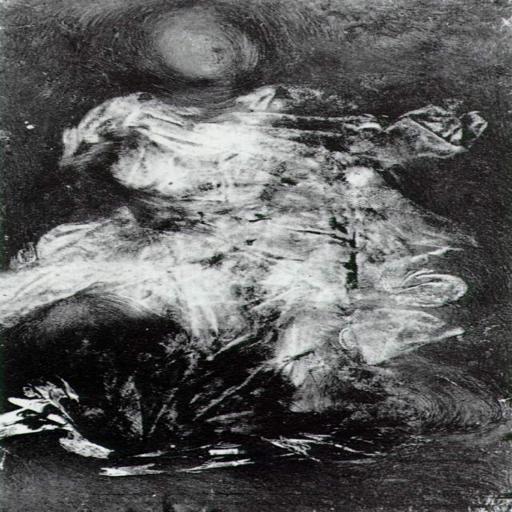}  & \includegraphics[width=0.14\linewidth]{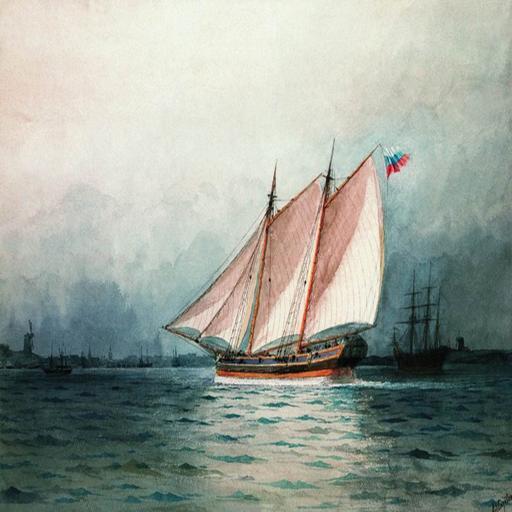} &
\includegraphics[width=0.14\linewidth]{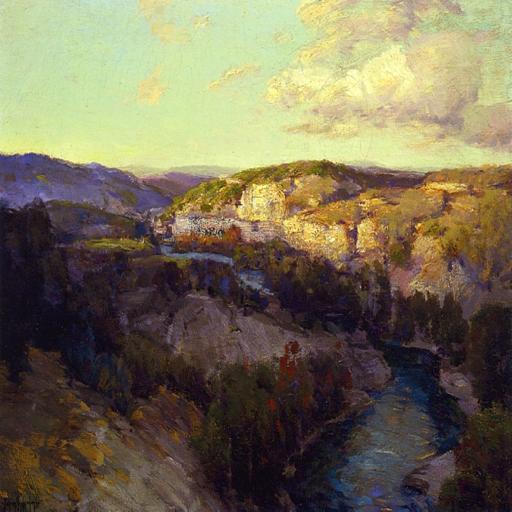} & \includegraphics[width=0.14\linewidth]{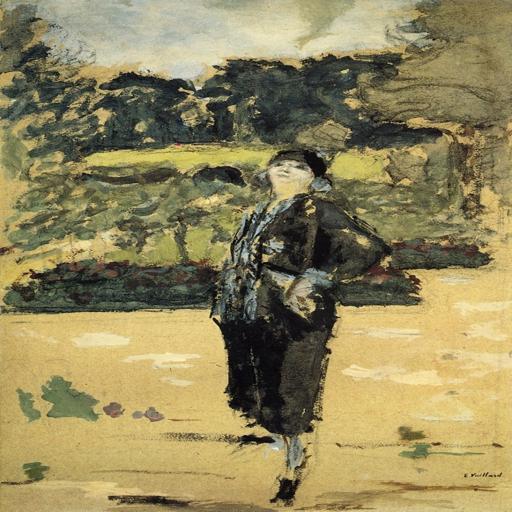}  \\ 
 & 
\includegraphics[width=0.14\linewidth]{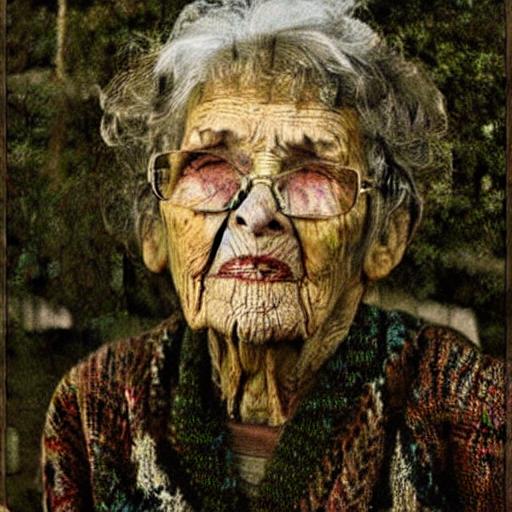} & \includegraphics[width=0.14\linewidth]{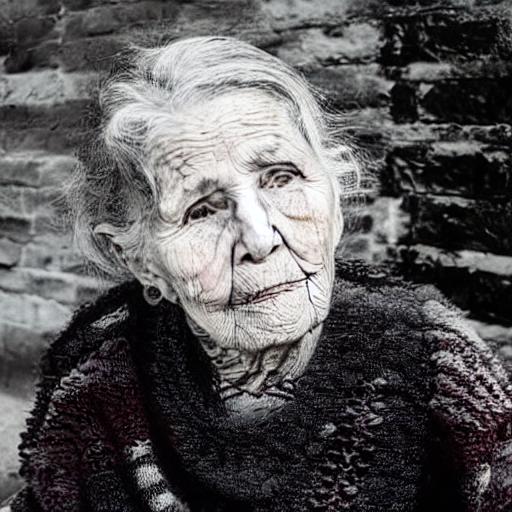}  & \includegraphics[width=0.14\linewidth]{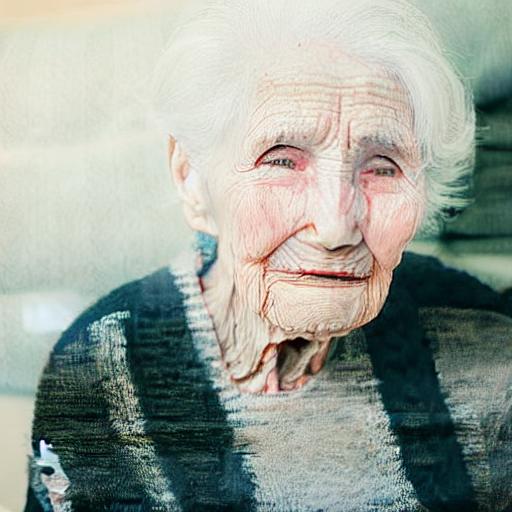} &
\includegraphics[width=0.14\linewidth]{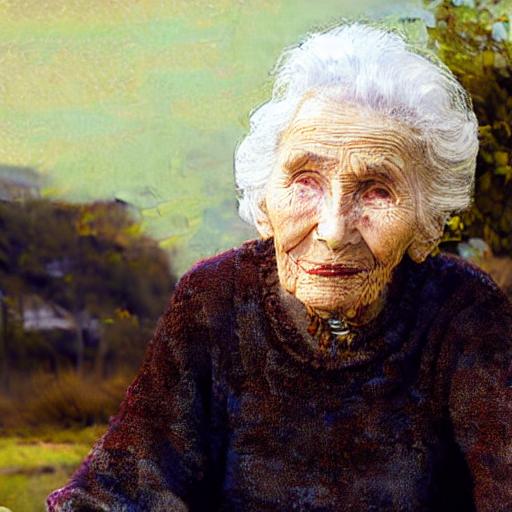} & \includegraphics[width=0.14\linewidth]{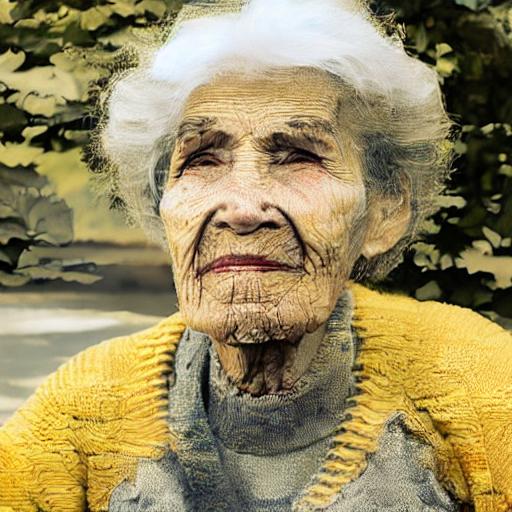} \\
\includegraphics[width=0.14\linewidth]{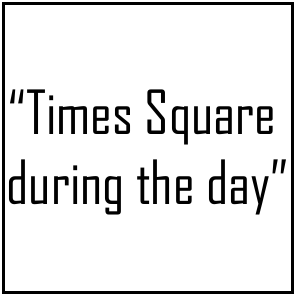} & 
\includegraphics[width=0.14\linewidth]{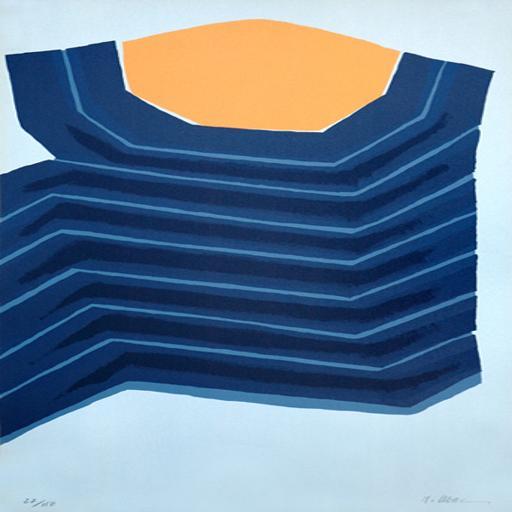} & \includegraphics[width=0.14\linewidth]{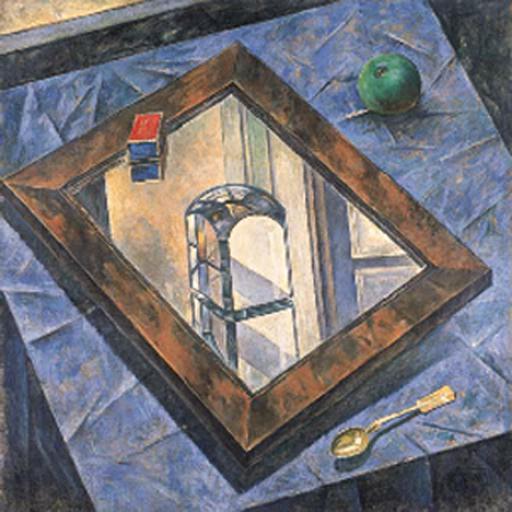}  & \includegraphics[width=0.14\linewidth]{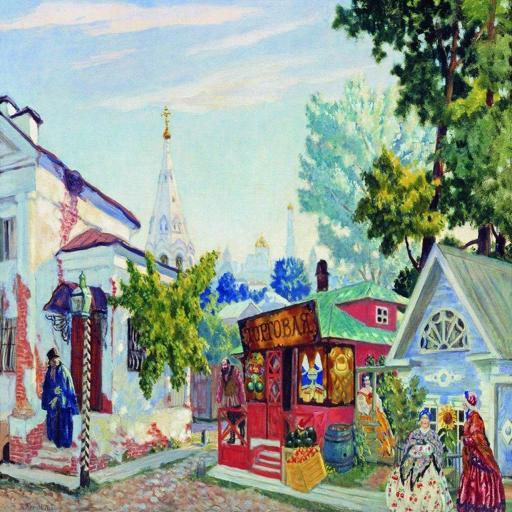} &
\includegraphics[width=0.14\linewidth]{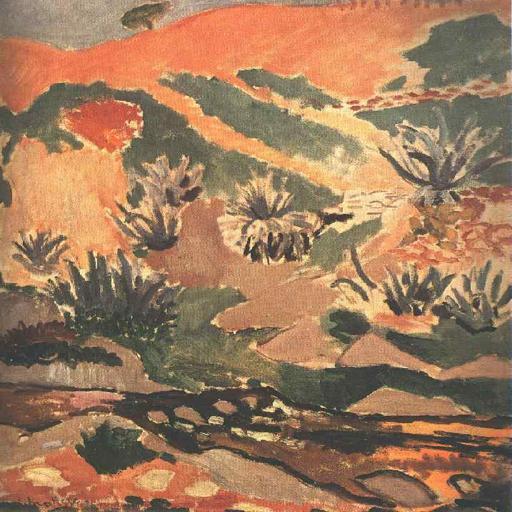} & \includegraphics[width=0.14\linewidth]{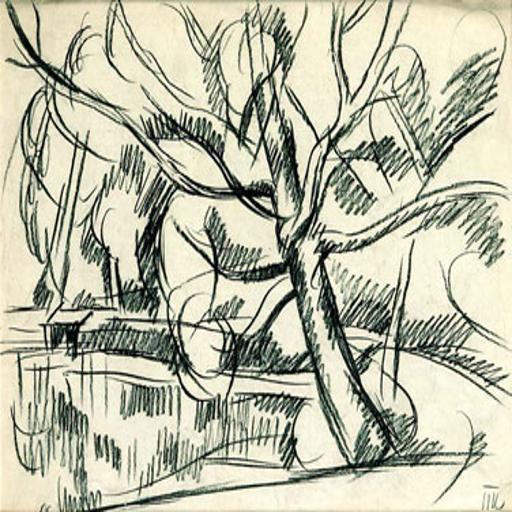}  \\ 
 & 
\includegraphics[width=0.14\linewidth]{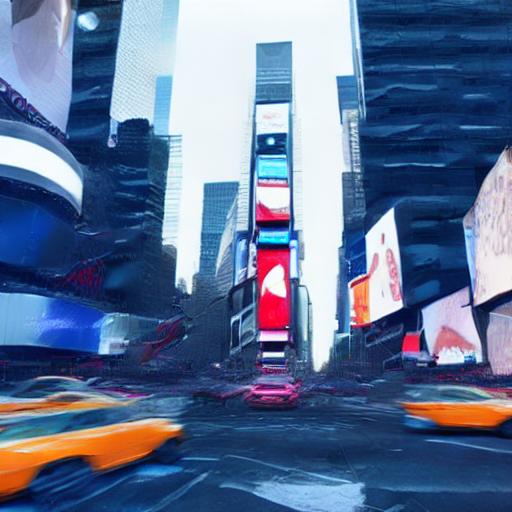} & \includegraphics[width=0.14\linewidth]{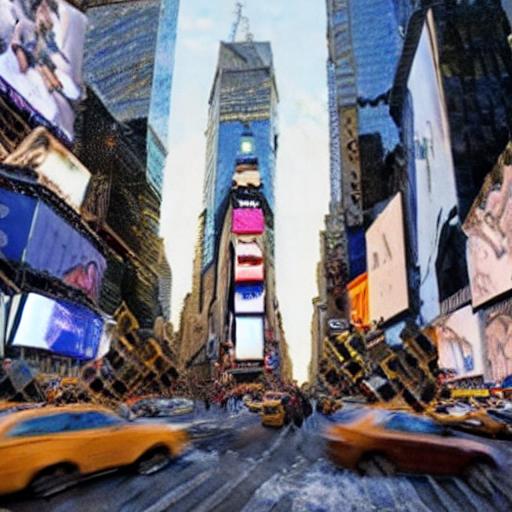}  & \includegraphics[width=0.14\linewidth]{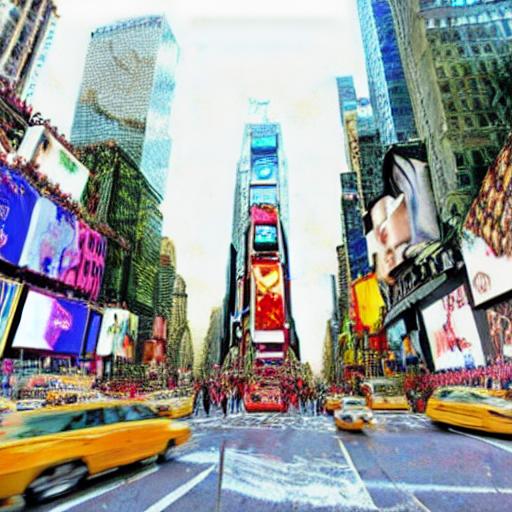} &
\includegraphics[width=0.14\linewidth]{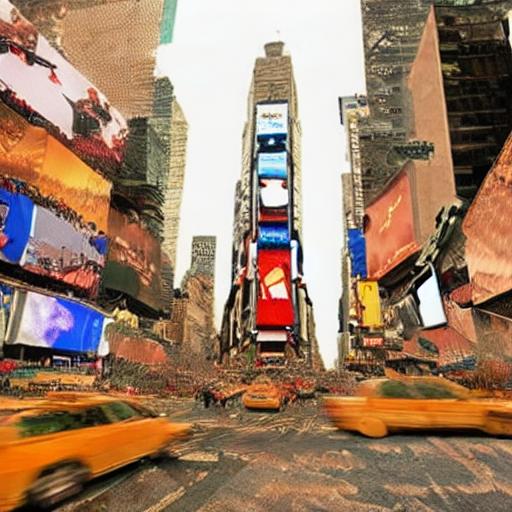} & \includegraphics[width=0.14\linewidth]{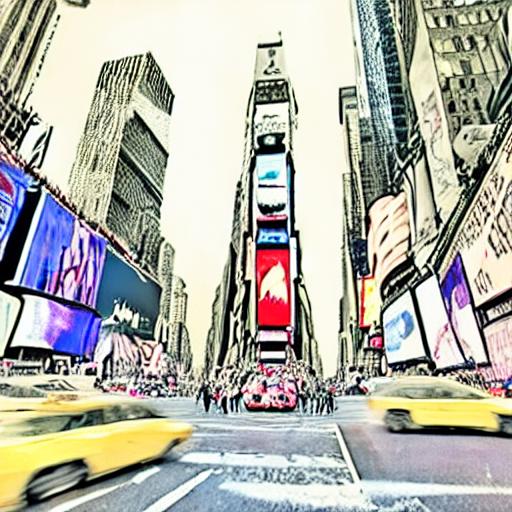} 
\end{tabular}
}

\caption{Qualitative results on the style transfer task. For each group, the top row shows the content and style references, and the bottom row is the resulting stylized output.}
\label{fig:a1}
\end{figure}

\begin{figure}[t]
\centering

\resizebox{0.98\textwidth}{!}{

\setlength{\tabcolsep}{0.05cm} % 调整列间距
\renewcommand{\arraystretch}{0.5}  % 调整行距
\begin{tabular}{ccccccc}
\includegraphics[width=0.14\linewidth]{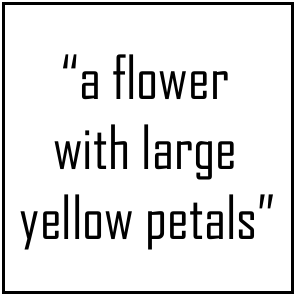} & 
\includegraphics[width=0.14\linewidth]{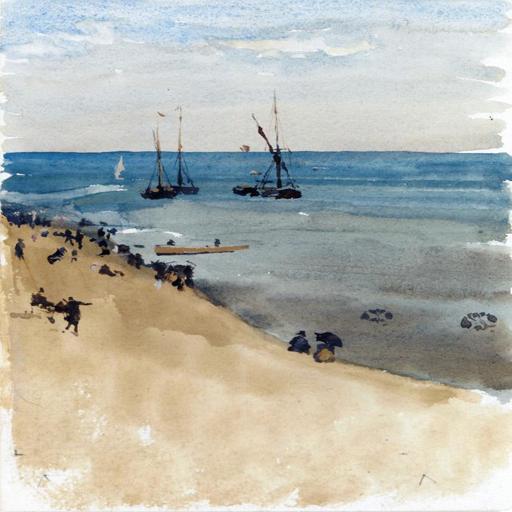} & \includegraphics[width=0.14\linewidth]{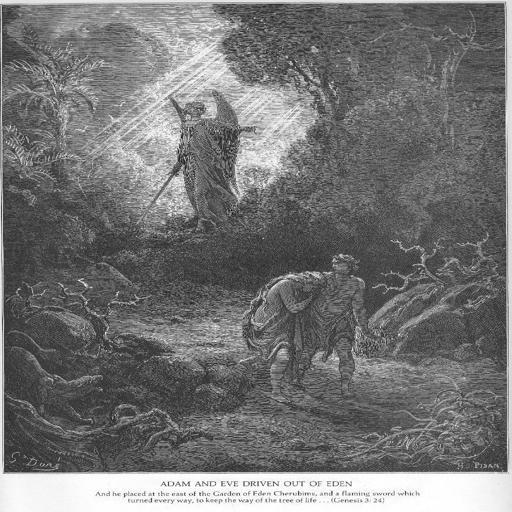}  & \includegraphics[width=0.14\linewidth]{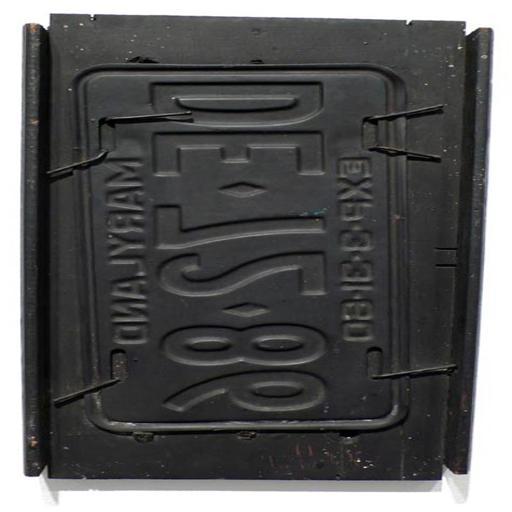} &
\includegraphics[width=0.14\linewidth]{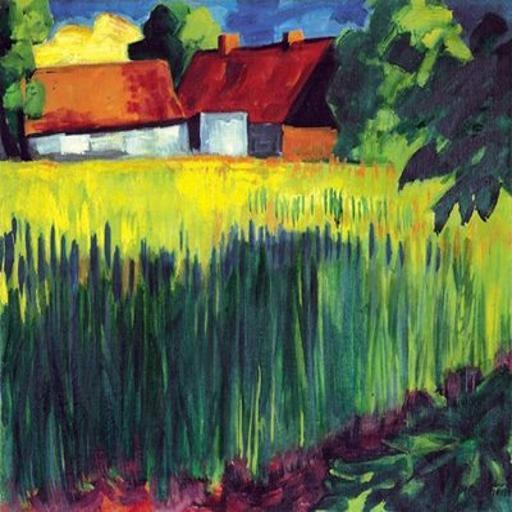} & \includegraphics[width=0.14\linewidth]{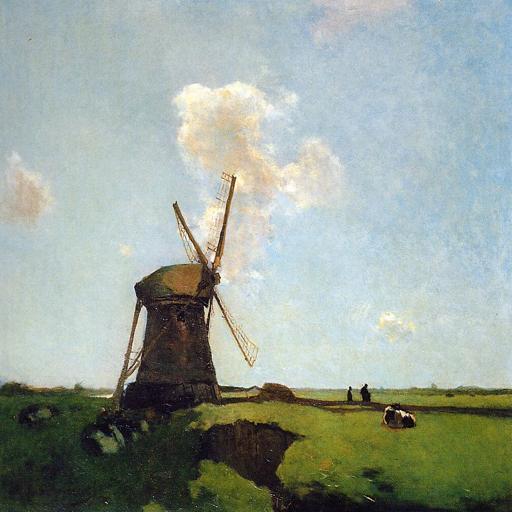} \\ & 
\includegraphics[width=0.14\linewidth]{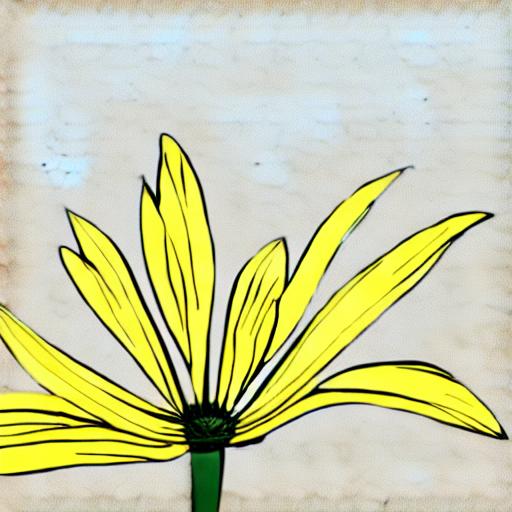} & \includegraphics[width=0.14\linewidth]{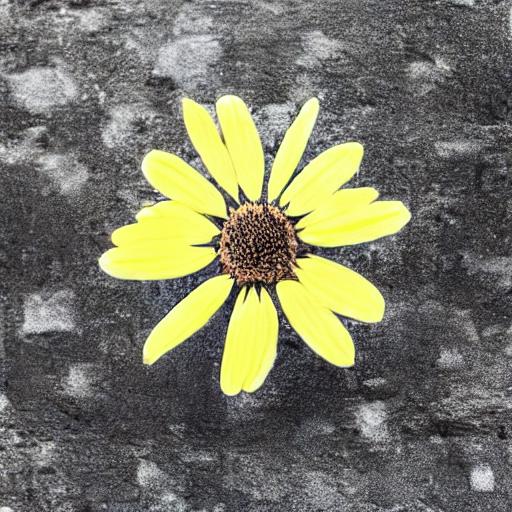}  & \includegraphics[width=0.14\linewidth]{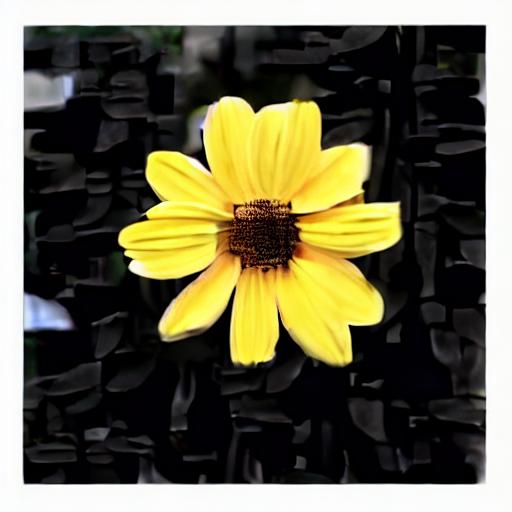} &
\includegraphics[width=0.14\linewidth]{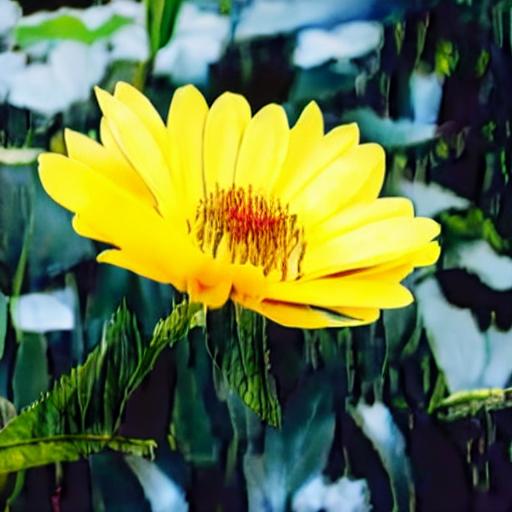} & \includegraphics[width=0.14\linewidth]{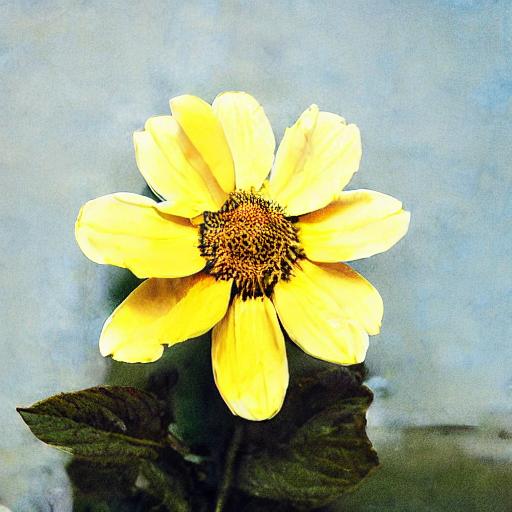} \\

\includegraphics[width=0.14\linewidth]{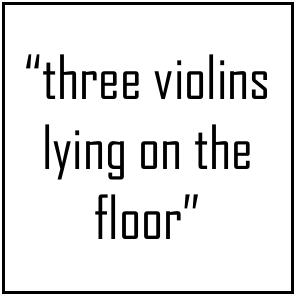} & 
\includegraphics[width=0.14\linewidth]{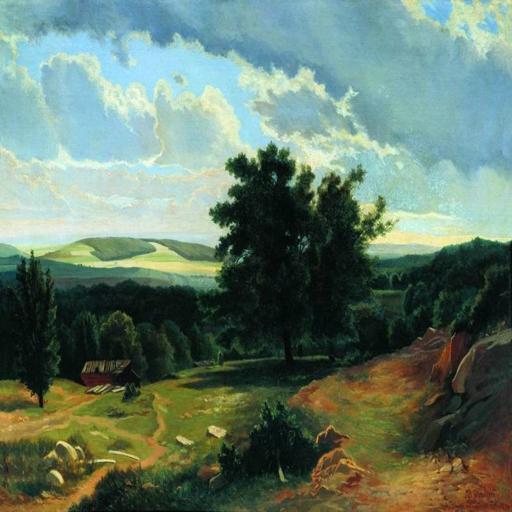} & \includegraphics[width=0.14\linewidth]{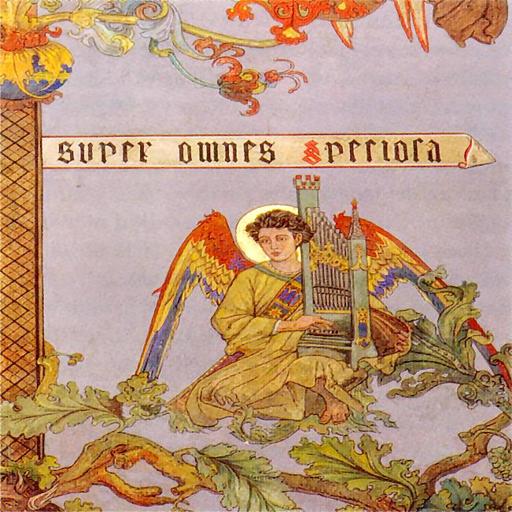}  & \includegraphics[width=0.14\linewidth]{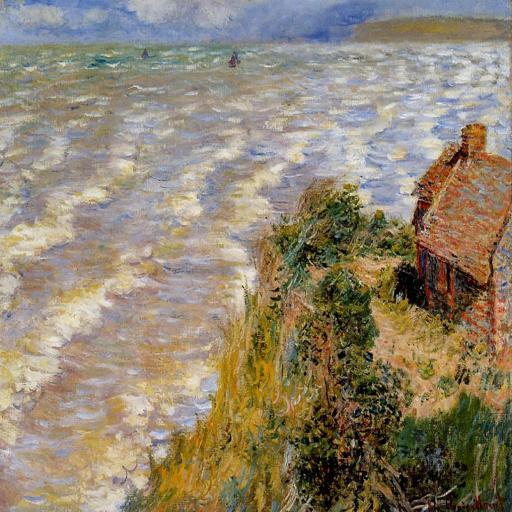} &
\includegraphics[width=0.14\linewidth]{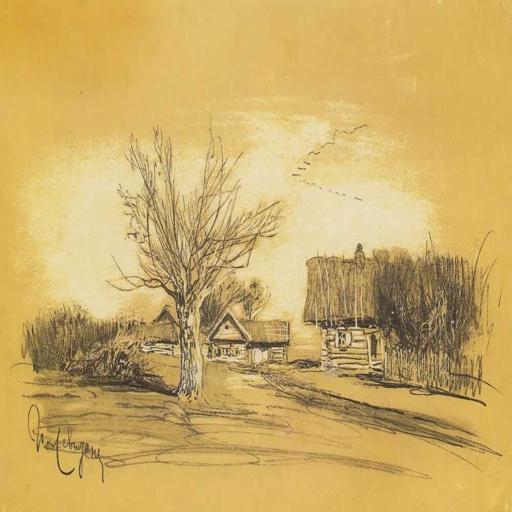} & \includegraphics[width=0.14\linewidth]{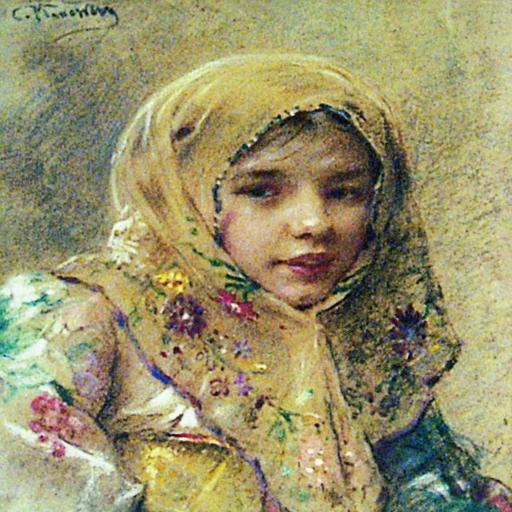} \\
& 
\includegraphics[width=0.14\linewidth]{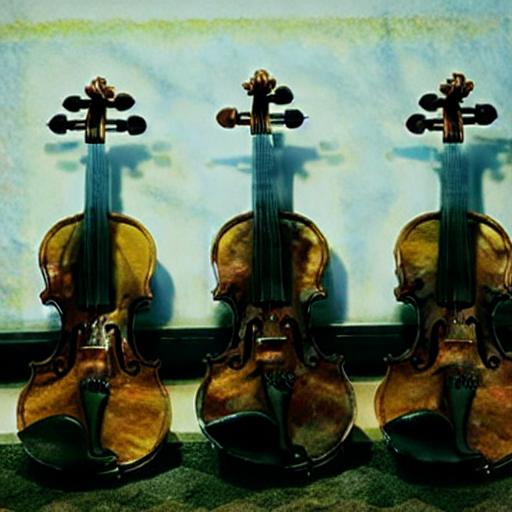} & \includegraphics[width=0.14\linewidth]{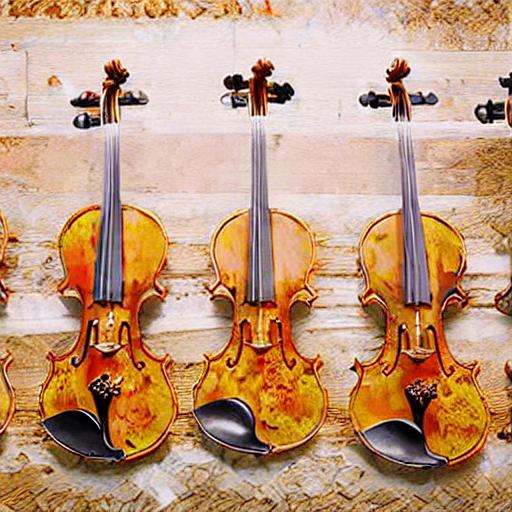}  & \includegraphics[width=0.14\linewidth]{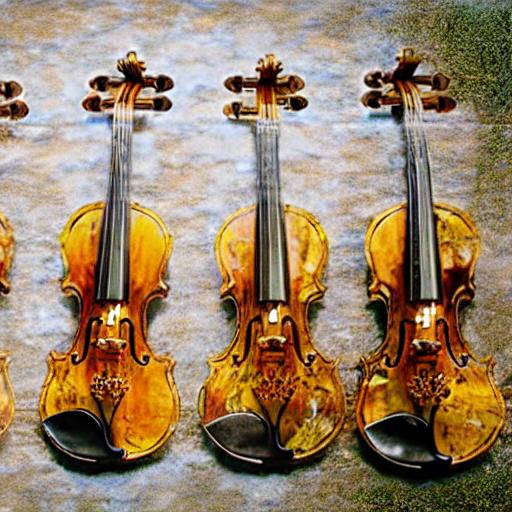} &
\includegraphics[width=0.14\linewidth]{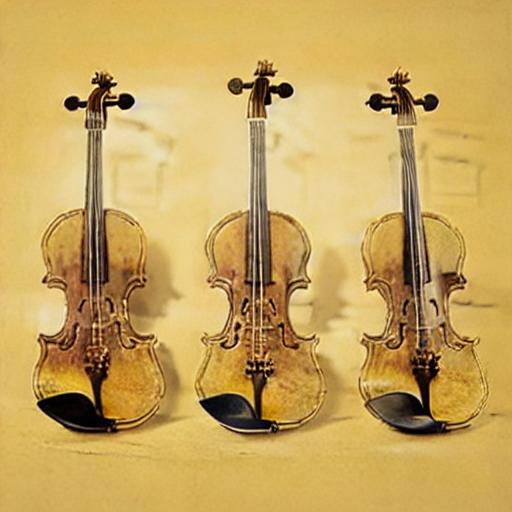} & \includegraphics[width=0.14\linewidth]{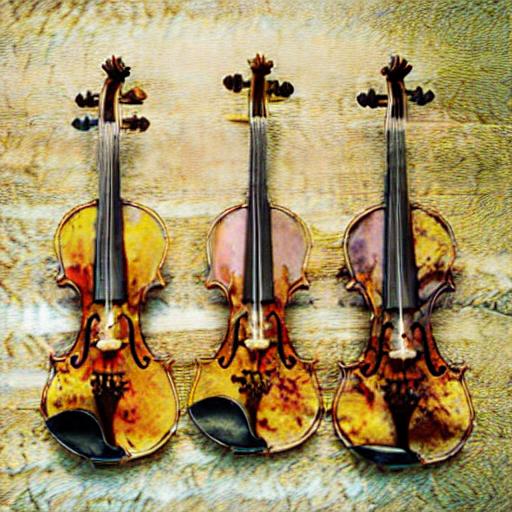} \\

\includegraphics[width=0.14\linewidth]{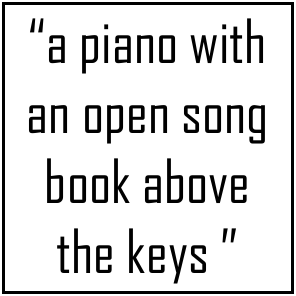} & 
\includegraphics[width=0.14\linewidth]{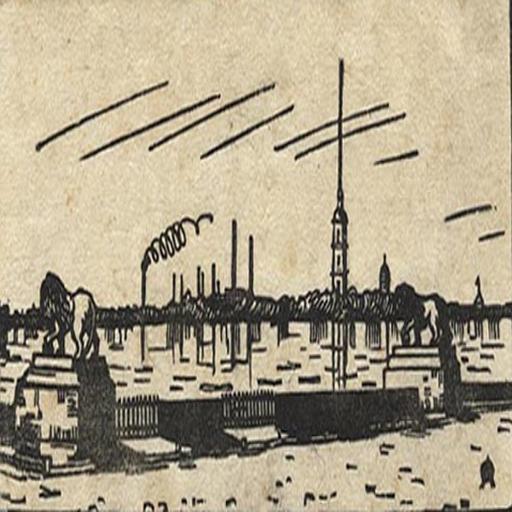} & \includegraphics[width=0.14\linewidth]{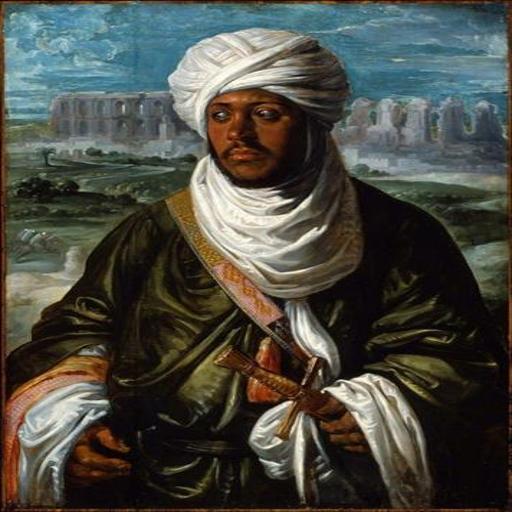}  & \includegraphics[width=0.14\linewidth]{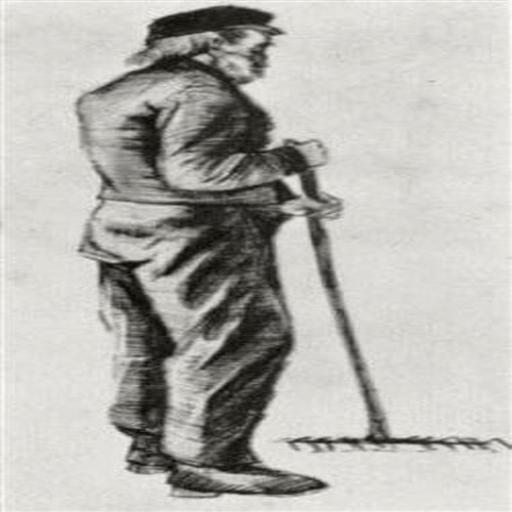} &
\includegraphics[width=0.14\linewidth]{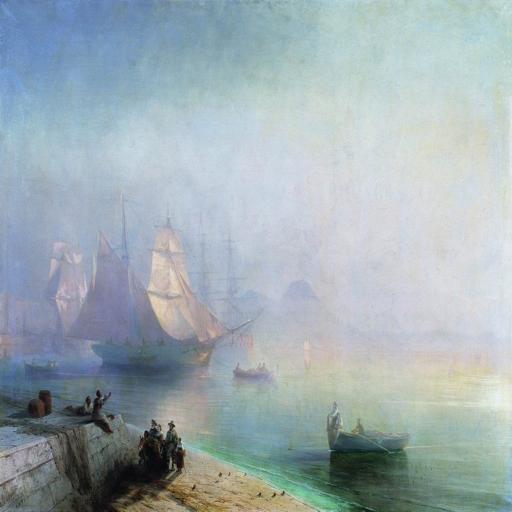} & \includegraphics[width=0.14\linewidth]{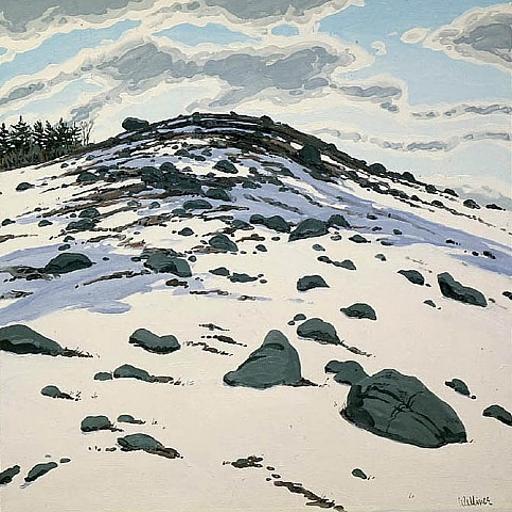} \\
& 
\includegraphics[width=0.14\linewidth]{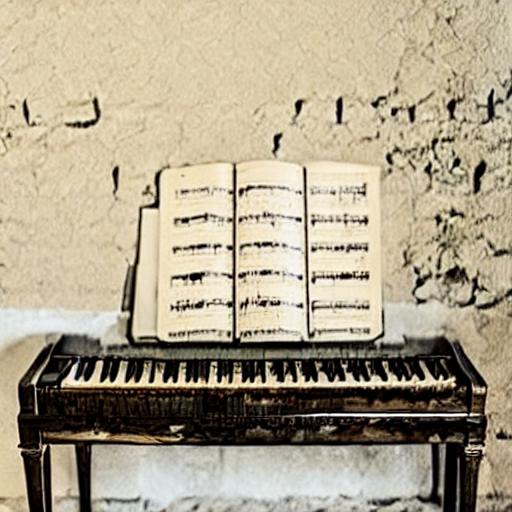} & \includegraphics[width=0.14\linewidth]{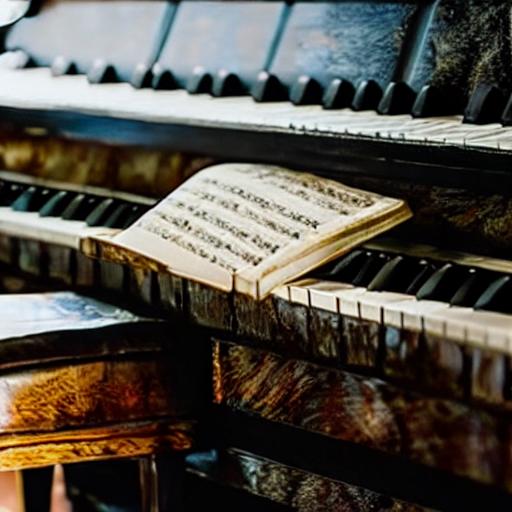}  & \includegraphics[width=0.14\linewidth]{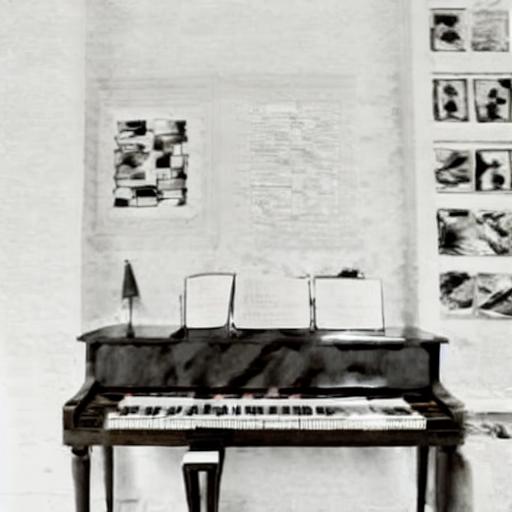} &
\includegraphics[width=0.14\linewidth]{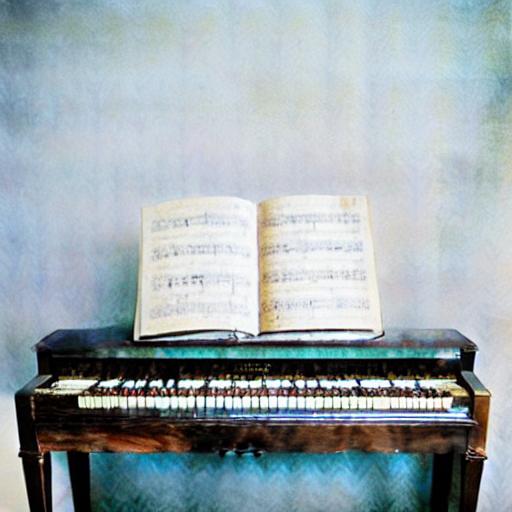} & \includegraphics[width=0.14\linewidth]{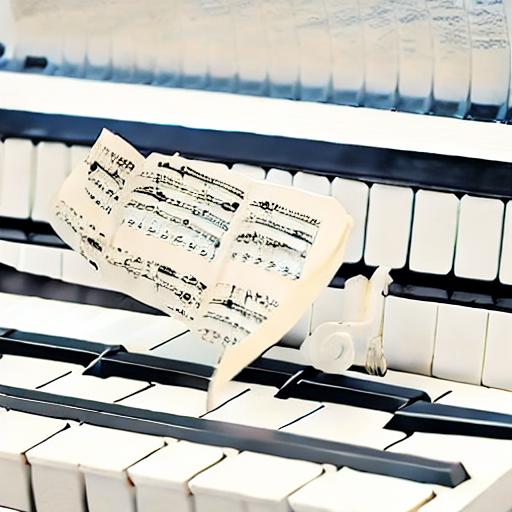} \\

\includegraphics[width=0.14\linewidth]{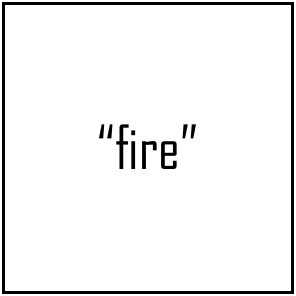} & 
\includegraphics[width=0.14\linewidth]{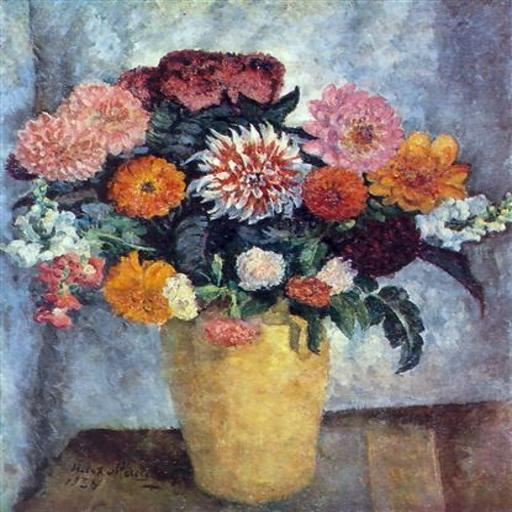} & \includegraphics[width=0.14\linewidth]{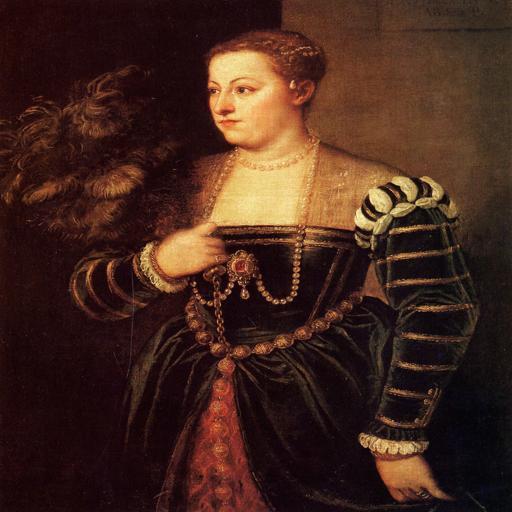}  & \includegraphics[width=0.14\linewidth]{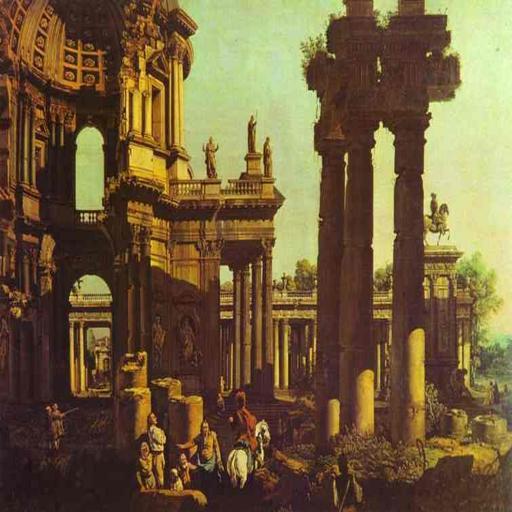} &
\includegraphics[width=0.14\linewidth]{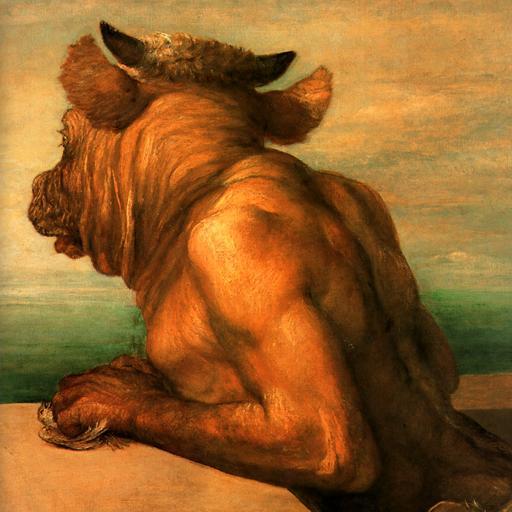} & \includegraphics[width=0.14\linewidth]{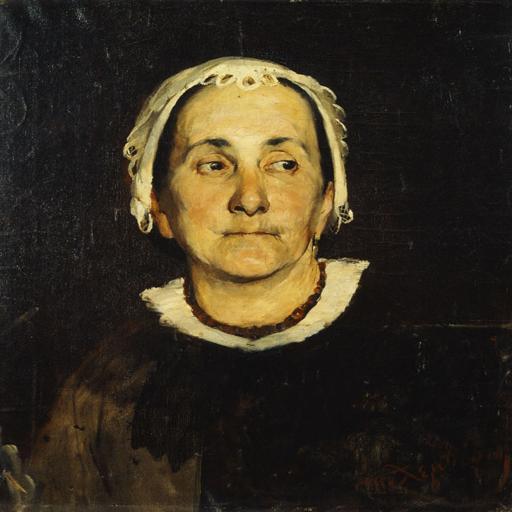} \\
& 
\includegraphics[width=0.14\linewidth]{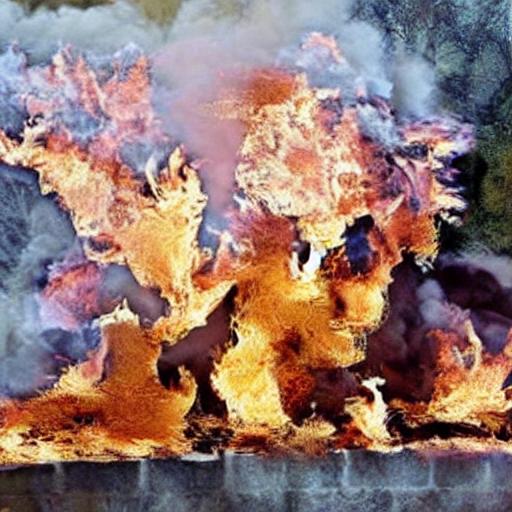} & \includegraphics[width=0.14\linewidth]{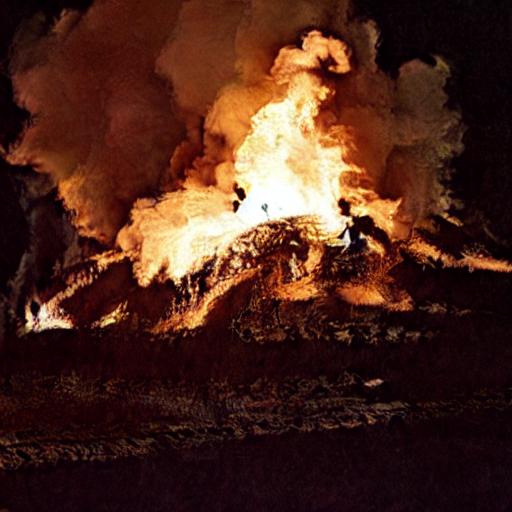}  & \includegraphics[width=0.14\linewidth]{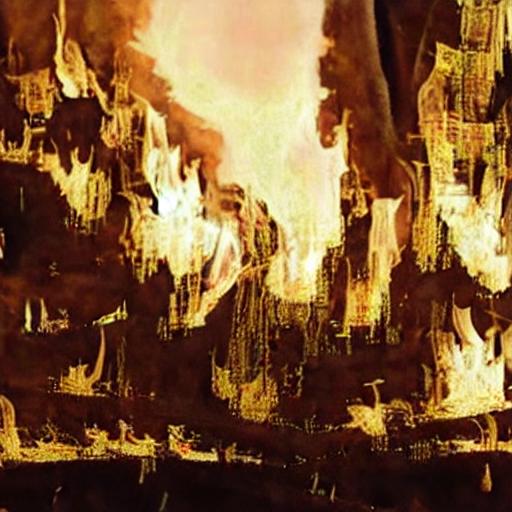} &
\includegraphics[width=0.14\linewidth]{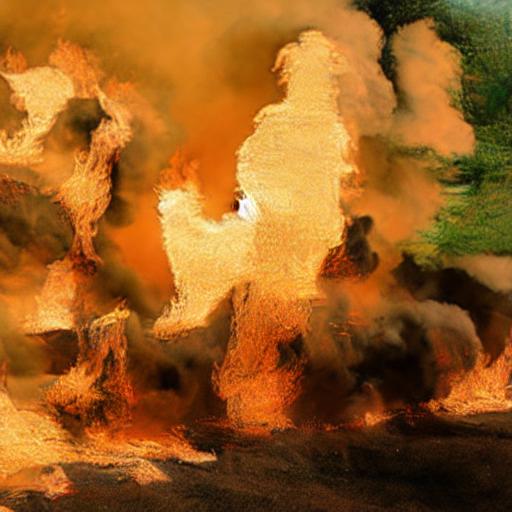} & \includegraphics[width=0.14\linewidth]{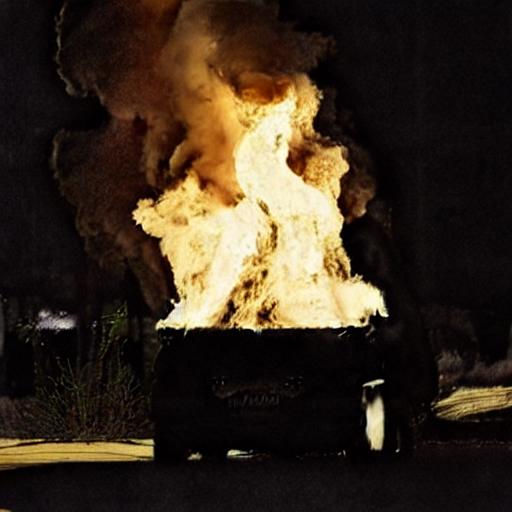}\\

\includegraphics[width=0.14\linewidth]{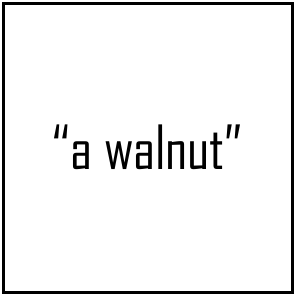} & 
\includegraphics[width=0.14\linewidth]{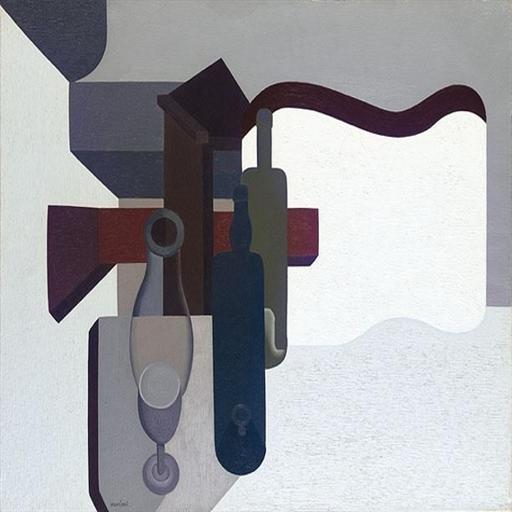} & \includegraphics[width=0.14\linewidth]{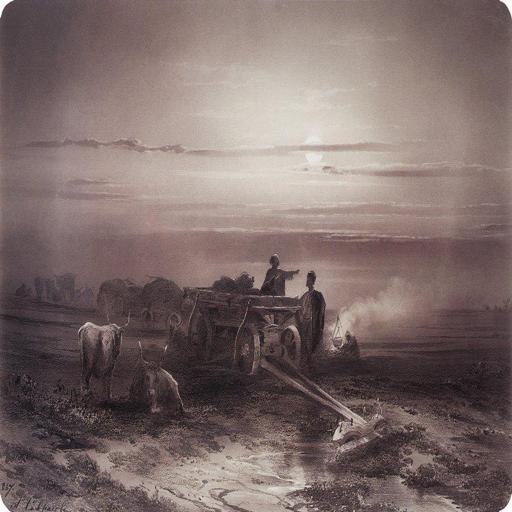}  & \includegraphics[width=0.14\linewidth]{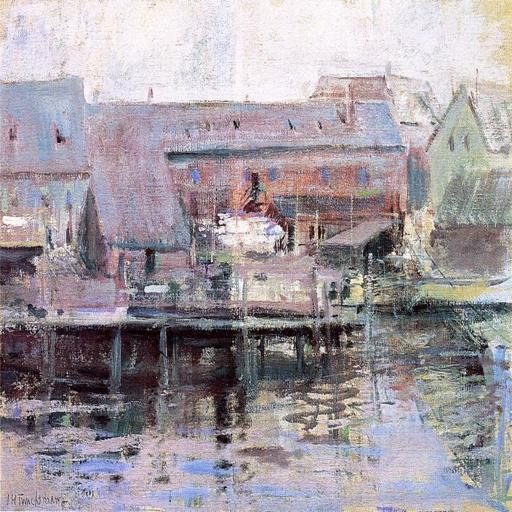} &
\includegraphics[width=0.14\linewidth]{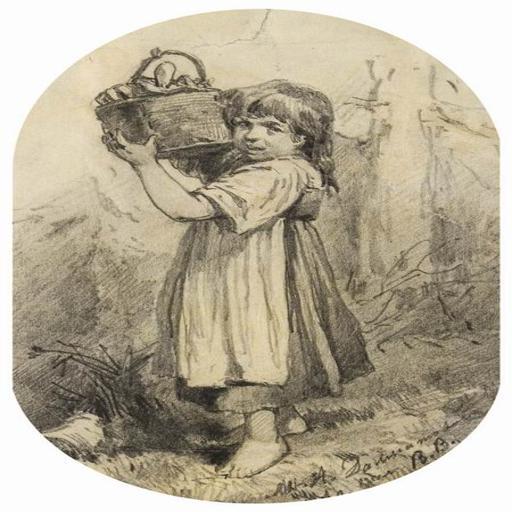} & \includegraphics[width=0.14\linewidth]{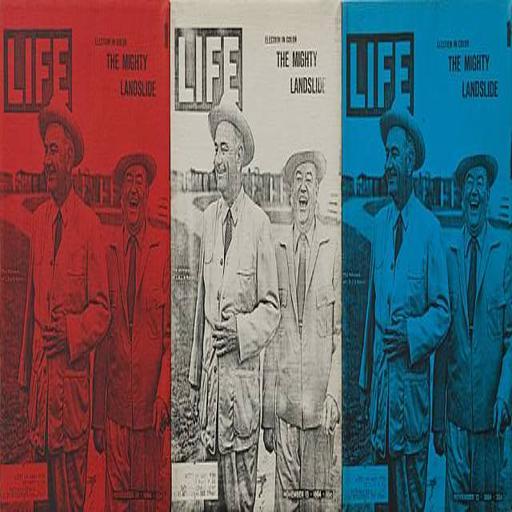} \\
 & 
\includegraphics[width=0.14\linewidth]{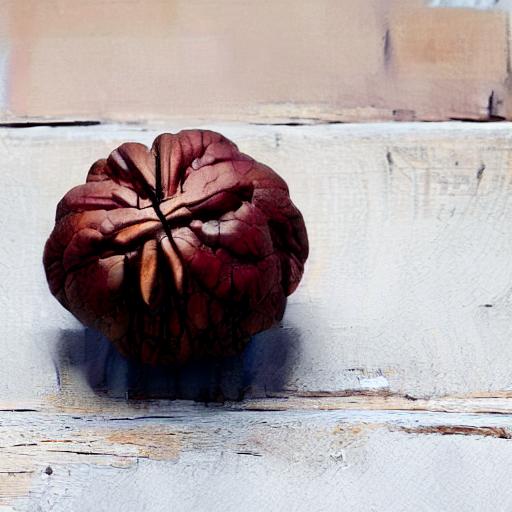} & \includegraphics[width=0.14\linewidth]{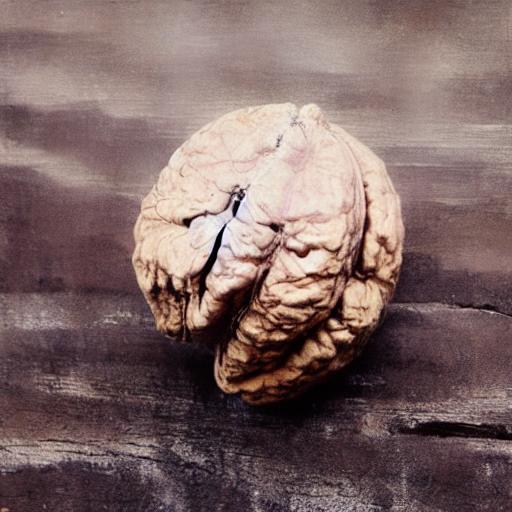}  & \includegraphics[width=0.14\linewidth]{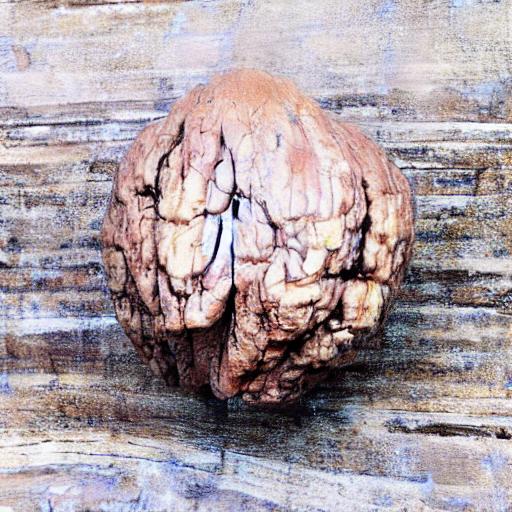} &
\includegraphics[width=0.14\linewidth]{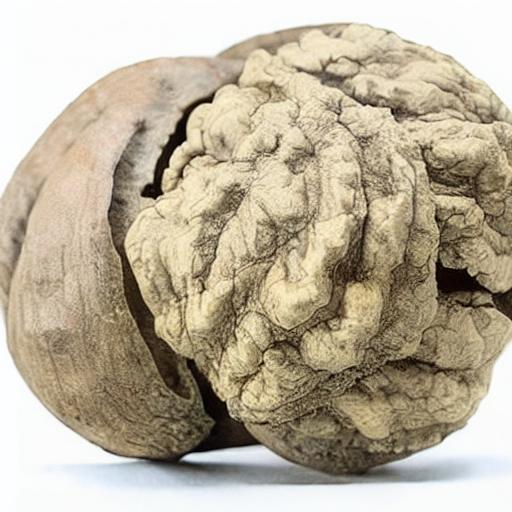} & \includegraphics[width=0.14\linewidth]{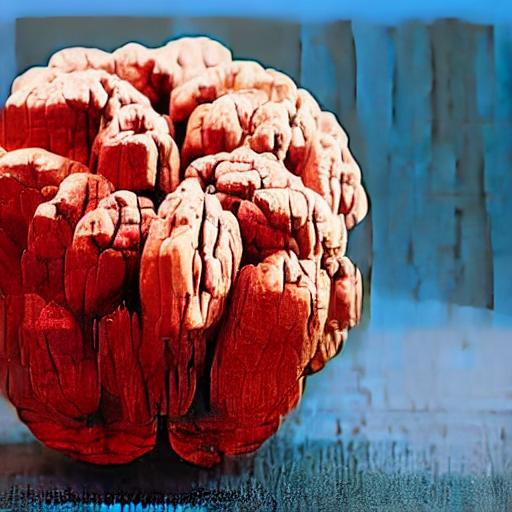} 
\end{tabular}
}

\caption{Qualitative results on the style transfer task. For each group, the top row shows the content and style references, and the bottom row is the resulting stylized output.}
\label{fig:a4}
\end{figure}

\begin{figure}[t]
\centering

\resizebox{0.98\textwidth}{!}{

\setlength{\tabcolsep}{0.05cm} % 调整列间距
\renewcommand{\arraystretch}{0.5}  % 调整行距
\begin{tabular}{ccc|cccc}
Condition & Target & DICT & Condition & Target & DICT \\
\includegraphics[width=0.14\linewidth]{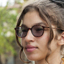} & 
\includegraphics[width=0.14\linewidth]{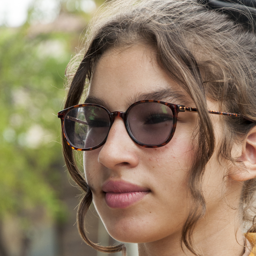} & \includegraphics[width=0.14\linewidth]{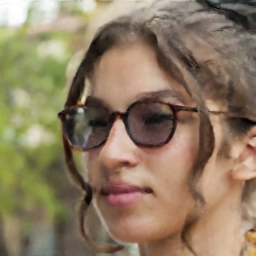}  & \includegraphics[width=0.14\linewidth]{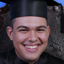} &
\includegraphics[width=0.14\linewidth]{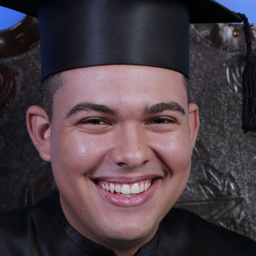} & \includegraphics[width=0.14\linewidth]{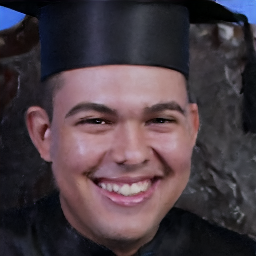} \\ \includegraphics[width=0.14\linewidth]{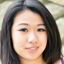} & 
\includegraphics[width=0.14\linewidth]{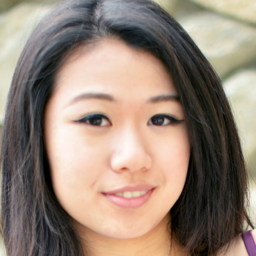} & \includegraphics[width=0.14\linewidth]{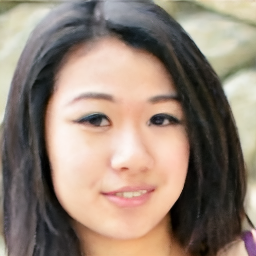}  & \includegraphics[width=0.14\linewidth]{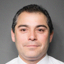} &
\includegraphics[width=0.14\linewidth]{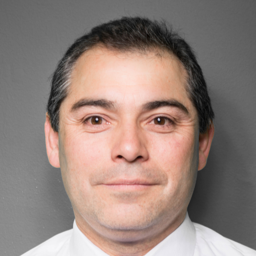} & \includegraphics[width=0.14\linewidth]{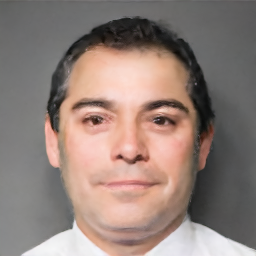} \\ 

\includegraphics[width=0.14\linewidth]{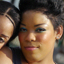} & 
\includegraphics[width=0.14\linewidth]{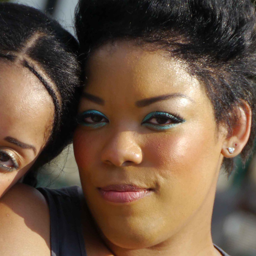} & \includegraphics[width=0.14\linewidth]{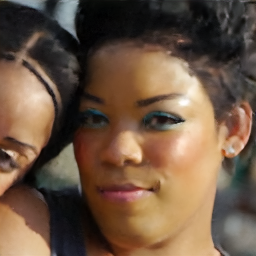}  & \includegraphics[width=0.14\linewidth]{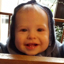} &
\includegraphics[width=0.14\linewidth]{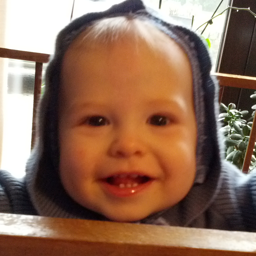} & \includegraphics[width=0.14\linewidth]{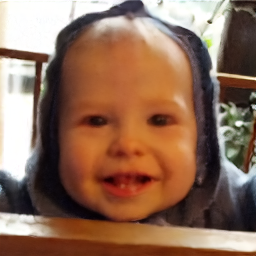} \\ 

\includegraphics[width=0.14\linewidth]{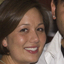} & 
\includegraphics[width=0.14\linewidth]{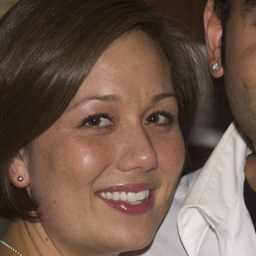} & \includegraphics[width=0.14\linewidth]{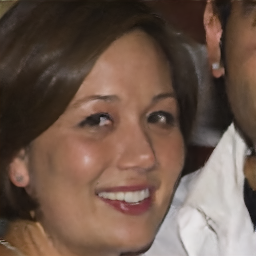}  & \includegraphics[width=0.14\linewidth]{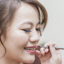} &
\includegraphics[width=0.14\linewidth]{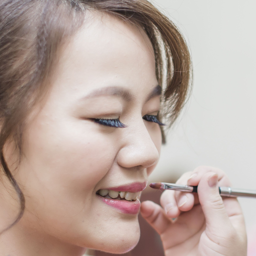} & \includegraphics[width=0.14\linewidth]{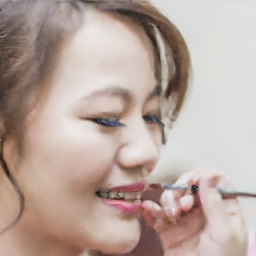} \\

\includegraphics[width=0.14\linewidth]{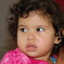} & 
\includegraphics[width=0.14\linewidth]{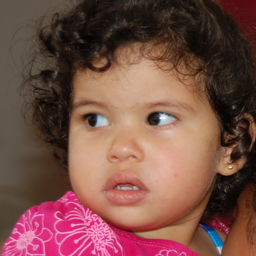} & \includegraphics[width=0.14\linewidth]{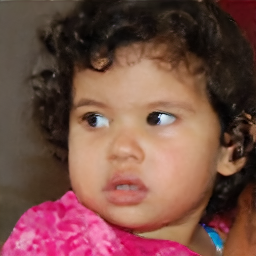}  & \includegraphics[width=0.14\linewidth]{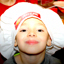} &
\includegraphics[width=0.14\linewidth]{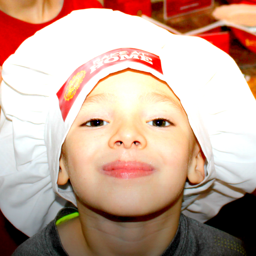} & \includegraphics[width=0.14\linewidth]{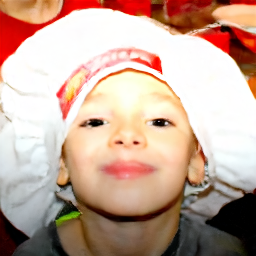} \\

\includegraphics[width=0.14\linewidth]{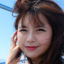} & 
\includegraphics[width=0.14\linewidth]{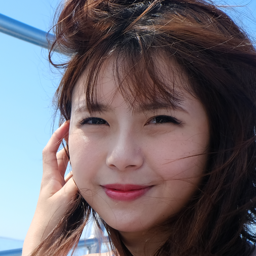} & \includegraphics[width=0.14\linewidth]{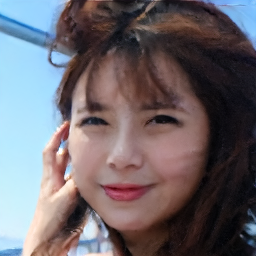}  & \includegraphics[width=0.14\linewidth]{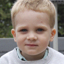} &
\includegraphics[width=0.14\linewidth]{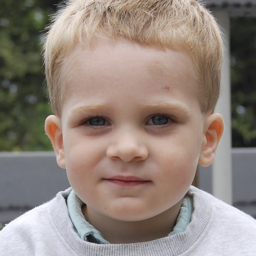} & \includegraphics[width=0.14\linewidth]{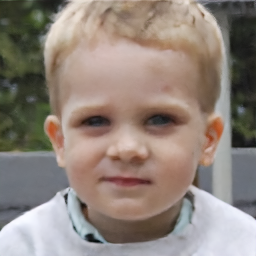} \\

\includegraphics[width=0.14\linewidth]{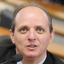} & 
\includegraphics[width=0.14\linewidth]{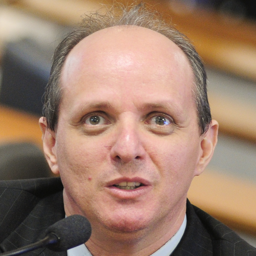} & \includegraphics[width=0.14\linewidth]{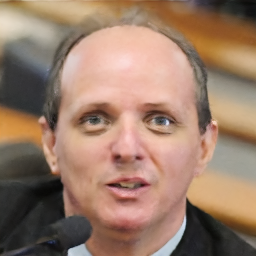}  & \includegraphics[width=0.14\linewidth]{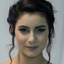} &
\includegraphics[width=0.14\linewidth]{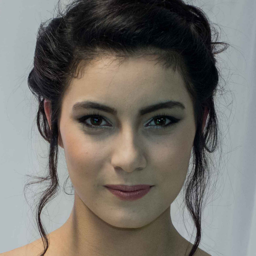} & \includegraphics[width=0.14\linewidth]{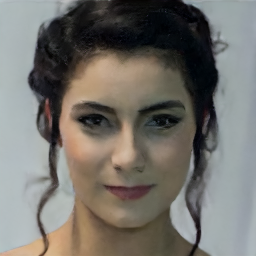} \\

\includegraphics[width=0.14\linewidth]{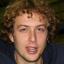} & 
\includegraphics[width=0.14\linewidth]{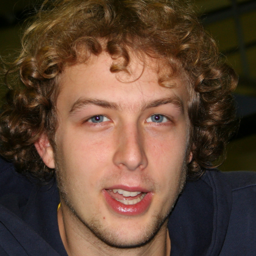} & \includegraphics[width=0.14\linewidth]{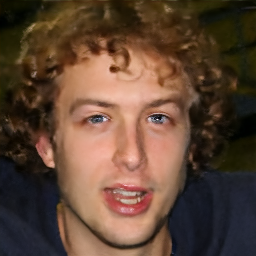}  & \includegraphics[width=0.14\linewidth]{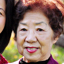} &
\includegraphics[width=0.14\linewidth]{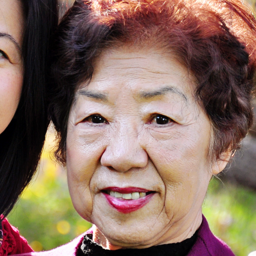} & \includegraphics[width=0.14\linewidth]{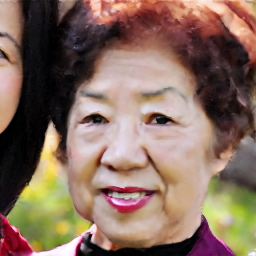} \\

\includegraphics[width=0.14\linewidth]{appendix_images/sr/60124_0_label.png} & 
\includegraphics[width=0.14\linewidth]{appendix_images/sr/60124_0_target.png} & \includegraphics[width=0.14\linewidth]{appendix_images/sr/60124_Ours.png}  & \includegraphics[width=0.14\linewidth]{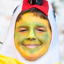} &
\includegraphics[width=0.14\linewidth]{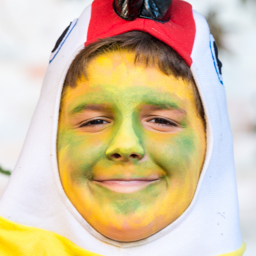} & \includegraphics[width=0.14\linewidth]{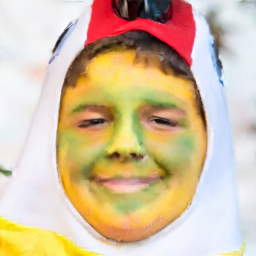} \\

\includegraphics[width=0.14\linewidth]{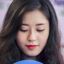} & 
\includegraphics[width=0.14\linewidth]{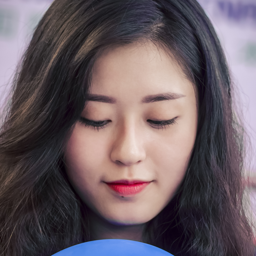} & \includegraphics[width=0.14\linewidth]{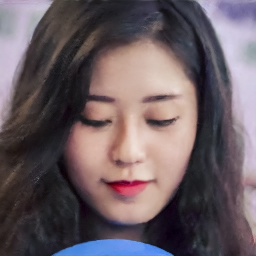}  & \includegraphics[width=0.14\linewidth]{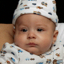} &
\includegraphics[width=0.14\linewidth]{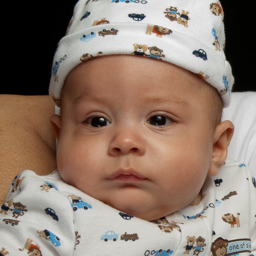} & \includegraphics[width=0.14\linewidth]{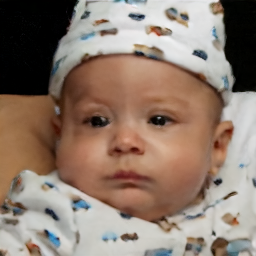} \\

\end{tabular}
}

\caption{Qualitative results on the image super-resolution task.}
\label{fig:a2}
\end{figure}

\begin{figure}[t]
\centering

\resizebox{0.98\textwidth}{!}{

\setlength{\tabcolsep}{0.05cm} % 调整列间距
\renewcommand{\arraystretch}{0.5}  % 调整行距
\begin{tabular}{ccc|cccc}
Condition & Target & DICT & Condition & Target & DICT \\
\includegraphics[width=0.14\linewidth]{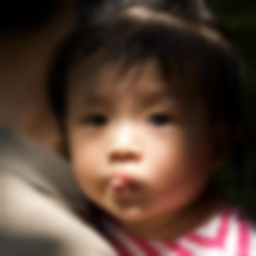} & 
\includegraphics[width=0.14\linewidth]{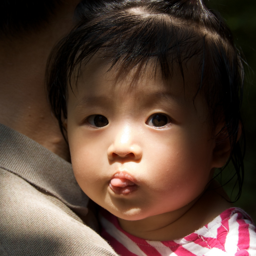} & \includegraphics[width=0.14\linewidth]{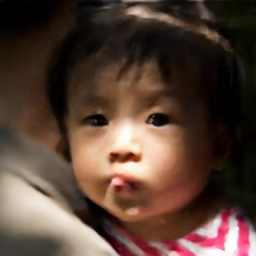}  & \includegraphics[width=0.14\linewidth]{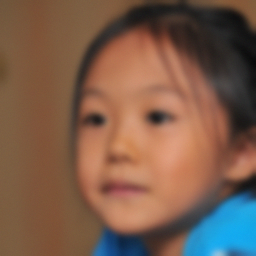} &
\includegraphics[width=0.14\linewidth]{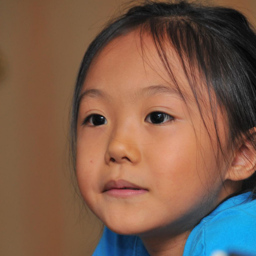} & \includegraphics[width=0.14\linewidth]{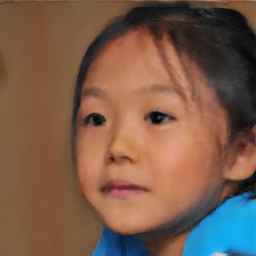} \\ \includegraphics[width=0.14\linewidth]{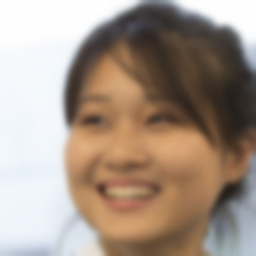} & 
\includegraphics[width=0.14\linewidth]{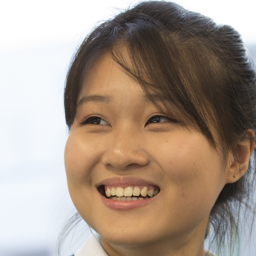} & \includegraphics[width=0.14\linewidth]{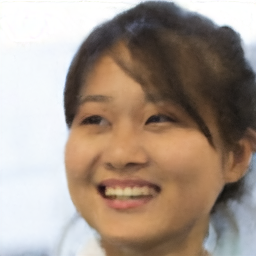}  & \includegraphics[width=0.14\linewidth]{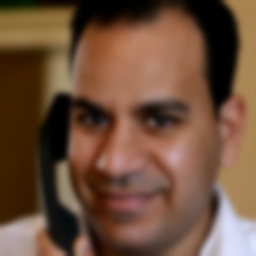} &
\includegraphics[width=0.14\linewidth]{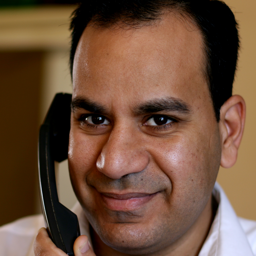} & \includegraphics[width=0.14\linewidth]{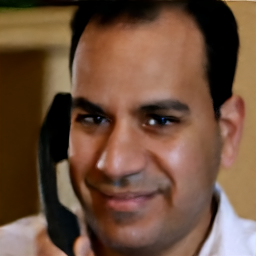} \\ 

\includegraphics[width=0.14\linewidth]{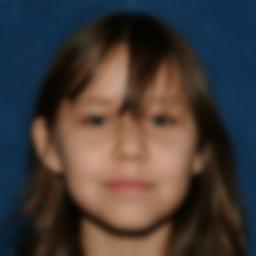} & 
\includegraphics[width=0.14\linewidth]{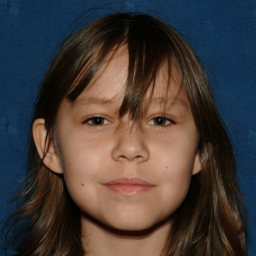} & \includegraphics[width=0.14\linewidth]{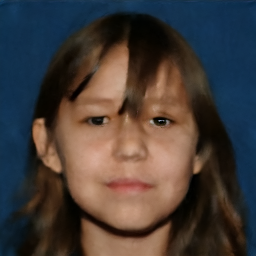}  & \includegraphics[width=0.14\linewidth]{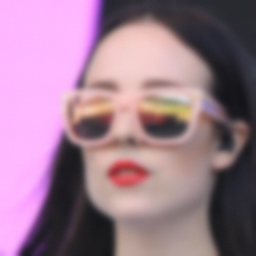} &
\includegraphics[width=0.14\linewidth]{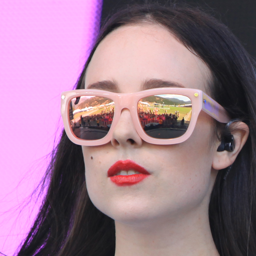} & \includegraphics[width=0.14\linewidth]{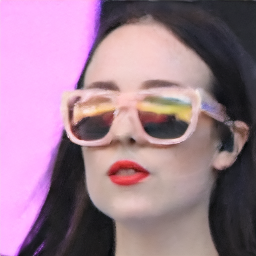} \\

\includegraphics[width=0.14\linewidth]{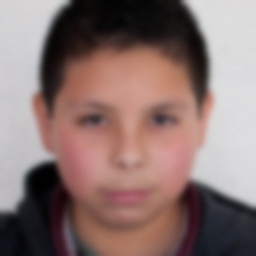} & 
\includegraphics[width=0.14\linewidth]{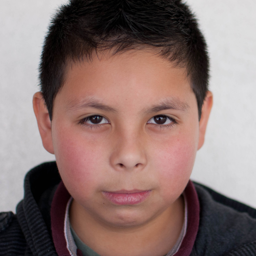} & \includegraphics[width=0.14\linewidth]{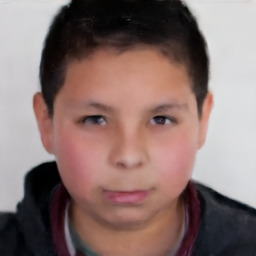}  & \includegraphics[width=0.14\linewidth]{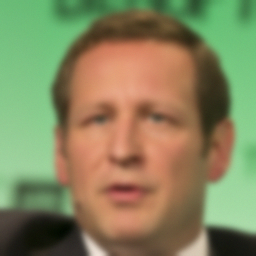} &
\includegraphics[width=0.14\linewidth]{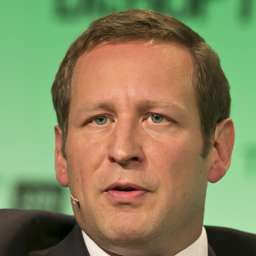} & \includegraphics[width=0.14\linewidth]{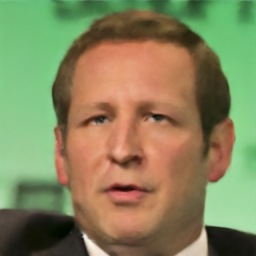} \\ 

\includegraphics[width=0.14\linewidth]{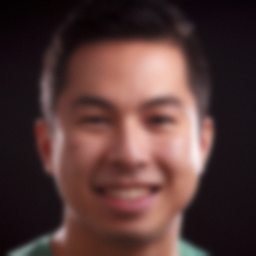} & 
\includegraphics[width=0.14\linewidth]{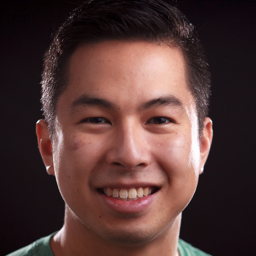} & \includegraphics[width=0.14\linewidth]{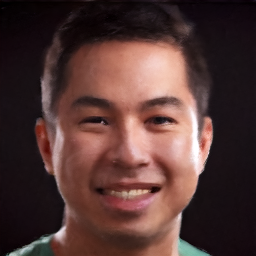}  & \includegraphics[width=0.14\linewidth]{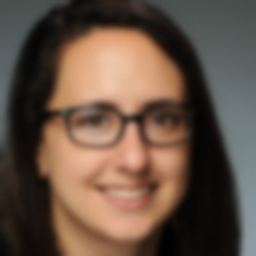} &
\includegraphics[width=0.14\linewidth]{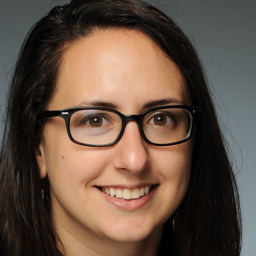} & \includegraphics[width=0.14\linewidth]{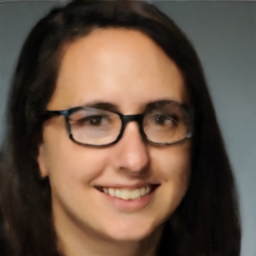} \\ 

\includegraphics[width=0.14\linewidth]{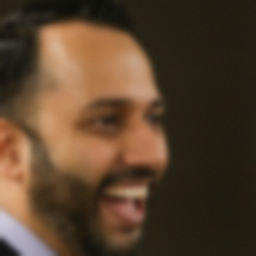} & 
\includegraphics[width=0.14\linewidth]{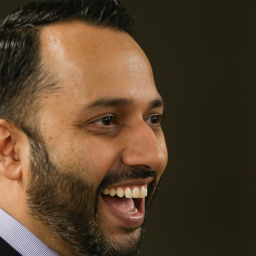} & \includegraphics[width=0.14\linewidth]{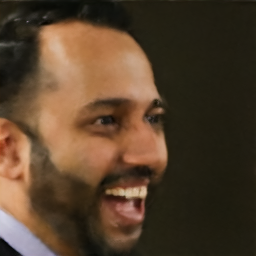}  & \includegraphics[width=0.14\linewidth]{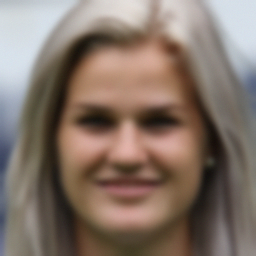} &
\includegraphics[width=0.14\linewidth]{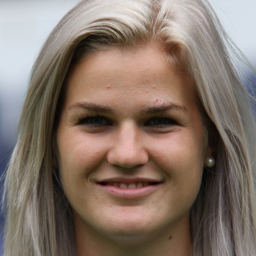} & \includegraphics[width=0.14\linewidth]{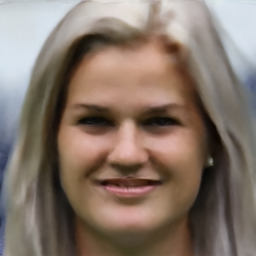} \\ 

\includegraphics[width=0.14\linewidth]{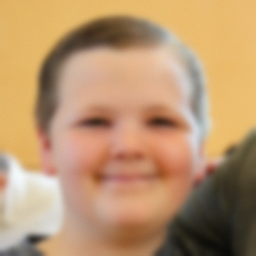} & 
\includegraphics[width=0.14\linewidth]{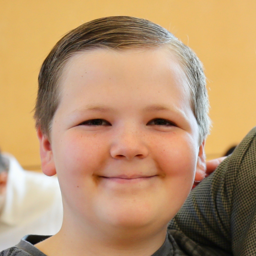} & \includegraphics[width=0.14\linewidth]{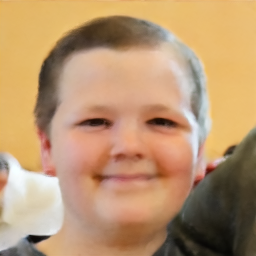}  & \includegraphics[width=0.14\linewidth]{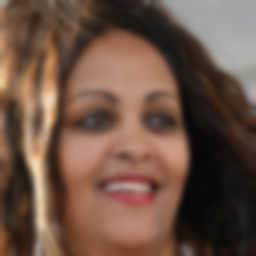} &
\includegraphics[width=0.14\linewidth]{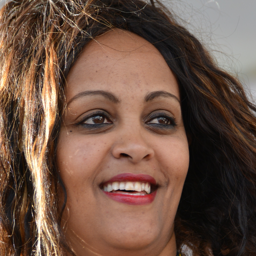} & \includegraphics[width=0.14\linewidth]{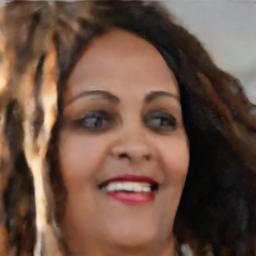} \\

\includegraphics[width=0.14\linewidth]{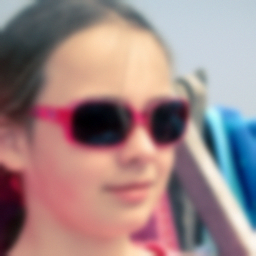} & 
\includegraphics[width=0.14\linewidth]{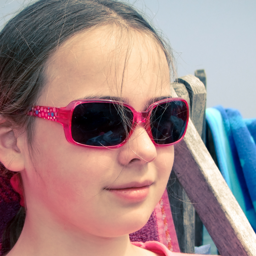} & \includegraphics[width=0.14\linewidth]{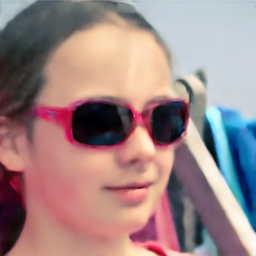}  & \includegraphics[width=0.14\linewidth]{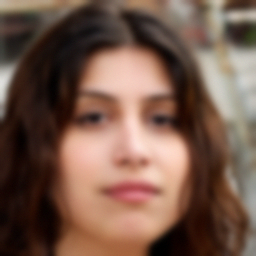} &
\includegraphics[width=0.14\linewidth]{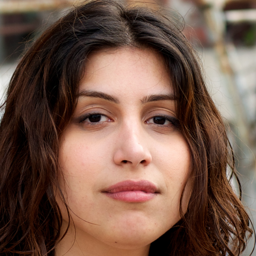} & \includegraphics[width=0.14\linewidth]{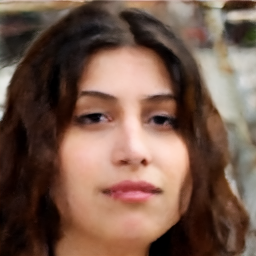} \\

\includegraphics[width=0.14\linewidth]{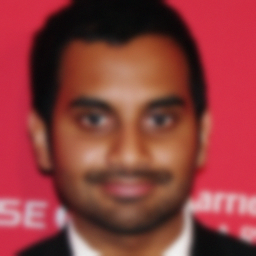} & 
\includegraphics[width=0.14\linewidth]{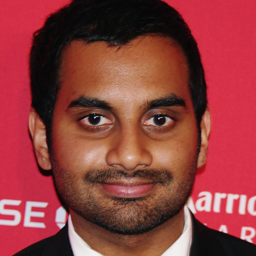} & \includegraphics[width=0.14\linewidth]{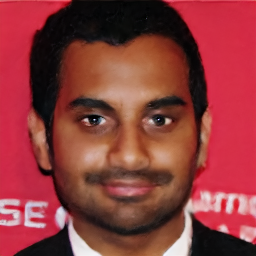}  & \includegraphics[width=0.14\linewidth]{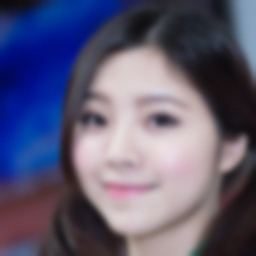} &
\includegraphics[width=0.14\linewidth]{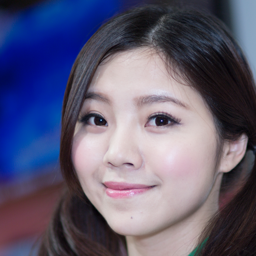} & \includegraphics[width=0.14\linewidth]{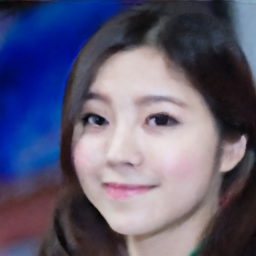} \\

\includegraphics[width=0.14\linewidth]{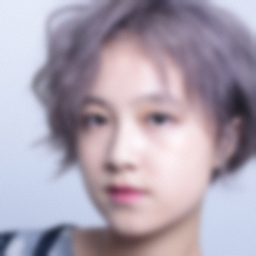} & 
\includegraphics[width=0.14\linewidth]{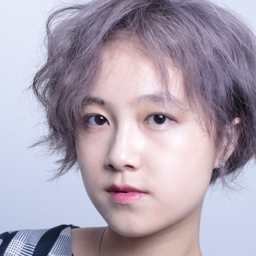} & \includegraphics[width=0.14\linewidth]{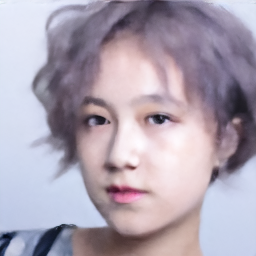}  & \includegraphics[width=0.14\linewidth]{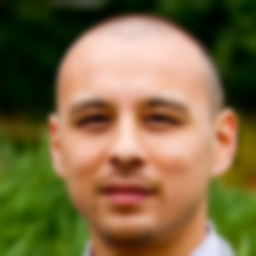} &
\includegraphics[width=0.14\linewidth]{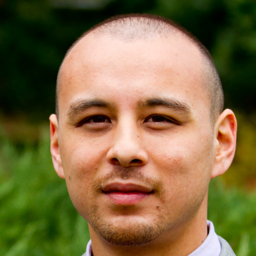} & \includegraphics[width=0.14\linewidth]{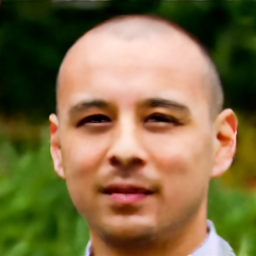} \\ 

\end{tabular}
}

\caption{Qualitative results on the image deblurring task.}
\label{fig:a3}
\end{figure}

\clearpage
% \bibliographystyle{splncs04}
% \bibliography{main}

\end{document}